%% file: aij_paper.tex
\documentclass[3p, a4paper]{elsarticle}
\usepackage{geometry}
\include{macros}

\usepackage{array}
\newcolumntype{L}[1]{>{\raggedright\let\newline\\\arraybackslash\hspace{0pt}}m{#1}}
\newcolumntype{C}[1]{>{\centering\let\newline\\\arraybackslash\hspace{0pt}}m{#1}}
\newcolumntype{R}[1]{>{\raggedleft\let\newline\\\arraybackslash\hspace{0pt}}m{#1}}

\theoremstyle{thmstyleone}%
\theoremstyle{thmstyletwo}%

\theoremstyle{thmstylethree}%
\newtcbtheorem[auto counter, number within = section]{box_definition}{Definition}{colback=gray!10, colframe=black, coltitle=white}{def}

\newtcbtheorem[auto counter, number within = section]{box_theorem}{Theorem}{colback=gray!10, colframe=black, coltitle=white}{thm}

\newtcbtheorem[auto counter, number within = section]{box_lemma}{Lemma}{colback=gray!10, colframe=black, coltitle=white}{lem}

\newtcolorbox{overview}[1][]
{
	breakable,
	colframe=blue!25,
	colback=blue!10,
	coltitle=black,
	title=\textbf{Overview}
}

\newif\ifshownotation
\shownotationtrue

\ifshownotation
	\newtcolorbox{notation}[1][]{
	  breakable,
	  parbox=false,
	  left=2mm,
	  right=2mm,
	  top=1mm,
	  bottom=1mm,
	  colframe=green!25,
	  colback=green!10,
	  coltitle=black,
	  title=\textbf{Notation},
	  before upper={
	    \setlength{\parindent}{0pt}%
	    \noindent\ignorespaces%
	    \setlength{\tabcolsep}{0pt}%
	  },
	  after upper={
	    \par\noindent\ignorespacesafterend%
	  },
	}
	\newcommand{\entry}[2]{%
	  \begin{tabularx}{\linewidth}{%
	    @{}>{\raggedright\arraybackslash\hsize=.25\hsize}X%
	    @{}>{\raggedright\arraybackslash\hsize=1.75\hsize}X@{}%
	  }
	    #1 & #2
	  \end{tabularx}%
	}
	\newcommand{\notsep}{\par\vspace{2pt}\noindent}
\else
  \newtcolorbox{notation}[1][]{blanker,#1}
  \newcommand\entry{m m}{}
  \newcommand{\notsep}{}
\fi

\usepackage{amssymb}
\usepackage{amsmath}

\journal{Nuclear Physics B}

\begin{document}

\begin{frontmatter}



\title{Life, uh, Finds a Way: Hyperadaptability by Behavioral Search}


\author{Alex Baranski\corref{cor1}}
\ead{alexander.baranski@oist.jp}

\author{Jun Tani}
\ead{jun.tani@oist.jp}

\affiliation{organization={Okinawa Institute of Science and Technology},
            addressline={1919-1 Tancha}, 
            city={Onna},
            postcode={1904-0412}, 
            state={Okinawa},
            country={Japan}}

\cortext[cor1]{Corresponding author}

\begin{abstract}
Living beings are able to solve a wide variety of problems that they encounter rarely or only once. Without the benefit of extensive and repeated experience with these problems, they can solve them in an ad-hoc manner. We call this capacity to always find a solution to a physically solvable problem \emph{hyperadaptability}. To explain how hyperadaptability can be achieved, we propose a theory that frames behavior as the physical manifestation of a self-modifying search procedure. Rather than exploring randomly, our system achieves robust problem-solving by dynamically ordering an infinite set of continuous behaviors according to simplicity and effectiveness. Behaviors are sampled from paths over cognitive graphs, their order determined by a tight behavior-execution/graph-modification feedback loop. We implement cognitive graphs using Hebbian-learning and a novel harmonic neural representation supporting flexible information storage. We validate our approach through simulation experiments showing rapid achievement of highly-robust navigation ability in complex mazes, as well as high reward on difficult extensions of classic reinforcement learning problems. This framework offers a new theoretical model for developmental learning and paves the way for robots that can autonomously master complex skills and handle exceptional circumstances.
\end{abstract}

\begin{graphicalabstract}
\includegraphics[width=\textwidth]{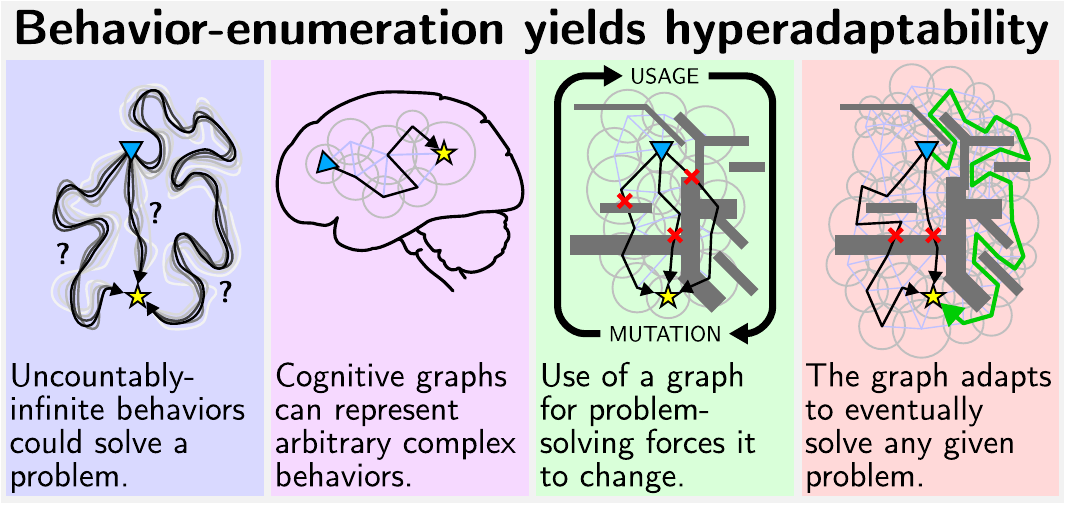}
\end{graphicalabstract}

\begin{highlights}
\item A hyperadaptable agent can solve any problem in real-time.
\item A complete (modulo equivalence) search over the set of all continuous behaviors is sufficient for hyperadaptability.
\item This search be guided by a self-organizing neural graph.
\end{highlights}


\end{frontmatter}


\section{Introduction}
What sets organisms apart from machines? Intuitively, it is their extraordinary adaptability to unforeseen circumstances, allowing them to create new behaviors to achieve their goals. For much of artificial intelligence research, this is pursued by designing models that can learn complex representations that generalize outside of their training data. However, generalization requires extensive experience and can be quite unreliable in totally new problems. For tasks that are repeated often this is acceptable, but most meaningfully distinct problems faced by an agent in the physical world will be encountered rarely or only once, providing no opportunity for repeated interactions to build up a generalizable model. 

Consider, for example, walking in an urban environment. Most people live in urban environments, and spend considerable time walking over flat, even surfaces, occasionally avoiding puddles, garbage, or potholes. If such a person is forced to bushwhack (traverse difficult, wild terrain full of brambles, gorges, boulders, and unstable ground), their automatic ``walking program'' will be severely strained; usually, ``walking'' will be forced to transform into a mixture of climbing, crawling, and sliding. Despite these difficulties, however, their lack of experience will only slow their progress, not halt it. Tellingly, the physical effort of bushwhacking will be matched by the mental toll of figuring out \emph{how} to bushwhack.

We contend that instead of powerful generalization abilities that require possession of relevant prior experience (which sometimes is unavailable, as in the case of our stranded urbanite), the adaptability of agents instead stems from an ability to (ideally) always eventually solve problems with physically achievable solutions by searching in real-time for a goal-achieving behavior, with or without good generalization. This is in part because in the real world, a problem faced by an agent often \emph{has} to be solved \emph{now}. We call this knowledge-invariant ability to always find a successful behavior in finite time \emph{hyperadaptability}, which is the conceptual complement to Ross Ashby's \emph{ultrastability}. While ultrastable systems \emph{maintain} a state in the face of arbitrary perturbations, \emph{hyperadaptable} systems \emph{achieve} arbitrary states via the construction of totally new behaviors. They accomplish this by using \emph{search} as a fundamental and universal problem-solving mechanism. By using search as an architectural primitive, we place problem-solving as prior to learning: instead of designing an agent that collects (essentially random) experiences and thereby learns how to solve problems, we envision an agent that has a very basic form of problem-solving out-of-the box, allowing it to quickly gain much higher-quality experiences from which it could, in-principle, learn better representations.

The hyperadaptability concept actually disentangles \emph{whether} a problem can be solved from \emph{how quickly} the problem can be solved. Relevant prior knowledge (crystal intelligence) can always be leveraged to quickly solve a problem, and excellent fluid intelligence can be used to rapidly recombine existing behaviors to solve a new problem. Hyperadaptability can interact with fluid and crystal intelligence, but technically does not \emph{rely} on them. The main difference between a smart, knowledgeable hyperadaptable agent and a ``rudimentary'' one is that the rudimentary hyperadaptable agent will take longer to solve any given problem. Within this framing, an astute urbanite might be able to quickly transfer their prior experience at a jungle gym to bushwhacking, whereas a slightly duller individual might struggle for longer and have to spend more time learning the hard way.

To operationalize hyperadaptability, we take inspiration from the hippocampus and entorhinal cortex, regions of the mammalian brain associated with learning, memory, and navigation. In particular, these regions are believed to implement cognitive graphs \cite{tolman1948cognitive} and cognitive maps \cite{peer2021structuring}, mental structures hypothesized to help animals organize learning and inference \cite{stachenfeld2017hippocampus, behrens2018cognitive}. There is strong evidence such graph-maps are used for a variety of non-spatial-navigation tasks, both across hippocampus \cite{deshmukh2011representation, julian2018human} and neocortex \cite{constantinescu2016organizing, long2021novel}. Beyond the biological motivation, we also think there is a practical one: in order for a search over behaviors to make sense, an agent needs to be able to try something, and after it fails, try something totally different. Representing behaviors implicitly in the weights of a deep network makes this very difficult, but representing behaviors semi-explicitly in the connections of a rapidly modifiable neural graph makes it much easier.

In this paper, we introduce and formalize the concept of hyperadaptability to explain the extreme adaptability of living organisms. We argue that hyperadaptability can be achieved by framing behavior as a kind of physically executed search, rather than a policy per-se, and we show how cognitive graphs can be used to implement this search. Under some fairly general criteria we prove that correct application of certain mutation operations to a cognitive graph can guarantee hyperadaptability, and we introduce a concrete, neurally implemented algorithm which we argue is hyperadaptable. We perform experiments to demonstrate the hyperadaptability of the algorithm, and we test the algorithm against a reinforcement learning (RL) baseline on several difficult variants of a classic RL problem, showing very strong performance. We finish by discussing the implications of this work and its limitations.

Our goal in this paper is not to advocate for symbolic architectures, it isn't to produce a superior planning algorithm, nor a state-of-the-art RL agent, and it isn't to provide a detailed biophysical model of cognition. Instead, we are aiming to introduce a general \emph{theory} for rapid online problem solving in humans and other animals, with an implementation that could in-principle be emulated by biological circuits. The graph-based formalism and subsequent algorithm that we introduce should be viewed mostly as conceptual scaffolding: by implementing this formalism with vectors and Hebbian learning, we leave open the possibility in the future of various extensions, approximations, and continuous relaxations of our system, whose forms \emph{echo} our formalism without strictly conforming to it.

\subsection{Related Work}
As we are discussing the question of how agents can adapt their behavior without previous experience (training data), our work falls squarely under the domain of reinforcement learning (RL). Unlike in classic RL though, we strip away many of the assumptions that undergird most algorithms. In order to create a setting as close to the real-world as possible;
\begin{itemize}
\item[-] We use no external resets of the environment, each problem has to be solved ``on-line''.
\item[-] We provide no extrinsic reward or only a single reward-spike upon goal-completion.
\item[-] We assume that problems come from a fat-tailed distribution, many unique problems are encountered once or only a few times \cite{clauset2009power, chan2022zipfian}, so learning a general and optimal policy is irrelevant and/or impossible, only solving the current problem quickly matters.
\end{itemize}
In contrast, classic RL assumes an episodic structure provided externally by the engineer, dense extrinsic rewards are used to define the task and guide the agent towards a useful behavior, and the goal of learning is to find a general policy that will work well on ``similar'' tasks. Several sub-areas of RL are more closely aligned to our setting: we are essentially operating under a combination of reset-free/single-life RL \cite{gupta2021reset, chen2022you}, intrinsic/reward-free RL \cite{martin2017count, jin2020reward}, and sample-efficient episodic control \cite{blundell2016model}. One major difference is our theoretical framework: RL is traditionally based on iteration of the Bellman update or an approxmation thereof in the hopes of finding an optimal policy, which is only guaranteed in tabular settings. In contrast, we \emph{start} from the assumption of a continuous spatiotemporal domain, and ask how we can guarantee the discovery of a solution, regardless of its optimality.

Within artificial intelligence, search, planning, and graphs are deeply intertwined. For discrete domains, the classic $\mathrm{A}^*$ pathfinding algorithm \cite{hart1968formal} finds optimal paths over a given graph. In the field of robotics, search has been successfully applied to continuous domains using techniques such as Rapidly-exploring Random Trees \cite{lavalle1998rapidly} and Probabilistic Roadmaps \cite{kavraki1996probabilistic}, which incrementally construct a graph over which planning can occur. Even within deep learning and deep reinforcement learning, Monte-Carlo search has been applied in architectures like MuZero \cite{schrittwieser2020mastering} to master complex tasks such as Atari games. 

The innately compositional structure of graphs also offers many advantages beyond planning. Graph Neural Networks \cite{scarselli2008graph} have been used to achieve state-of-the-art performance on a wide variety of structured tasks, extending the use of deep neural networks to non-parametric problems. From a cognitive-science perspective, the Tolman-Eichenbaum Machine (TEM) \cite{whittington2020tolman} shows that cognitive graphs can be used to organize inferences using learned structured representations that take advantage of the natural compositionality of graphs to efficiently generalize from a small amount of experience.

\subsection{How to Read this Paper}
The full, formal, and detailed description of the theory of behavioral search is quite long, and involves the introduction of many mathematical objects used to build intuition, as well as construct more complex mathematical objects. To mitigate the risk that the reader will get lost, at the beginning of each ``theory'' section (3, 4, 5, and 6) we provide a short overview of the section, as well as ``Notation'' box that acts as a glossary for notation introduced in that section. A less motivated reader could just read the overviews and have a basic grasp of the theory.

In \textbf{section 2} we give a general overview of our theory. In \textbf{section 3} we define some foundational terminology that later sections build off of. In \textbf{section 4} we formally define hyperadaptability, and provide a general recipe for enumerating the set of all behaviors. In \textbf{section 5} we introduce segraphs, a data-structure corresponding to a cognitive graph that organizes the enumeration. In \textbf{section 6} we describe how segraphs self-organize to actually perform the enumeration. In \textbf{section 7} we describe the concrete algorithm for performing behavioral search. In \textbf{section 8} we provide the details of the neural instantiation of segraphs. In \textbf{section 9} we present our experimental results of using the algorithm from section 7. In \textbf{section 10} we discuss our findings. We leave some details of methods for the appendices.

\subsection{Notational Conventions}

Scalars are denoted by normal lower-case italic letters ($a$), vectors by bold lower-case italic letters ($\vect{a}$), while matrices are denoted by bold upper-case letters ($\mat{A}$). Ordered sequences (and conceptually adjacent objects) are denoted by lower-case Fraktur-font letters ($\mathfrak{a}$). Sets of such sequences are denoted by upper-case Fraktur-font letters ($\mathfrak{A}$). If $\mathfrak{a}$ is a sequence, or $\vect{a}$ is a vector, then $\mathfrak{a}\idx{j}$ and $\vect{a}\idx{j}$ are the $0$-based j\textsuperscript{th} index of $\mathfrak{a}$ and $\vect{a}$, respectively. For finite-length $\mathfrak{a}$, if $j<0$ we let $\mathfrak{a}\idx{j} = \mathfrak{a}\idx{|\mathfrak{a}| + j}$, so that $\mathfrak{a}\idx{\n{1}}$ is the last item of $\mathfrak{a}$, $\mathfrak{a}\idx{\n{2}}$ the second-to-last item of $\mathfrak{a}$, etc. When appropriate, we will sometimes extend this notation to subscripts, so we may write that a sequence $\mathfrak{a} = (a_0, a_1, a_2, ... a_{\n{1}})$, with $a_{\n{1}}$ indicating the last element of the sequence.

To provide a reference for the reader, we provide a glossary of all symbols introduced in this paper in \ref{sec:notation_glossary}.

\section{Overview of Behavioral Search Theory}
Our goal in sections 3-6 is to lay-out, in detail, a theory of how an agent can eventually solve any problem, regardless of prior knowledge. To do this, we have to imagine what it means to have \emph{no} prior knowledge. Any random behavior that an agent tries is almost sure to fail: how can the agent, from so little, eventually find a solution to its problem? The key is in the word ``find'': as the agent has no prior knowledge, it cannot rely on simply \emph{knowing} the solution, it has to perform a \emph{search} for a solution.

Imagine an agent in a statespace; the statespace could be physical space, it could be the configuration space of a body, it could be an abstract space. The agent has a \emph{problem} whenever it is not currently in the state it wants to be. Under this view, a \emph{solution} is simply a behavior that takes the agent to its goal. To formally characterize a behavior, we make a distinction between a \emph{plan}, or high-level strategy for reaching a goal, and an \emph{execution}, the physical trajectory generated by following the plan using a controller. This endows behaviors with a teleological aspect, allowing our agent to try a behavior and either \emph{succeed} or \emph{fail}. To solve a given problem merely requires that a single solution be found, so all plans that solve the same problem are considered to be equivalent.

For an agent to be \emph{hyperadaptable}, it must be able to find a solution to any problem in finite time. Doing this without prior knowledge seems to necessitate an exhaustive search over the set of all plans, modulo the equivalence of solutions to the same problem. But plan-space is uncountably infinite! Thankfully, we can construct a set of plans that is countably infinite and dense in the set of all plans, having much the same relationship as that between the rational and real numbers. Using this set of ``rational'' plans, we can approximate to arbitrary precision any ``real'' behavior. Ultimately, this construction involves systematically growing a set of waypoints, corresponding to subgoals of the agent's plans, which in the limit is dense in the agent's statespace.

A naive enumeration, while technically possible, would be extremely inefficient. There is at least one simple and problem-agnostic way to bias the search over behaviors \emph{without} making it incomplete, which relies on the fact that many plans can share the same subplan. If a subplan fails, any plan containing that subplan will also fail, and any plans containing a \emph{similar} subplan will be \emph{likely} to fail. Lacking prior knowledge, our agent has no idea what the relationship between subplan similarity and subplan failure likelihood is, so it is forced to make a non-binding \emph{guess} about the appropriate scale of generalization. After a subplan fails, the agent will temporarily generalize the failure to similar subplans and avoid re-using them, \emph{until} other options have been exhausted. At this point, the agent may revisit failed subplans in more detail.

In order to simplify this search, we initially give our agent a finite set of waypoints, which it can add to as needed. In order to make the search intelligent and adaptive, we use a graph whose vertices correspond to the waypoints. Each graph vertex encodes the agent's ``guess'' about the scale of generalization, which we represent as a hyperball around the waypoint. The graph's edges perform bookkeeping on subplan success and failure, which is then used to guide the selection of paths, as well as the selection of new waypoints to add to the set. By interleaving the addition of new waypoints ``outside'' the graph with the addition of new waypoints ``inside'' the graph, we can guarantee that in the limit the full set of waypoints is dense in the statespace, but at every step is finite. In the limit, any plan can be generated, allowing the agent's search to be complete. Because the search is complete, any behavior needed to solve a problem is eventually found, making the agent hyperadaptable.

\section{Problems, Behaviors, and Solutions}\label{sec:problems_behaviors_solutions}
\begin{overview}
An agent inhabits some kind of statespace; a problem is simply the task of getting from one state to another (a goal). We model goal-directed behavior with \emph{plans} (sequences of subgoals or \emph{waypoints}) executed by a low-level controller. A plan that when executed reaches the goal is a solution; if it still succeeds under bounded perturbation, it is \emph{robust}. Over the entire set of possible robust plans (robust plan-space), we define solutions to be equivalent if they solve the same problem. This equivalence relation partitions the robust plan-space into \emph{achievement classes}, so solving a problem reduces to finding just one representative from the class corresponding to that problem.
\end{overview}

\begin{notation}
\entry{$\agent$}{An agent} \notsep
\entry{$\SS$}{A Euclidean \textbf{statespace}} \notsep
\entry{$\SS_0$}{The subset of $\SS$ physically accessible from the agent's starting state} \notsep
\entry{$\s, \goal{\s}$}{A state in $\SS$, and a goal-state in $\SS$ (together, a \textbf{problem})} \notsep
\entry{$\K$}{A statefull, goal-oriented \textbf{controller} that tries to reach a goal-state} \notsep
\entry{$\Ks(\vect{h}_t)$}{Controller \textbf{success} ($0$ for not yet, $1$ for success)} \notsep
\entry{$\Kf(\vect{h}_t)$}{Controller \textbf{failure} ($0$ for not yet, $1$ for failure)} \notsep
\entry{$\termtime{\s}{\goal{\s}}$}{Termination time (either $\Ks(\vect{h}_t)=1$ or $\Kf(\vect{h}_t)=1$) for controller $\K$ trying to reach $\goal{\s}$ from $\s$} \notsep
\entry{$\trajK(\s, \goal{\s})$}{Finite physical \textbf{trajectory} generated by controller $\K$ going from $\s$ to $\goal{\s}$} \notsep
\entry{$\rK(\s, \goal{\s})$}{\textbf{Reachability} for controller $\K$ from $\s$ to $\goal{\s}$ (failure=$0$, success=$1$)} \notsep
\entry{$\conK(\s)$}{The points reachable by controller $\K$ from state $\s$} \notsep
\entry{$\preK(\s)$}{The points that can be reached from state $\s$ using controller $\K$} \notsep
\entry{$\plan$}{A \textbf{plan}, or sequence of states (waypoints)} \notsep
\entry{$\trajK(\plan)$}{Finite physical \textbf{trajectory} generated by controller $\K$ following plan $\plan$} \notsep
\entry{$\rK(\plan)$}{\textbf{Reachability} for controller $\K$ following plan $\plan$ (failure=$0$, success=$1$)} \notsep
\entry{$\ball{\s}{\epsilon}$}{A hyperball around $\s$ with radius $\epsilon$, $\ball{\s}{\epsilon} = \{\s' \in \SS : \norm{s' - s} \leq \epsilon\}$} \notsep
\entry{$\splans[][]{}{}$}{The \textbf{set of all plans}. This symbol admits many variations. The set of plans of length $n$ is $\splans[][]{}{n}$, of any length, $\splans[][]{}{*}$. The set of plans with waypoints in the set $A$ is $\splans[][]{A}{}$. The set of robust plans is $\splans[r][]{}{}$. The set of plans that start at $\s$ and end at $\s'$ is $\splans[][]{}{}(\s, \s')$} \notsep
\entry{$\tube{\eps}(\plan)$}{The radius-$\eps$ tube around plan $\plan$ of alternative plans} \notsep
\entry{$\stube{\eps}(\plan)$}{The subset of $\tube{\eps}(\plan)$ that is successful} \notsep
\entry{$\rKe{\eps}(\plan)$}{The fraction of $\tube{\eps}(\plan)$ that succeeds, if $\rKe{\eps}(\plan)=1$ for $\eps>0$ then $\plan$ is \textbf{robust}} \notsep
\entry{$\SSK$}{The subset of statespace reachable using the controller $\K$} \notsep
\entry{$\plan_1 \sim \plan_2$}{Plans are equivalent if they are both solutions to the same problem, starting at the same point $\start{\s}$, ending at the same point $\goal{\s}$, and succeeding, meaning $\rK(\plan_1) = 1$ and $\rK(\plan_2) = 1$. Equivalently, $\plan_1, \plan_2 \in \splansf[r][]{}{*}{\start{\s}, \goal{\s}}$} \notsep
\entry{$\achclass{\SSK}{\s}{\s'}$}{Equivalence class of plans from $\s$ to $\s'$ with waypoints in $\SSK$, equal to $\splansf[r][]{\SSK}{*}{\s, \s'}$} \notsep
\entry{$\achset{}{\SSK}$}{The set $\splans[r][]{\SSK}{*}$ partitioned by $\sim$, i.e. the \textbf{set of solveable problems} in $\SSK$, or the set of all equivalence classes of plans with waypoints in $\SSK$} \notsep
\entry{$\achset{\agent}{\SSK}$}{The set of equivalence classes of solutions with waypoints in $\SSK$ that the agent $\agent$ can generate}
\end{notation}

Imagine that an agent $\agent$ inhabits a Euclidean statespace $\SS$, upon which it can act through an action space $\actionspace$ via some deterministic dynamics. In this work, we do not consider the problem of \emph{finding} nor \emph{accessing knowledge about} such a statespace, nor do we consider the problem of \emph{noise}, instead granting the agent direct access to $\SS$ and the immediate consequences of its actions on its state. A \emph{problem} is when the agent \emph{wants} to be at a specific state $\goal{\s} \in \SS$, but is currently in a different state $\s\in\SS, \s\neq\goal{\s}$. A \emph{solution} is a \emph{behavior} that takes the agent from $\s$ to $\goal{\s}$. In general, not all of statespace may be physically accessible to our agent. We refer to the subset of $\SS$ accessible from the agent's starting state as $\SS_0 \subseteq \SS$. While our agent may have a specific goal it wants to reach, a hyperadaptable agent should be able to eventually find the right behavior to reach \emph{any} goal in $\SS_0$.

What exactly \emph{is} a behavior? Ultimately, we will use the word to loosely refer to three different levels of abstraction:
\begin{enumerate}
\item A physically executed trajectory.
\item A plan that deterministically generates a physically executed trajectory.
\item A probability distribution over plans.
\end{enumerate}
At first glance, the first sense would seem to be sufficient, but later we will want to speak of specific behaviors ``failing'' or ``succeeding'', giving \emph{behaviors} a teleological component that is lacking in a mere \emph{physical trajectory}. Tautologically, a physical trajectory reaches the final point it reaches, it cannot ``fail'' or ``succeed'' to reach that point, as there is no intrinsic aspirational quality to such a trajectory: by definition it reaches its final point. In order to grant behaviors a proper teleology, we restrict our consideration to behaviors that are generated by a stateful, goal-directed low-level controller:
\begin{box_definition}{Controller}{controller}
A controller is a tuple $(\K, \Ks, \Kf)$, with:
\begin{itemize}
\item $\K$ a function $\vect{a}_t, \vect{h}_{t+1} = \K(\s_t, \goal{\s}, \vect{o}_t, \vect{h}_t)$ which maps a current state $\s_t$, a target state $\s$, any additional observations $\vect{o}_t$, and a hidden state $\vect{h}_t$ to an action $\vect{a}_t$ and the controllers next hidden state $\vect{h}_{t+1}$.
\item $\Ks$ an indicator function that decides if the controller has reached it's target. If it has, then $\Ks(\vect{h}_t) = 1$, otherwise, $\Ks(\vect{h}_t) = 0$.
\item $\Kf$ an indicator function that decides if the controller should \emph{give up} on trying to reach its target. If so, then $\Kf(\vect{h}_t) = 1$, otherwise $\Kf(\vect{h}_t) = 0$.
\end{itemize}
This definition can be suitably modified for continuous-time domains, which anyways must be discretized for computer implementation. We generally refer to the whole controller $(\K, \Ks, \Kf)$ simply as $\K$ for brevity, mentioning $\Ks$ and $\Kf$ only when necessary.
\end{box_definition}

\begin{figure}
\includegraphics[width=\textwidth]{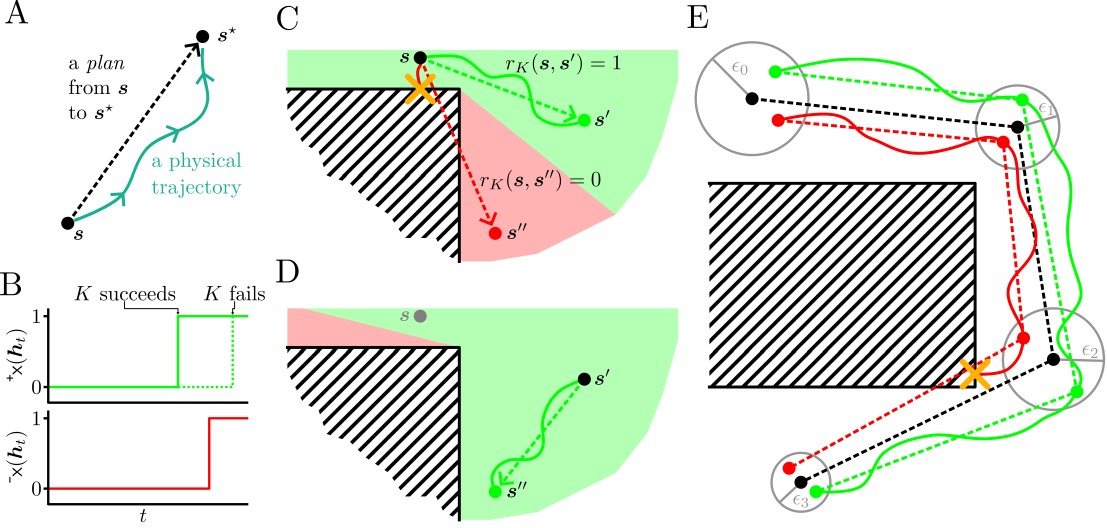}
\caption{\label{fig:problems}(A) A plan is a teleological object, representing an intention to go from one state to another. Using a controller to follow a plan generates a physical trajectory. If the physical trajectory reaches the target, the plan \emph{succeeds}, otherwise it \emph{fails}. (B) Termination is ultimately determined by the relative timing of the success and failure functions, $\Ks$ and $\Kf$. (C) Some points, such as $\s'$, can be reached by the controller from $\s$. Many points such as $\s''$ cannot, however. (D) Some points that cannot be reached in one step can be reached in multiple. If the agent first goes to $\s'$, then it can subsequently reach $\s''$. (E) Complex plans (black dashed line) can be created by sequentially composing waypoints (black dots), generating more complex physical trajectories. For each plan we can define a neighborhood of nearby plans, called it's \emph{$\eps$-tube}. Alternative plans can be sampled from the $\eps$-tube. Some may succeed (green behavior), others may fail (red behavior), the more that succeed, the more robust the plan is.}
\end{figure}

The start, target point-pair  $(\s, \s^{\star})$ corresponds to a very simple plan (Figure \ref{fig:problems}A). While both $\Ks(\vect{h}_t)=0$ and $\Kf(\vect{h}_t)=0$, the controller generates actions to reach the target, generating a physical trajectory or \emph{execution}. As soon as $\Ks(\vect{h}_t)=1$ or $\Kf(\vect{h}_t)=1$, the execution is terminated, and the plan either \emph{succeeds} or \emph{fails}, respectively (Figure \ref{fig:problems}B). $\Kf$ is defined so that it will always eventually trigger, meaning that every execution ends at some finite time $\termtime{\s}{\goal{\s}}>0$. We denote the execution by $\K$ of the plan to reach $\goal{s}$ from $\s$ as $\trajK(\s, \goal{\s}) : [0, \termtime{\s}{\goal{\s}}] \rightarrow \SS$. This is a ``behavior'' in the first sense. We define a \emph{reachability function} $\rK(\s, \goal{\s})$ to indicate whether or not $\K$ reaches the goal $\goal{s}$, which equals $1$ if $\Ks(\vect{h}_{\termtime{\s}{\goal{\s}}}) = 1$ and equals $0$ if $\Kf(\vect{h}_{\termtime{\s}{\goal{\s}}}) = 0$ (Figure \ref{fig:problems}C). Our method is comparable to, and formally, a specific manifestation of, the \emph{options framework} \cite{sutton1999between}.

We define $\conK(\s) = \{\goal{\s}\in\SS : \rK(\s, \goal{\s}) = 1\}$ as the set of points that $\K$ can reach from $\s$, and $\preK(\goal{\s}) = \{\s\in\SS : \rK(\s, \goal{\s}) = 1\}$ as the set of points from which $\s^{\star}$ can be reached by $\K$. For any sufficiently complex environment, the controller will not be able to reach most targets from most starting points, that is, $\forall \s\in\SS_0, \conK(\s) \subset \SS_0$ (Figure \ref{fig:problems}C). It may be the case, however, that after reaching a point $\s'$ in $\conK(\s)$, \emph{new} points are reachable (Figure \ref{fig:problems}D). That is, it is possible that $\exists \s' \in \conK(\s)$ such that $\conK(\s') - \conK(\s) \neq \varnothing$. If this is the case, then a point $\s' \in \conK(\s) \cap{} \preK(\s'')$ can act as a \emph{waypoint} between $\s$ and $\s''$. We call a sequence of waypoints $\plan = (\s, \s_1, \s_2, ... \s_{n-1}, \s^{\star})$ a \emph{plan} from $\s$ to $\s^{\star}$ (Figure \ref{fig:problems}E). A plan is a behavior in the second sense.

We denote the composite-execution generated by $\K$ following the plan $\plan$ as $\trajK(\plan)$, which like a ``single-waypoint'' execution, always terminates at some finite time $\plantermtime{\plan}$. We say that the plan succeeds if every sub-plan succeeds, but if any of the sub-plans fails (a waypoint isn't reached), the entire plan fails, and the execution terminates at a non-target point. We denote success in reaching the final target as $\rK(\plan) = 1$ and failure as $\rK(\plan) = 0$. We say that a plan $\plan$ is \emph{robust} if every ``nearby'' plan $\plan$ succeeds. To formalize this, we consider the set $\splans[][]{S}{n}$ of length $n$ plans with waypoints in $\SS$, and introduce the \emph{$\eps$-tube}:

\begin{box_definition}{$\eps$-tube}{epsilon-tube}
Let $\ball{\s}{\epsilon} = \{\s'\in\SS : \norm{\s-\s'}\leq\epsilon\}$ be the $\epsilon$-neighborhood of $\s$, a hyperball with radius $\epsilon$ centered on $\s$. Let $\plan = (\s_0, \s_1, \s_2, ... \s_n)$ be a plan, and $\eps = (\epsilon_0, \epsilon_1, \epsilon_2, ... \epsilon_n)$ a corresponding sequence of positive numbers. Then
\begin{equation*}
\tube{\eps}(\plan) = \{(\s_0', \s'_1, \s'_2, ... \s'_n) \in \splans[][]{S}{n} : \forall i\in[0, n], \s'_i \in \ball{\s_i}{\epsilon_i}\}
\end{equation*}
is the \emph{$\eps$-tube} around $\plan$, a bundle of ``similar'' plans. (See Figure \ref{fig:problems}E for example).
\end{box_definition}
\noindent{}Ultimately, we are not interested in non-robust plans: if a plan requires infinitely precise control in order to successfully execute it, then the plan is effectively useless under conditions with noisy physics, noisy controllers, or noisy sensors. Let $\stube{\eps}(\plan) = \{\plan' \in \tube{\eps}(\plan) : \rK(\plan') = 1\}$ be the set of all plans in the $\eps$-tube of $\plan$ which \emph{succeed}. We extend the function $\rK(\plan)$ (the ``reachability'' of the plan) to it's $\eps$-tube, and refer to the \emph{$\eps$-reliability} of $\plan$ as:
\begin{equation}
\rKe{\eps}(\plan) = \tfrac{|\stube{\eps}(\plan)|}{|\tube{\eps}(\plan)|}
\label{eq:eps-reliability}
\end{equation}
which is the fraction of successful plans in the $\eps$-tube of $\plan$. We say that the plan $\plan$ is \emph{robust} if for some $\eps>0$, $\rKe{\eps}(\plan) = 1$.

Now, we let $\splans[r][]{S}{n}$ be the set of all plans of length $n$ that are robust, and $\splans[r][]{S}{*}$ the set of all such plans of any finite length. Then, slightly overloading notation, we say that $\conK^*(\s) \subset \SS$ is the set of states that can be reached by $\K$ via some plan in $\plans^*$. If $\s_0$ is the initial state of the agent $\agent$, then $\SS_{\K} = \conK^*(\s_0)$ is the subset of $\SS$ that $\agent$ can reach. For the rest of this work, we assume that there are neither ``Garden of Eden'' nor ``black-hole'' states\footnote{States or regions which once left, can never be returned to, and states or regions which once entered, can never be left.}, meaning that $\forall \s'\in\conK^*(\s), \conK^*(\s') = \conK^*(\s)$. Any state in $\SS_{\K}$ is a potential problem that is physically solvable by our agent. An agent that is \emph{hyperadaptable} can eventually solve any problem (reach any state) in $\SSK$, meaning that it must be able to eventually find \emph{a plan that reaches the goal}.

$\splans[r][]{\SSK}{*} \subset \splans[][]{\SSK}{*}$ is the set of all possible robust plans with waypoints in $\SSK$, with $\splansf[r][]{\SSK}{*}{\s, \s'}$ the restriction of this set to only plans which start at $\s$ and end at $\s'$. By definition, every plan in $\splansf[r][]{\SSK}{*}{\s, \s'}$ is a viable way to go from $\s$ to $\s'$, so from the perspective of solving a problem (going from one point to another), every plan in $\splansf[r][]{\SSK}{*}{\s, \s'}$ is equivalent (Figure \ref{fig:behavior_space}A). We say that $\plan \sim \plan' \iff \plan, \plan' \in \splansf[r][]{\SSK}{*}{\s, \s'}$, and refer to the equivalence class $\splansf[r][]{\SSK}{*}{\s, \s'}$ as an \emph{achievement class}, denoted $\achclass{\SSK}{\s}{\s'}$. The set of all achievement classes partitions $\splans[r][]{\SSK}{*}$, yielding a quotient set (Figure \ref{fig:behavior_space}B):
\begin{equation*}
\achset{}{\SSK} = \splans[r][]{\SSK}{*}\hspace{-0.2em}\backslash{}{\sim} = \{\achclass{\SSK}{\s}{\s'} : \s, \s' \in \SSK\}
\end{equation*}
which effectively represents the existence of solutions to problems inside of $\SSK$, called the \emph{capability set} of $\SSK$. If $\agent$ is an agent, and $\splans[r][]{\agent}{*}$ is the set of behaviors that can be generated by $\agent$, then $\achset{\agent}{\SSK} = \splans[r][]{\agent}{*}\hspace{-0.2em}\backslash{}{\sim}$ represents the \emph{capability of} $\agent$.

\section{Hyperadaptability and Behavioral Search}
\begin{overview}
An agent is defined as \emph{hyperadaptable} if, for every solvable problem, it can find a solution in finite time. Without prior knowledge, this requires search, or \emph{enumeration}. Because plan-space is uncountably infinite, direct enumeration is impossible. Instead, we construct a countable set of ``rational'' plans that is dense in plan-space. We start with a finite set of waypoints, generating finitely many plans. Each waypoint has a corresponding hyperball, with which we can define \emph{mutation} operations to grow the set of waypoints. A hyperball can be \emph{extended} by adding new larger hyperballs around it, placing waypoints in new regions of statespace, or it can be \emph{refined}, replaced with new smaller hyperballs inside it to increase the local density of waypoints. A \emph{balanced} sequence of these mutations in the limit makes the waypoint set dense in the statespace, ensuring that search over the corresponding plan-set will eventually visit at least one member of every achievement class. Finally, because plans share sub-plans, the agent can intelligently bias the search by deprioritising candidates that reuse failed segments, making an otherwise intractable search a practical problem-solving strategy.
\end{overview}

\begin{notation}
\entry{$\WP$}{A discrete set of waypoints in the statespace $\SS$} \notsep
\entry{$E$}{A function assigning an $\epsilon$ to each waypoint in $\WP$} \notsep
\entry{$B_E$}{A function assigning a hyperball to each waypoint in $\WP$ with radius given by $E$} \notsep
\entry{$\hball_{\s}$}{The hyperball assigned to the waypoint $\s$} \notsep
\entry{$\hballs$}{A set of hyperballs. $\hballs(\WP)$ are the hyperballs corresponding to points in $\WP$} \notsep
\entry{$\bath$}{A sequence of hyperballs or \textbf{path}. If $\plan$ is a plan drawn from $\splans[][]{\WP}{*}$, then $\bath_{\plan}$ is the corresponding sequence of hyperballs} \notsep
\entry{$\product{\bath}$}{The bundle of plans ($\eps$-tube) represented by the hyperball sequence $\bath$} \notsep
\entry{$\rK(\bath)$}{The $\eps_{\bath}$-reliability of $\bath$, where $\eps_{\bath}$ is the vector of corresonding hyperball radii. $\bath$ is \textbf{robust} if $\rK(\bath)=1$} \notsep
\entry{$\refine$}{The \textbf{refinement mutation}, replaces a hyperball with several smaller hyperballs in the same area, increasing the local resolution of $\hballs(\WP)$} \notsep
\entry{$\allbaths[][]{}{}$}{The \textbf{set of all paths}. This symbol admits many variations. The set of paths of length $n$ is $\allbaths[][]{}{n}$, of any length, $\allbaths[][]{}{*}$. The set of paths with hyperballs in the set $A$ is $\allbaths[][]{A}{}$, all paths that are robust is $\allbaths[r][]{}{}$. All paths from $\hball_1$ to $\hball_2$ is $\allbaths[][]{}{}(\hball_1, \hball_2)$} \notsep
\entry{$\bath_1 \sim \bath_2$}{Paths are equivalent if they are both robust, start at the same hyperball, and end at the same hyperball. Equivalently, $\bath_1, \bath_2 \in \allbaths[r][]{}{*}(\hball, \hball')$} \notsep
\entry{$\achclass{\hballs}{\hball}{\hball'}$}{Equivalence class of paths from $\hball$ to $\hball'$ with hyperballs in $\hballs$, equal to $\allbaths[r][]{\hballs}{*}(\hball, \hball')$} \notsep
\entry{$\achset{\hballs}{\hballs}$}{The set $\allbaths[r][]{\hballs}{*}$ partitioned by $\sim$, i.e. the set of all equivalence classes of \emph{paths} with hyperballs in $\hballs$} \notsep
\entry{$\achset{\hballs}{\SSK}$}{The set of equivalence classes of \emph{plans} with members inside a path in $\allbaths[r][]{\hballs}{*}$} \notsep
\entry{$\extend$}{The \textbf{extension mutation}, adds new larger hyperballs around an existing hyperball, increasing the area of statespace covered by $\hballs(\WP)$}
\end{notation}

\begin{figure}
\includegraphics[width=\textwidth]{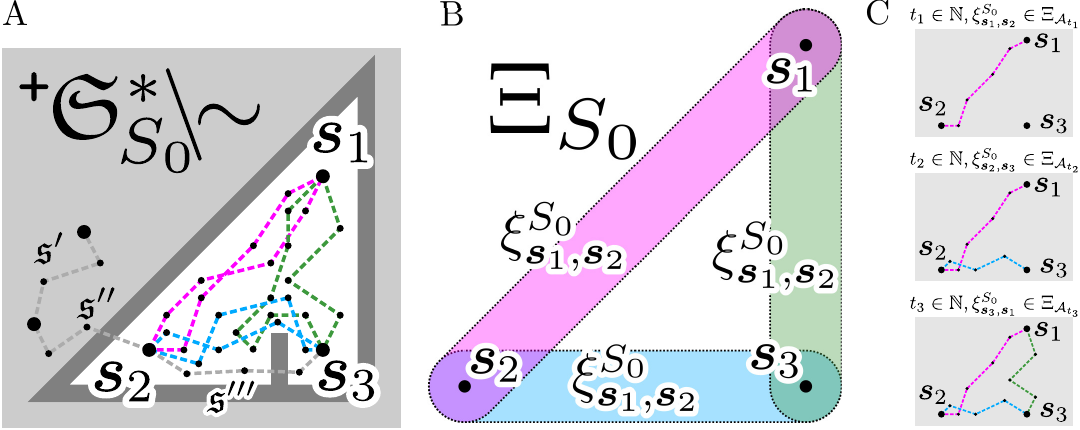}
\caption{\label{fig:behavior_space}The set of all behaviors (plans) through statespace is uncountably infinite, but there is internal structure to this set which can be taken advantage of. (A) The set of interest for our agent is the set of all robust plans of any length with waypoints inside the reachable region of statespace, $\splans[r][]{\SSK}{*} \subset \splans[][]{\SSK}{*} \subset \splans[][]{\SS}{*}$, which excludes plans containing waypoints outside of $\SSK$ (plans $\plan'$ and $\plan''$), and any plan that fails or otherwise isn't robust ($\plan''$ and $\plan'''$). All plans linking the same two points form an equivalence class (plans in the same equivalence class have the same color). The equivalence relation partitions the set of robust plans through $\SSK$, $\splans[r][]{\SSK}{*}\hspace{-0.2em}\backslash{}{\sim}$. (B) This partition is equivalent to the set of solvable problems $\Xi_{\SSK}$ in the reachable statespace, each achievement class $\xi$, or set of equivalent behaviors, corresponds to a solvable problem. (C) An agent is hyperadaptable if, for every solvable problem in $\Xi_{\SSK}$, the agent finds in finite time a solution to that problem.}
\end{figure}

Before, we gave an informal definition of a hyperadaptable agent as one that can always eventually solve any solveable problem. Now, we can now provide a formal definition of \emph{hyperadaptability}:

\begin{box_definition}{Hyperadaptability}{hyperadaptability}
Let $\agent_t$ be an agent at time $t\in\mathbb{N}$.
\begin{equation*}
\agent \text{ is \emph{hyperadaptable}} \iff \forall \achclass{\SSK}{\s}{\s'} \in \achset{}{\SSK}, \exists t\in\mathbb{N} \text{ s.t. } \achclass{\SSK}{\s}{\s'} \in \achset{\agent_t}{\SSK}
\end{equation*}

In other words, an agent is hyperadaptable if and only if for any achievement class, there is some finite time at which that class is part of the agent's capability (Figure \ref{fig:behavior_space}C).
\end{box_definition}

You may notice that this definition makes no mention of any priors (crystal intelligence) or inference ability (fluid intelligence), as indeed, hyperadaptability is independent of these notions. Fluid intelligence might allow a hyperadaptable agent to make more complex inferences about which behaviors could work, prior knowledge might be able to narrow down the set of possible behaviors from the start, but hyperadaptability cannot depend on either of these concepts. In fact, because of this, an agent can only be hyperadaptable if it can perform an \emph{exhaustive search} of $\splans[][]{\SSK}{*}$, as otherwise it would be possible for it to miss a necessary behavior. We immediately face a stark problem: $\splans[][]{\SSK}{*}$ is uncountably infinite, meaning that it cannot be enumerated, meaning that it is mathematically \emph{impossible} to exhaustively search it.

\begin{figure}
\includegraphics[width=\textwidth]{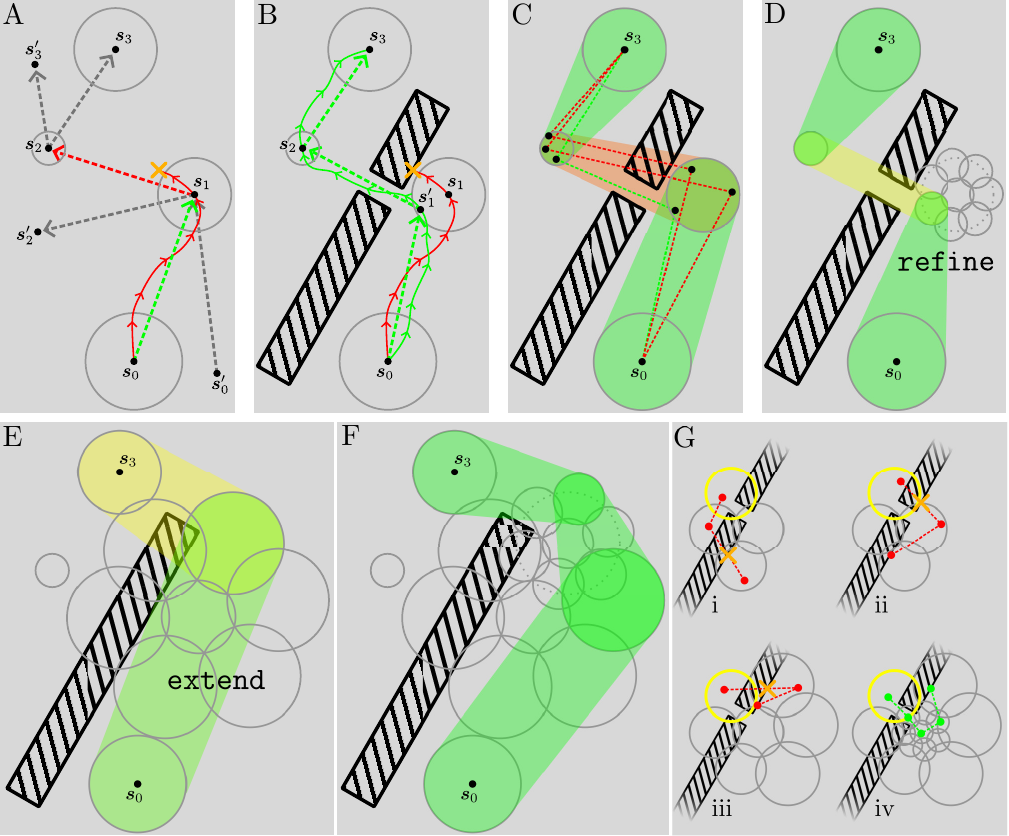}
\caption{\label{fig:decomopsition}Where along a plan the plans fails contains rich information about other plans. (A) If the plan $\plan = (\s_0, \s_1, \s_2, \s_3)$ fails between $\s_1$ and $\s_2$, then we know that any other plan that contains $(\s_1, \s_2)$, such as $(\s'_0, \s_1, \s_2, \s'_3)$, will \emph{also} fail at $(\s_1, \s_2)$. Conversely, any plan containing $(\s_0, \s_1)$ will at least not fail at $(\s_0, \s_1)$, such as $(\s_0, \s_1, \s'_2)$. (B) Just because $(\s_1, \s_2)$ failed, doesn't mean that other plans inside its $\eps$-tube will also fail: it could be that $(\s_1, \s_2)$ was a ``near miss''. (C) If there is a solution in $\plan$'s $\eps$-tube $\product{\bath_{\plan}}$, then randomly sampling plans can eventually find a solution. (D) A better way is to replace the single $\eps$-tube with several higher-resolution $\eps$-tubes by refining one of the waypoint's hyperballs. (E) However, if a valid solution isn't inside the original $\eps$-tube, then the agent can add qualitatively new waypoints through extension. (F) Extension followed by refinement can create arbitrarily complex plans. (G) Ideally, the agent should attempt simpler and coarser plans before more complex and fine-grained plans (i $\rightarrow$ ii $\rightarrow$ iii $\rightarrow$ iv).}
\end{figure}

Thankfully, the situation is not hopeless. First, we can construct a countably infinite set of behaviors that is dense in $\splans[][]{\SSK}{*}$, which \emph{can}, mathematically, be exhaustively searched. Second, even for this countable subset, our agent only needs to find one behavior for each achievement class: after finding one behavior $\plan \in \achclass{\SSK}{\s}{\s'}$, it is not necessary to try any other behaviors in this class. Third, there is significant modular structure in $\splans[][]{\SSK}{*}$ that we can take advantage of: basically, every sub-plan of a plan is shared with infinitely many other plans, a fact that can be used to de-prioritize plans that are unlikely to succeed based on their substructure. Ultimately, the necessity to construct a set which is dense in $\splans[][]{\SSK}{*}$ and the ability to leverage shared subplan structure will provide us with the outlines of a recipe for performing a search across the set of behaviors.

Suppose that our agent has a finite set of waypoints $\WP\subset\SS$ from which it can construct a behavior $\plan$. If this behavior fails, then in the deterministic conditions that we are assuming, the next behavior our agent tries should \emph{not} be $\plan$ again, as it will definitely fail. Suppose that $\plan = (\s_0, \s_1, \s_2, \s_3)$, so that $\plan$ naturally decomposes into three ``sub-behaviors'', $\plan_1 = (\s_0, \s_1)$, $\plan_2 = (\s_1, \s_2)$, and $\plan_3 = (\s_2, \s_3)$ (Figure \ref{fig:decomopsition}A). Suppose that $\rK(\plan_1)=1$ and $\rK(\plan_2)=0$, meaning that $\plan_1$ succeeded but $\plan_2$ failed, and that the agent never got a chance to actually try $\plan_3$. By definition, if any sub-behavior of a behavior fails, the entire behavior fails, and since we are assuming that environment dynamics are deterministic, it follows that:
\begin{enumerate}
\item Any behavior containing $\plan_1$ will at least not fail at $\plan_1$, and
\item Any behavior containing $\plan_2$ will fail at $\plan_2$ or an earlier sub-behavior.
\end{enumerate}
Thus, our agent should \emph{at least} not pick as its next behavior any behavior containing $\plan_2$.

Now technically, for any $\eps>0$, any \emph{other} behavior $\plan'$ in $\plan$'s $\eps$-tube \emph{could} be successful (Figure \ref{fig:decomopsition}B). Essentially, it's possible that $\plan$ was a near-miss, and if the agent looks for long enough in $\tube{\eps}(\plan)$ for a plan, it will eventually find the solution (Figure \ref{fig:decomopsition}C). Suppose that each waypoint $\s\in\WP$ has a fixed $\epsilon$, encoded by a function $E : \s \mapsto \epsilon_{\s}$. Then each waypoint $\s \in \WP$ can be treated as a hyperball, encoded by a function $B_E : \s \mapsto \hball_{\s}$, with $\hball_{\s} = \ball{\s}{\epsilon_{\s}}$. For brevity, we will sometimes substitute $\hball_i$ for $\hball_{\s_i}$. We denote the set of hyberballs corresponding to waypoints in $\WP$ as $\hballs(\WP)$. We extend these functions to sequences, letting $E : \plan \mapsto \eps_{\plan}$, with $\eps_{\plan} = (\epsilon_{\s_0}, \epsilon_{\s_1}, ... \epsilon_{\s_{\n{1}}})$ being the vector of corresponding $\epsilon$'s and $B_E : \plan \mapsto \bath_{\plan}$, with $\bath_{\plan} = (\hball_{\s_0}, \hball_{\s_1}, ... \hball_{\s_{\n{1}}})$ being the sequence of corresponding hyperballs. In the reverse direction, we let $\s(\hball_i) = \s_i$ be the center-point of the hyperball $\hball_i$, and $\epsilon(\hball_i) = \epsilon_{\s_i}$ the radius of $\hball_i$. Extending again to sequences, we let $\eps_{\bath} = (\epsilon_{\s_0}, \epsilon_{\s_1}, ... \epsilon_{\s_{\n{1}}})$ be the vector of corresponding hyperball radii, and $\plan_{\bath} = (\s_0, \s_1, ... \s_{\n{1}})$ the plan of corresponding hyperball centers.

This allows us to re-represent the $\eps$-tube around $\plan$, $\tube{\eps(\plan)}(\plan)$ as:
\begin{equation*}
\product{\bath_{\plan}} = \cartprod_{\mathclap{\hball_i \in \bath_{\plan}}}{\hball_i} = \hball_0 \times \hball_1 \times ... \times \hball_{\n{1}} = \{(\s'_0, \s'_1, ... \s'_{\n{1}}) : \s'_i \in \hball_i\} = \tube{\eps(\plan)}(\plan)
\end{equation*}
Where we define $\product{\bath_{\plan}}$ to be the Cartesian product of the hyperballs in $\bath_{\plan}$. If $\exists \plan^{\star} \in \product{\bath_{\plan}}$ with $\rK(\plan^{\star})=1$, then the agent could randomly sample plans from $\product{\bath_{\plan}}$ until $\plan^{\star}$ is found, but that would be haphazard, and how would the agent repeat $\plan^{\star}$? Instead, the agent can take a more systematic approach, taking some or all of its hyperballs and \emph{refining} them, that is, adding \emph{new} smaller hyperballs:
\begin{box_definition}{Refinement}{refinement}
The refinement function $\refine$ takes in two arguments, $\hball$ a hyperball and $a\in(0, 1)$ a scaling parameter, and returns a set of new ``denser'' hyperballs:
\begin{equation*}
\refine(\hball, a) = \{\hball^{(1)}, \hball^{(2)}, ... \hball^{(k)}\}
\end{equation*}
with the following properties:
\begin{enumerate}
\item $\s(\hball^{(1)}) = \s(\hball)$
\item $\forall \hball' \in \refine(\hball, a), \epsilon(\hball') = a \cdot \epsilon(\hball)$
\item $\hball \subset \hball^{(1)} \cup \hball^{(2)} \cup ... \cup \hball^{(k)}$
\end{enumerate}
\end{box_definition}

We can use the $\refine$ function to replace a hyperball in $\bath_{\plan}$ with several smaller, ``more precise'' options in the same area, so that the agent has more control over the behaviors it executes (Figure \ref{fig:decomopsition}D). If the original tube was $\bath_{\plan} = (\hball_0, \hball_1, \hball_2, \hball_3)$, and the agent refined $\hball_2$, then the agent would have a new set of tubes $\baths = \{(\hball_0, \hball_1, \hball'_2, \hball_3) : \hball'_2 \in \refine(\hball_2, a)\}$, with the property that $\forall \plan' \in \product{\bath_{\plan}}, \exists \bath' \in \baths$ s.t. $\plan' \in \product{\bath'}$. Recalling the definition of \emph{$\eps$-reliability} (Equation \ref{eq:eps-reliability}), we can denote the reliability of a sequence of hyberballs (or a \emph{path}) as $\rK(\bath) = \rKe{\eps_{\bath}}(\plan_{\bath})$. Thus, when sampling plans from a path, a path corresponds to a behavior in the third sense.

If $\hballs$ is a set of hyperballs, and $\allbaths[][]{\hballs}{*}$ the set of all paths of any length using hyperballs in $\hballs$, then $\allbaths[r][]{\hballs}{*} = \{\bath \in \allbaths[][]{\hballs}{*} : \rK(\bath) = 1\}$ is the set of all robust paths in $\allbaths[][]{\hballs}{*}$. If two paths $\bath_1, \bath_2 \in \allbaths[r][]{\hballs}{*}$ start at $\hball$ and end at $\goal{\hball}$, then $\bath_1 \sim \bath_2$, so we say they are both in the equivalence class $\achclass{\hballs}{\hball}{\goal{\hball}}$. This is an equivalence class of \emph{paths}. We let $\achset{\hballs}{\hballs} = \{\achclass{\hballs}{\hball}{\goal{\hball}} : \hball, \goal{\hball} \in \hballs\}$ be the set of all equivalence classes of paths. Then, $\achset{\hballs}{\SSK} = \{\achclass{\SSK}{\s}{\goal{\s}} \in \achset{}{\SSK} : \exists \achclass{\hballs}{\hball}{\goal{\hball}} \in \achset{\hballs}{\hballs} \text{ s.t. } \s \in \hball \text{ and } \goal{\s} \in \goal{\hball}\}$ represents the total set of problems that can be solved using $\hballs$. Suppose that there is a path $\bath \in \allbaths[][]{\hballs}{*}$ for which $\plan_{\bath}$ has failed, with $\plan_{\bath}\idx{0} = \s_0$ and $\plan_{\bath}\idx{\n{1}} = \goal{\s}$, but $\exists \plan^{\star}\in\product{\bath_{\plan}}$ such that $\rKe{\eps}(\plan^{\star})=1$. If $\hballs$ is repeatedly refined, producing an infinite sequence $\hballs_1$, $\hballs_2$, ... such that eventually every hyperball is refined, then it is guaranteed that $\exists n\in\mathbb{N}$ such that $\achclass{\SSK}{\s_0}{\goal{\s}} \in \achset{\hballs_n}{\SSK}$, which is to say, if there \emph{is} a robust plan in $\product{\bath}$, our agent is sure to find it. However, if there \emph{isn't}, our agent could spend forever trying various minute variations on $\plan_{\bath}$. If there is no solution inside of $\product{\bath}$, then the agent will waste an infinite amount of time. Imagine our bushwhacking urbanite coming across a steep cliff, occluded by foliage to either side. They could spend an infinite amount of time attempting different variations on climbing the cliff, without ever checking for a shallower slope past the foliage.

Alternatively, there are an infinite number of behaviors outside of $\product{\bath}$, in $\splans[][]{\SSK}{} - \product{\bath}$. If there is no solution inside of $\product{\bath}$, then some necessary waypoints for a solution lay outside of $\product{\bath}$, and the agent has no hope of sampling those plans, that is, \emph{unless} it can add hyperballs that cover new regions of statespace, that is, \emph{extending} the scope of its behaviors:

\begin{box_definition}{Extension}{extension}
The extension function $\extend$ takes in three arguments, a hyperball $\hball$ and two scaling parameters $a>1$ and $c>1$, and returns a new set of hyperballs covering ``new'' regions of $\SS$:
\begin{equation*}
\extend(\hball, a, c) = \{\hball^{(1)}, \hball^{(2)}, ... \hball^{(k)}\}
\end{equation*}
with the following properties:
\begin{enumerate}
\item $\forall \hball' \in \extend(\hball, a, c), \epsilon(\hball') = a \cdot \epsilon(\hball)$
\item $\ball{\s(\hball)}{c\cdot\epsilon(\hball)} \subset \hball^{(1)} \cup \hball^{(2)} \cup ... \cup \hball^{(k)}$
\end{enumerate}
\end{box_definition}

We can use the $\extend$ function to add qualitatively new options for our agent: if a waypoint is needed for a solution and isn't currently inside of one of our agent's hyperballs, then repeated extension can expand the scope of the agent's behaviors (Figure \ref{fig:decomopsition}E). This certainly doesn't guarantee that once the waypoint is covered by a hyperball that it can be incorporated into a robust behavior, and if our agent only ever extends, it will almost certainly miss important behaviors... but repeated refinement can fix this (Figure \ref{fig:decomopsition}F). In fact, these two operations give us the tools to deliver on our earlier promise of a countably infinite set of plans dense in $\splans[][]{\SSK}{*}$.

Consider the set of hyperballs $\hballs$, with $\WP(\hballs)$ the set of center-points of those hyperballs. When we refine a hyperball $\hball \in \hballs$, we essentially ``update'' or ``mutate'' $\hballs$, as $\hballs_{i+1} = (\hballs_{i} - \{\hball\}) + \refine(\hball, a)$. Similarly, when we extend a hyperball $\hball \in \hballs$, we are also mutating $\hballs$, as $\hballs_{i+1} = \hballs_{i} \cup \extend(\hball, a, c)$. Repeated application of extension and refinement will produce an infinitely long but countable sequence of sets of hyperballs, $\hballs_0, \hballs_1, \hballs_2, ... $, where $\forall i$, $\abs{\hballs_i}$ is finite. If, in a certain sense, the sequence of mutations is \emph{balanced}, the set $\WP(\hballs_i)$ will be \emph{dense} in the infinite limit:

\begin{box_definition}{Balanced Hyperball Sequence}{balanced-hyperball-sequence}
Let $\hballs_0, \hballs_1, \hballs_2, ... $ be an infinite sequence of sets of hyperballs, where $\hballs_{i+1}$ is produced through either refinement or extension of $\hballs_{i}$. If, $\forall i\in\mathbb{N}$, and $\forall\hball\in\hballs_i, \exists m, n\in\mathbb{N}$ such that $\hball$ is \emph{extended} in $\hballs_{i+m}$ and \emph{refined} in $\hballs_{i+m+n}$, then $\forall \s \in \SSK, \forall \epsilon>0, \exists j\in\mathbb{N}$ such that $\exists \s' \in \WP(\hballs_i)$ with $\norm{\s - \s'} < \epsilon$.
\end{box_definition}

As $\forall i \in \mathbb{N}$, $\WP(\hballs_i)$ is countable, and in the limit dense in $\SSK$, it further follows that the set of all behaviors which can be generated by this set, $\splans[][]{\hballs_i}{*}$, is also countable and, in the limit, dense in $\splans[][]{\SSK}{*}$, from which it immediately follows that our agent can perform an exhaustive search over this set of behaviors, guaranteeing that any robust behavior that \emph{needs} to be found \emph{can} be found. First, unlike in most reinforcement learning, we are assuming that there is no external reset of the environment, meaning that just like in real life, our agent cannot ``teleport''. This provides a severe constraint on which behaviors can even be tried in the first place, that is, every plan has to start where the agent currently is. Additionally, the agent really only needs to actually enumerate $\achset{\hballs_i}{\hballs_i}$, so there are some tricks we can use to dramatically reduce the scope of the search. Basically, if a subpath failed, it might be a good idea to defer trying \emph{other} paths that use a similar subpath, until other options have been exhausted (Figure \ref{fig:decomopsition}G). The question is, how do we actually organize such a search in practical terms?

\section{Segraphs}
\begin{overview}
In order to organize the enumeration of plans and addition of new waypoints, we use a graph whose vertices are hyperballs that segment statespace, a \emph{segraph}. The segraph's edges track the empirical reliability of the agent's low-level controller for traversing between hyperballs. By recording successful and failed traversals over edges, the agent is able to both avoid unreliable edges and select novel paths to drive plan enumeration while simultaneously regulating segraph mutation. This creates a tight feedback loop: path enumeration supplies experiences that guide graph mutation, and each mutation expands the future path-set, allowing the agent to adaptively search the space of achievement classes without repeating failed sub-behaviors. Because any mutated segraph contains its parent, disconnected subgraphs that once blocked enumeration are eventually contained by better-connected segraphs, ensuring that local exploration results in global problem-solving.
\end{overview}

\begin{notation}
\entry{$\segraph=(\segverts, \segedges)$}{A \textbf{segraph}, a graph with directed edges $\segedges$ that connect vertices $\segverts$ that segment statespace, which is used to organize the enumeration of plans} \notsep
\entry{$\vertex_i$}{A segraph \textbf{vertex} $\vertex_i$ representing a ``chunk'' of state-space, a corresponding hyperball $\hball_i \subset \SS$} \notsep
\entry{$\edge_{i,j}$}{The directed segraph \textbf{edge} between vertices $\vertex_i$ and $\vertex_j$, representing a chunk of connection space $\connection_{i, j} = \hball_i \times \hball_j \subset \SS^2$} \notsep
\entry{$\measure$}{The \textbf{measure} of a vertex or edge, $\measure(\vertex_i) = \abs{\hball_i}$, $\measure(\edge_{i, j}) = \abs{\connection_{i,j}}$} \notsep
\entry{$\rK(\edge_{i,j})$}{The theoretical \textbf{reliability} of the edge $\edge_{i, j}$, equal to $\rK((\hball_i, \hball_j))$} \notsep
\entry{$n_s(\edge)$}{The number of \textbf{successful traversals} by $\agent$ over $\edge$} \notsep
\entry{$n_f(\edge)$}{The number of \textbf{failed traversals} by $\agent$ over $\edge$} \notsep
\entry{$n(\edge)$}{The total number of \textbf{traversals} by $\agent$ over $\edge$, equal to $n_s(\edge) + n_f(\edge)$} \notsep
\entry{$\erK(\edge)$}{The empirically \textbf{estimated reliability} of $\edge$} \notsep
\entry{$\p$}{A \textbf{path}, a sequence of unique vertices} \notsep
\entry{$\allpaths[][]{}{}$}{The \textbf{set of all paths}. This symbol admits many variations. The set of paths of length $n$ is $\allpaths[][]{}{n}$, of any length, $\allpaths[][]{}{*}$. The set of paths with vertices in the set $A$ is $\allpaths[][]{A}{}$. The set of robust paths is $\allpaths[r][]{}{}$. The set of paths that start at $\vertex$ and end at $\vertex'$ is $\allpaths[][]{}{}(\vertex, \vertex')$} \notsep
\entry{$\goalPsym$}{The probability distribution over goals used to represent the way an agent selects problems to try to solve} \notsep
\entry{$\pathPsym$}{The probability distribution over paths used to represent the way an agent selects plans to try to solve a problem} \notsep
\entry{$\rK(\p)$}{The reliability of a path, equal to the reliability of the corresponding hyperball sequence. When $\rK(\p)=1$, the path is \textbf{robust}} \notsep
\entry{$\achclass{\segraph}{\vertex}{\vertex'}$}{The equivalence class of robust paths over $\segraph$ between $\vertex$ and $\vertex'$} \notsep
\entry{$\achset{\segraph}{\segverts}$}{The set of all equivalence classes of robust paths over $\segraph$} \notsep
\entry{$\try(\p)$}{A function modeling the action of an agent \textbf{trying to follow a path}} \notsep
\entry{$\efailed{\segedges}$}{The set of edges that fail when an agent tries to follow a path. If $\efailed{\segedges}=\varnothing$, then the path succeeds} \notsep
\entry{$\eforcedinactive{\segedges}$}{The set of edges that are inhibited (excluded from pathfinding by the agent) due to having failed while trying paths} \notsep
\entry{$\eactive{\segedges}$}{The set of edges that are not inhibited and can be used in pathfinding}
\end{notation}

If a transition between two hyperballs $\hball_1$ and $\hball_2$ along a path has failed, then to de-prioritize paths that use that transition, our agent needs some way to record the fact of the failure. Indeed, if the transition succeeded, this would be evidence that other paths using that transition might also succeed, and our agent would want to record the fact of the success as well. This information can easily be represented by a \emph{graph}, where each \emph{vertex} is identified with a hyperball, and each directed \emph{edge} from one vertex to another encodes information about the reliability of transitioning between the corresponding hyperballs. As the vertices of this graph effectively segment statespace, we call it a \emph{segraph}:

\begin{box_definition}{Segraph}{segraph}
A \emph{segraph} is a pair $\segraph = (\segverts, \segedges)$ with:
\begin{itemize}
\item $\segverts$ - a finite set of vertices; each vertex $\vertex_i \in \segverts$ is a tuple $(i, \hball_i)$ where $i\in\mathbb{N}$ is an identifier and $\hball_i$ is the corresponding hyperball, called the vertex's \emph{field} in statespace. The measure (size) of this field is denoted $\measure_i = \measure(\vertex_i) = \abs{\hball_i}$.
\item $\segedges$ - a finite set of directed edges; each edge $\edge \in \segedges$ is an ordered tuple $(\vertex_i, \vertex_j)$ representing a connection from $\vertex_i$ to $\vertex_j$. We say that $\connection_{i, j} = \hball_i \times \hball_j \subset \SS^2$ is the edge's \emph{field} in connection-space. The measure (size) of this field is denoted $\measure_{i,j} = \measure(\edge_{i,j}) = \abs{\connection_{i, j}}$.
\end{itemize}
We let $\SS(\segraph) = \bigcup_{\vertex_i\in \segverts}\hball_i$ be the \emph{coverage} of $\segraph$.
\end{box_definition}

The definitions of $\refine$ and $\extend$ can be seamlessly extended to segraph vertices, with the only additional machinery being that for pragmatic reasons, the segraph enforces a redundancy constraint on added vertices, which is to say, if there is an existing vertex (hyperball) in the segraph that is very similar to a new vertex that \emph{would} be added by refinement or extension, then that new vertex is actually unnecessary and is not added. Additionally, we do not assume that $\segraph$ is fully-connected, so to ensure that every possible plan can actually be generated, we also use a linking function $\link$ that creates an edge between two vertices.

The segraph has two deeply intertwined jobs: the first is to help the agent navigate to distant goals by providing robust \emph{paths} (sequences of vertices), the second is to regulate its own mutation to produce a balanced sequence of segraphs, so that eventually the agent can generate any behavior necessary to solve any problem. We accomplish this by establishing an extremely tight bidirectional feedback loop between how the segraph is \emph{used} by the agent and how the segraph gets \emph{mutated}. Over-time, the segraph self-organizes in response to the needs of the agent, in much the same way that bone, muscle, or nervous tissue respond to usage by an organism.

The feedback loop between \emph{usage} of the segraph and \emph{mutation} of the segraph is absolutely critical: as \emph{using} the segraph just means following a path sampled from it (selecting a goal and finding a path to it), repeated \emph{use} of a given segraph is (or should be) equivalent to \emph{enumerating paths} over the segraph. Likewise, \emph{mutating} the graph just means creating a modified segraph, repeated \emph{mutation} creates a sequence of segraphs; in other words, the enumeration of segraphs. The enumeration of segraphs shapes the set of paths that can be enumerated at each stage, and the enumeration of paths collects information that guides or biases the enumeration of segraphs.

\begin{figure}
\includegraphics[width=\textwidth]{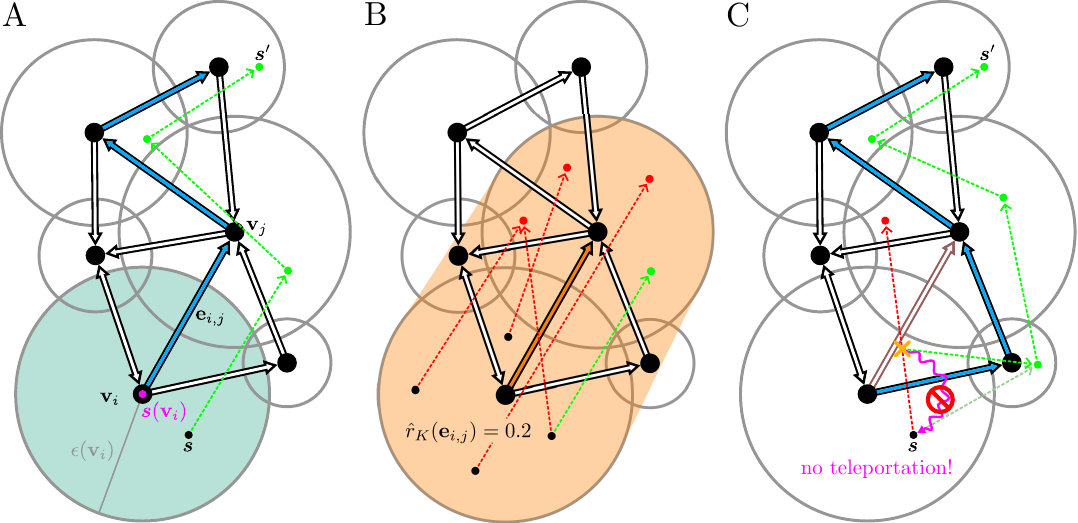}
\caption{\label{fig:segraph}(A) A \emph{segraph} is a directed graph whose vertices segment statespace. Each vertex's segment of statespace, or \emph{field}, is a hyperball centered at $\s(\vertex_i)$ with radius $\epsilon(\vertex_i)$. An edge represents a connection between two regions of statespace. In order to help the agent reach a distant goal such as $\s'$ the segraph will find a path (sequence of vertices highlighted in blue) between the vertex the agent is currently in and a vertex containing the goal. Then, the agent will sample waypoints from the fields of the vertices along the path, generating a plan from it's current state to the target state (green dots connected by dashed line). (B) Experience trying to cross the edges of the segraph allows the segraph to estimate the reliability of each of it's edges. (C) Unreliable edges (such as the one marked in orange), can be \emph{inhibited}, prevented from use in pathfinding, which can allow for potentially better plans to be tried (example alternative path highlighted in orange). The new plan that is generated has to start where the last (failed) plan left the agent, as there are no external resets.}
\end{figure}

For either the usage or mutation of the segraph to be \emph{adaptive}, our agent has to be able to remember information about its experiences while using the segraph. Ultimately, both mutation and usage will hinge on the agent having an estimate of the reliability of each edge of the segraph, which it accomplishes by storing the number of failed and successful transitions over each edge $\edge_{i,j}$ to empirically estimate the true reliability $\rK(\edge_{i,j}) = \rK((\hball_i, \hball_j))$ as:
\begin{equation*}
\erK(\edge_{i,j}) = \tfrac{n_s(\edge_{i,j}) + \tilde{n}_s}{n_s(\edge_{i,j}) + \tilde{n}_s + n_f(\edge_{i,j}) + \tilde{n}_f} = \tfrac{n_s(\edge_{i,j}) + \tilde{n}_s}{n(\edge_{i,j}) + \tilde{n}_s + \tilde{n}_f}
\end{equation*}
where $n_s(\edge_{i,j})$ and $n_f(\edge_{i,j})$ are simply the empirical counts of successful and failed traversals by $\K$ over the edge $\edge_{i,j}$, with $\tilde{n}_s$ and $\tilde{n}_f$ pseudo-counts encoding the agent's initial guess about $\edge_{i,j}$'s reliability, and $n(\edge) = n_s(\edge) + n_f(\edge)$. These traversal counts will ultimately control how the segraph gets mutated (segraph enumeration), as well as how goals and paths are selected (path enumeration).

\subsection{Enumerating Paths}
As each plan must start where the agent currently is, the main degrees of freedom available to our agent while using the segraph are which \emph{targets} it tries to reach and which \emph{paths} it chooses to take to those targets. We model the agent's \emph{selection of goals} with the distribution $\goalPsym$ over $\segverts$. A path $\p = (\vertex_0, \vertex_1, ... \goal{\vertex})$ from $\vertex_0$ to $\goal{\vertex}$ is just a sequence of vertices, with no vertex repeated. As the agent is \emph{at} a specific state, it has to choose a starting vertex whose hyperball contains it's state. Let $\allpaths[][]{\segraph}{*} = \allpaths[][]{(\segverts, \segedges)}{*}$ be the set of all paths over the segraph $\segraph$. We model the agent's \emph{selection of a path} with the distribution $\pathP{(\segverts, \segedges)}{\vertex}{\goal{\vertex}}$ over $\allpaths[][]{(\segverts, \segedges)}{*}$. In a way, $\allpaths[][]{\segraph}{*}$ acts like a discretized version of $\splans[][]{\SS(\segraph)}{*} \subset \splans[][]{\SS}{*}$, which in the infinite limit it approaches. As a finite set, we can choose any order for enumerating it, but efficient problem solving requires an \emph{intelligent} order.

The first thing to realize is that, just as our agent needs to exhaust $\achset{}{\SSK}$ rather than $\splans[r][]{\SSK}{*}$, for every $\vertex, \vertex' \in \segverts$ it is sufficient for our agent to find a \emph{single} path $\p$ between them with $\rK(\p)=1$. Let $\allpathsf[][]{\segraph}{*}{\vertex, \vertex'} = \{\p \in \allpaths[][]{\segraph}{*} : \p\idx{0}=\vertex, \p\idx{\n{1}}=\vertex'\}$ be the set of paths connecting $\vertex$ to $\vertex'$, and $\allpathsf[r][]{\segraph}{*}{\vertex, \vertex'} = \{\p \in \allpathsf[][]{\segraph}{*}{\vertex, \vertex'} : \rK(\p)=1\}$ the set of robust paths from $\vertex$ to $\vertex'$ over $\segraph$. You may recall the construction of plan-equivalence from section \ref{sec:problems_behaviors_solutions}: here, two \emph{paths} $\p_i, \p_j$ are equivalent if $\p_i, \p_j \in \allpathsf[r][]{\segraph}{*}{\vertex, \vertex'}$, we denote the equivalence class as $\achclass{\segraph}{\vertex}{\vertex'}$, and the set of all such classes as $\achset{\segraph}{\segverts} = \{\achclass{\segraph}{\vertex}{\vertex'} : \vertex, \vertex' \in \segverts\}$. It is this set that must actually be exhaustively searched, and thus \emph{enumerated}. Naturally, our agent doesn't know a-priori which paths are robust (if any are), and so must perform a (non-exhaustive) search over $\allpaths[][]{\segraph}{*}$. Ultimately, this enumeration will be performed via the \emph{inhibition} of different sets of edges.

\subsection{Try and Try Again: Failure-Driven Enumeration} \label{sec:failure_order}
One of the basic insights of this paper is that, after our agent has tried a behavior and failed, the most basic way that our agent can adapt is to simply not repeat the failed behavior (or failed sub-behavior). As the novelist Rita May Brown said (misattributed to Albert Einstein), ``Insanity is doing the same thing over and over and expecting different results.'' In this sense, the necessarily gradual weight updates of deep networks make them nearly insane. By never repeating failed behavior, our agent is in some sense guaranteed to eventually find a successful behavior. We accomplish this by simply \emph{inhibiting} any edge that the agent fails to traverse. Imagine that our agent started at the vertex $\vertex$, tried to reach the goal $\goal{\vertex}$, and tried to follow the path $\p$. We can express this with the function:
\begin{equation}
\try(\p) \mapsto \vertex', \segedges^{\times}
\end{equation}
where $\vertex'$ is the final state of the agent after trying to follow $\p$, and $\efailed{\segedges}$ is the set of edges (if any) that failed while following the path. If the agent succeeded, then $\vertex' = \goal{\vertex}$ and $\efailed{\segedges} = \varnothing$, otherwise, $\vertex' \neq \goal{\vertex}$ and $\segedges^{\times} \neq \varnothing$. Recall that the goal at this stage for our agent is to enumerate $\achset{\segraph}{\segverts}$, \emph{not} $\allpaths[][]{\segraph}{*}$. If $\p$ failed, then we know that $\rK(\p) < 1$, meaning that $\p$ cannot be a representative of $\achclass{\segraph}{\vertex}{\goal{\vertex}}$. Indeed, \emph{no} path $\p'$ containing an edge in $\efailed{\segedges}$ can be a member of \emph{any} class in $\achset{\segraph}{\segverts}$, so all such paths can be \emph{safely excluded} from our agent's enumeration.

To do this, we simply preclude the use of failed edges in future paths. If $\segraph = (\segverts, \segedges)$, we accomplish this by splitting $\segedges$ into two disjoint sets, $\eforcedinactive{\segedges}$ and $\eactive{\segedges}$, where $\eforcedinactive{\segedges}$ are \emph{inhibited}, and $\eactive{\segedges}$ are \emph{disinhibited}. Initially, we let $\eactive{\segedges} = \segedges$ and $\eforcedinactive{\segedges} = \varnothing$, then after every call of $\try$, we perform the updates: $\eforcedinactive{\segedges} \leftarrow \eforcedinactive{\segedges} \cup \efailed{\segedges}$, $\eactive{\segedges} \leftarrow \eactive{\segedges} - \efailed{\segedges}$. Basically, the agent \emph{inhibits} every edge that failed. Instead of sampling paths from $\pathP{(\segverts, \segedges)}{\vertex}{\goal{\vertex}}$, our agent samples only from $\pathP{(\segverts, \eactive{\segedges})}{\vertex}{\goal{\vertex}}$, the set of paths over $\segraph$ with only disinhibited edges. Because of this, the agent is inherently \emph{resilient} to failure, as it merely tries again to reach the goal, now from a new location, and with some options trimmed away.

\subsection{The Curse of Locality}

\newcommand{\acg}[2]{\achclass{\segraph}{\vertex_{#1}}{\vertex_{#2}}}

\begin{figure}
\includegraphics[width=\textwidth]{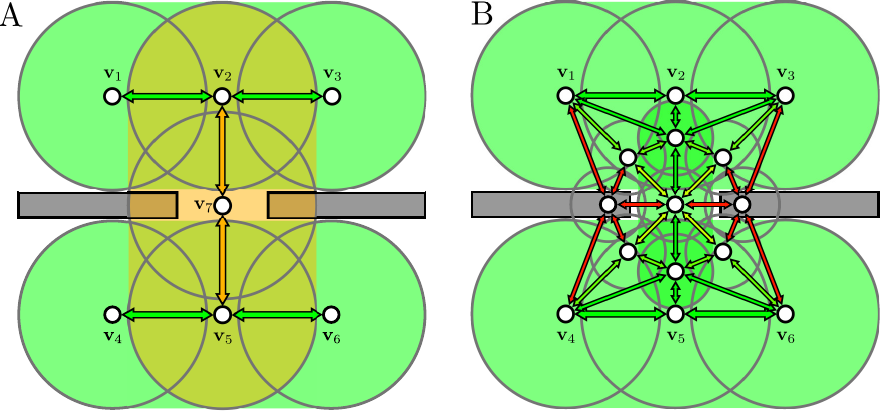}
\caption{\label{fig:affordance} (A) For the shown segraph $\segraph$, the elements of $\achset{\segraph}{\segverts}$ are not connected to each other, meaning there is no guarantee that our agent can actually enumerate $\achset{\segraph}{\segverts}$, complicating the agent's task of trying every behavior. (B) However, through systematic refinement, it can be possible to construct a new segraph $\segraph'$ which connects the elements of $\segraph$, allowing the agent to enumerate $\achset{\segraph}{\segverts}$, just using a \emph{different} segraph.}
\end{figure}

There is a problem with the failure-driven enumeration of paths, caused by inhibition. Consider the segraph $\segraph$ shown in Figure \ref{fig:affordance}A. All the edges among $\segverts' = \{\vertex_1, \vertex_2, \vertex_3\}$ are robust, as are all the edges among $\segverts'' = \{\vertex_4, \vertex_5, \vertex_6\}$, whereas the edges among $\{\vertex_2, \vertex_7, \vertex_5\}$ are not robust. Due to the structure of the segraph, the capability set of the segraph can be split into two halves, $\achset{\segraph}{\segverts} = \{\acg{i}{j} : \vertex_i, \vertex_j \in \segverts'\} \cup \{\acg{i}{j} : \vertex_i, \vertex_j \in \segverts''\}$. There is no achievement class connecting a vertex in $\segverts'$ to $\segverts''$, or visa versa! That means that if the agent tries to follow a path from, say, $\vertex_1$ to $\vertex_4$, it will likely fail at edge $\edge_{2, 7}$ or $\edge_{7, 5}$, which will cause that edge to be inhibited, thus \emph{disconnecting} $\segraph$, making it impossible to enumerate $\achset{\segraph}{\segverts}$.

This is not as catastrophic as it first appears. Recall that a path over a segraph is a sequence of hyperballs, which itself corresponds to a specific $\eps$-tube, a bundle of \emph{plans}. For a given graph, let $\splans[][]{\segraph}{*} = \{\plan \in \p : \p \in \allpaths[][]{\segraph}{*}\}$ be the set of all plans that can be sampled from any path over $\segraph$. We say that one segraph $\segraph'$ \emph{contains} another segraph, $\segraph$, if and only if $\splans[][]{\segraph}{*} \subset \splans[][]{\segraph'}{*}$, which we write as $\segraph' \contains \segraph$. What is interesting is that, for any mutation operator $\texttt{mutation} \in \{\extend, \refine, \link\}$, $\texttt{mutation}(\segraph) \contains \segraph$. A natural consequence of this is that, if $\segraph'$ is a mutation of $\segraph$, that $\achset{\segraph}{\SSK} \subset \achset{\segraph'}{\SSK}$. As these relationships are transitive, for any sequence of mutations leading from $\segraph_i$ to $\segraph_{i+k}$, it will be the case that $\achset{\segraph_i}{\SSK} \subset \achset{\segraph_{i+k}}{\SSK}$. The key takeaway here is that, even if our agent cannot enumerate $\achset{\segraph_{i}}{\segverts}$ because it is disconnected, it could be the case at a later time that it can still find every element of $\achset{\segraph_{i}}{\segverts}$, just as a subset of an appropriately mutated segraph with achievement set $\achset{\segraph_{i+k}}{\SSK}$ (Figure \ref{fig:affordance}B).

What this speaks to is the necessity that mutation occur \emph{where the agent needs it}, because in some sense, to reach the broader graph, the agent must first be able to negotiate the ``local'' graph immediately around it, necessitating a sort of nested enumeration of sub-graphs, as-needed. Because mutation will ultimately happen, when it happens, wherever the agent currently is, it must be the case that in order to achieve a balanced sequence of segraphs (produced through mutation), it will be necessary to carefully control \emph{where} the agent is \emph{on} the segraph. In other words, while mutation of the graph will ultimately control which paths are available for the agent to explore, the primary way that we can \emph{control} mutation is by \emph{controlling} which paths are followed... the enumeration of plans loops back on itself.

\section{Segraph Self-Organization}
\begin{overview}
A self-organizing segraph regulates its own development by coupling goal-oriented usage and stress-induced mutation together. A per-edge stress score signals where refinement (splitting a hyperball to raise statespace resolution) can increase the segraph's utility, while a dynamic affordance threshold forces pathfinding to bias the agent towards traversing edges that are either highly reliable or likely to benefit from refinement. Traversal failures actively inhibit edges and raise the threshold; when no paths are left, inhibition is reset and the threshold is lowered, completing a homeostatic loop that re-opens blocked paths as-needed. Every successfully visited vertex is extended before it can be refinement, and overlapping or co-traversed vertices are linked to increase connectivity. Exploration favors edges with high epistemic value (large, under-sampled regions), and paths are tried by simplicity (shortest first), so enumeration advances systematically as failures prune the search space. These coupled mechanisms ensure that over an unbounded sequence of segraphs, coverage becomes dense and the agent ultimately acquires a robust plan for every solvable problems.
\end{overview}

\begin{notation}
\entry{$\potential(\p)$}{The total affordance provided by a path $\p$} \notsep
\entry{$\q(\edge)$}{The utility of an edge $\edge$ in terms of the incremental affordance it provides} \notsep
\entry{$\q_{\text{ref}}(\edge)$}{The total utility of the edges produced by refining $\edge$} \notsep
\entry{$\Delta q(\edge)$}{The theoretical change in utility resulting from refining $\edge$} \notsep
\entry{$\stress(\edge)$}{An empirical estimate of $\Delta q(\edge)$} \notsep
\entry{$\potentialbar$}{An affordance-threshold used to regulate the order in which paths are tried}
\entry{$\potential(\edge)$}{The affordance of the edge $\edge$, $\potential(\edge) = \rK(\edge) \cdot \measure(\edge)$} \notsep
\entry{$\epotential(\edge)$}{The estimated affordance of the edge $\edge$, $\epotential(\edge) = \erK(\edge) \cdot \measure(\edge)$} \notsep
\entry{$\segElt{\potentialbar}$}{The set of edges that are below the affordance threshold $\potentialbar$} \notsep
\entry{$\segEgte{\potentialbar}$}{The set of edges that are above the affordance threshold $\potentialbar$} \notsep 
\entry{$\allpathsf[][]{\raisemath{0.1ex}{\segraph}}{}{\vertex, \edge}$}{The set of all paths over $\segraph$ that start at $\vertex$ and end by crossing $\edge$} \notsep
\entry{$\pathP{\segraph}{\vertex}{\edge}$}{The probability distribution of paths from $\vertex$ to $\edge$ over $\segraph$} \notsep
\entry{$\epistemicscore(\edge)$}{The epistemic value of an edge $\edge$, used to weigh the probability that $\edge$ is chosen as a goal, modeled by the distribution $\goalP$ over edges}
\end{notation}

In the following section, we lay-out \emph{one} possible way of arranging the self-organized usage and mutation of a segraph: there are surely other ways, and our particular method should only be considered as illustrative, rather than definitive. The development of these self-organizing mechanisms was guided primarily by intuition and experimentation, so a broader theory of different types of ordering is required in the future. We begin by laying out the conditions under which segraph refinement is adaptive, rather than detrimental or redundant, and end by describing some practical ways that paths can be sensibly ordered over a given segraph. To understand the way that usage and mutation interact to produce an infinite sequence of segraphs which, in the limit, allow the agent to generate a behavior to solve any problem, we have to reframe the issue: what is the agent trying to \emph{maximize}? Because we want our agent to be able to eventually solve any problem (be able to go to any point from any point), we are arguably trying to maximize $\abs{\achset{\agent}{\SSK}}$, the measure of the set of points in $\SSK$ the agent can go to and from.

Technically, the ability of the agent to go from one point $\s$ to another $\s'$ using its controller $\K$ corresponds to an \emph{affordance} \cite{gibson2014theory} of the statespace. In this sense, $\conK(\s)$ represents the total affordance of the agent while in the state $\s$, but as previously discussed, $\abs{\conK(\s)} \ll \abs{\SSK}$, meaning that most points in $\SSK$ cannot be reached by $\K$ from any given point. This, of course, motivated the introduction of plans, sequences of waypoints that, in effect, extend the capability of the agent by allowing it to reach more of statespace via \emph{intermediary} steps. While we don't treat noise in this paper, successful plans that aren't robust are practically unrealizable, so we restrict our attention to plans that have a non-zero-measure ``tube'' of successful plans around them. This gives our agent some ``wiggle room''. Organizing information about the set of such plans given a finite set of waypoints motivated the introduction of segraphs.

A segraph, along with a probability distribution over paths, provides a means of selecting a path to a goal. In general, a path $\p$ can be treated as a \emph{bundle} of plans, with $\product{\p} = \cartprod_{\vertex \in \p}{\hball(\vertex)}$ representing this bundle (an $\eps$-tube). The path is not a valuable object in it's own right: it only matters as a means of getting from points inside of $\p\idx{0}$ to points inside of $\p\idx{\n{1}}$. If $\rK(\p)$ is the reliability of $\p$, then we can express the affordance of $\p$ as $\potential(\p) = \rK(\p) \cdot \measure(\p\idx{0}) \cdot \measure(\p\idx{\n{1}})$. From the perspective of the agent, a path is only as useful as it is accessible: a path that is never sampled has only virtual affordance. So, the affordance of the agent is $\potential(\agent) = \sum_{\smash{\p\in\allpaths[][]{\segraph}{}}}\pathPsym(\p)\potential(\p)$. How is $\potential(\agent)$ effected by mutation of $\segraph$?

\subsection{Refinement}
While ``refinement'' is an operation that occurs to an individual \emph{vertex}, the relevant information about when to refine a vertex actually exists on an \emph{edge}, as we are ultimately interested in increasing the resolution of the corresponding $\eps$-tubes. Henceforth, when we refer to ``refinement'', we will be talking about an operation that occurs \emph{on an edge}, with the decision to ``refine an edge'' resulting in the refinement of \emph{one of its vertices}. Ultimately, we are interested in the effect that refinement has on our agent's affordance, $\potential(\agent)$. While calculating the contribution of an individual edge $\edge$ on the total $\potential(\agent)$ is probably intractable, it is not unreasonable to assume that it is related to the total set of paths that pass through the edge. We can model the global utility of an edge as $\q(\edge) = \sum_{\p\in\paths_{\segraph}(\edge)}{\pathPsym(\p)\potential(\p)}$, where $\paths_{\segraph}(\edge)$ is the set of all paths passing over the edge $\edge$.

Assuming that refining the edge $\edge$ will split it into $k$ smaller edges $\{\edge'_1, ... \edge'_k\}$ with disjoint fields\footnote{This is a simplification, in reality they will overlap.}, the total ``utility'' of these refined edges would be $\q_{\text{ref}}(\edge) = \sum_{i=1}^k{q(\edge'_i)}$. Then, the change in overall affordance from refining the edge $\edge$ would be $\Delta \q(\edge) = \q_{\text{ref}}(\edge) - \q(\edge)$. To get a tractable closed-form expression, we have to make several assumptions, which we elaborate on in section \ref{meth:deltaQ}. However, using these simplifications, we get that:
\begin{equation}
\Delta \q(\edge) = c \cdot N_{\p}(\edge)\rK(\edge)(1 - \rK(\edge))
\end{equation}
Where $N_{\p}(\edge)$ is the weighted number of paths that go through the edge. This score, which we call the edge's \emph{stress}, is an indicator of how useful it would be to refine an edge. An edge that has never been used ($N_{\p}(\edge)=0$) is unlikely to benefit from refinement. If an edge is completely reliable ($(1 - \rK(\edge))=0$), then the edge is fine and doesn't need to be refined. If the edge is completely unreliable ($\rK(\edge)=0$), then all of it's child-edges will \emph{also} be completely unreliable, so nothing will have been gained. If some used paths actually pass through $\edge$, and it is of intermediate reliability, then $\Delta \q(\edge)>0$, because refining the edge creates the possibility that some child-edges will be \emph{more} reliable than the parent, while others less: these lesser children can be filtered out (inhibited), allowing our agent to separate the wheat from the chafe, so to speak.

While ideally our agent would just refine edges with maximum $\Delta \q(e)$, in fact, our agent only has an estimate of this score, $\stress(\edge) = c \cdot n(\edge)\erK'(\edge)(1 - \erK'(\edge))$, where $n(\edge)$ empirically estimates how important $\edge$ is, and $\erK'(\edge)$ is the empirical estimate of the edge's reliability, without pseudo-counts.

\subsection{Ordering Paths by Affordance to Regulate Refinement}
In order to bias our agent towards refining edges with a high $\Delta q$, we need to somehow encourage our agent to traverse edges through which high-affordance paths pass. Directly choosing paths based on their affordance is intractable, but simply \emph{enforcing} that all edges used for pathfinding are themselves high-affordance guarantees that the resulting paths are also high-affordance. The affordance of an edge is just $\potential(\edge) = \rK(\edge) \cdot \measure(\edge)$, it's estimated value being $\epotential(\edge) = \erK(\edge) \cdot \measure(\edge)$. Thresholding \emph{edges} by their affordance rather than \emph{paths} is an approximation, and perhaps a hack, but in practice it seems to work. By prioritizing high-affordance paths before lower-affordance paths, we increase the odds that our agent will encounter edges with high $\Delta q$ (even when $\stress$ is a poor estimate), thus helping our agent to maximize $\potential(\agent)$ in the long-run.

We accomplish this by the use of a dynamic affordance-threshold, $\potentialbar$, which separates all edges into two sets: $\segElt{\potentialbar} = \{\edge \in \segedges : \epotential(\edge) < \potentialbar\}$ and $\segEgte{\potentialbar} = \{\edge \in \segedges : \epotential(\edge) \geq \potentialbar\}$, then restricting pathfinding to use only edges in $\segEgte{\potentialbar}$. Recall from section \ref{sec:failure_order} that edges that have failed are put in a set $\eforcedinactive{\segedges}$, with $\eactive{\segedges} = \segedges - \eforcedinactive{\segedges}$. We extend this logic, constructing the set of edges inhibited \emph{for any reason} as $\einactive{\segedges} = \segElt{\potentialbar} \cup \eforcedinactive{\segedges}$, with $\eactive{\segedges} = \segedges - \einactive{\segedges} = \segEgte{\potentialbar} - \eforcedinactive{\segedges}$. By starting $\potentialbar$ high and lowering it if a path cannot be found, our agent $\agent$ can prioritize higher-affordance paths.

Notably, prioritizing high-affordance paths does \emph{not} necessarily prioritize highly \emph{reliable} paths, so in a sense this is slightly unproductive. However, it's primary purpose is not to induce a sensible order on \emph{paths}, but instead to induce a sensible order on \emph{refinements}. Recall that mutation (including refinement) will happen, when it happens, where the agent is, and so controlling where the agent is ultimately controls where refinement happens. Essentially this order exposes edges that are large and reliable (good for using the segraph), \emph{or}, are very large and somewhat reliable, which would benefit from refinement (good for mutating the segraph).

For now though, we have to grapple with how to adapt the threshold, $\potentialbar$. By setting $\potentialbar$ high, our agent is guaranteed to sample only paths with high (estimated) potential from $\allpathsf[][]{\raisemath{0.1ex}{(\segverts, \eactive{\segedges})}}{}{\vertex, \vertex'}$... however if $\potentialbar$ is \emph{too} high, then $\allpathsf[][]{\raisemath{0.1ex}{(\segverts, \eactive{\segedges})}}{}{\vertex, \vertex'} = \varnothing$, and no path can be found! This can be fixed by simply lowering $\potentialbar$ until $\allpathsf[][]{\raisemath{0.1ex}{(\segverts, \eactive{\segedges})}}{}{\vertex, \vertex'} \neq \varnothing$ and a path can be sampled. If the path succeeds, that particular problem is solved, and the agent can try to solve other problems. Otherwise, the path will fail and be inhibited, eliminating it from $\allpathsf[][]{\raisemath{0.1ex}{(\segverts, \eactive{\segedges})}}{}{\vertex, \vertex'}$, driving enumeration until no paths are left and $\potentialbar$ is lowered again.

But, what happens when $\potentialbar = 0$ and so many edges have been inhibited that no path can be found to a goal?

\subsection{Usage and Mutation Feedback-Loop}
Thresholding edges by their affordance and gradually lowering the threshold can help the agent prioritize high-affordance paths, but in the end, the agent is unlikely to be able to actually enumerate $\achset{\segraph}{\SSK}$ due to locality constraints, necessitating that it mutate the graph before trying to enumerate the paths it ``missed'' in the last segraph.

To handle this, we have to address a basic problem with both the failure-driven enumeration and the affordance-ordering: they are irreversible! Once an edge has been actively inhibited (added to $\eforcedinactive{\segedges}$) due to a traversal failure, there is no mechanism to \emph{disinhibit} the edge, and thus, no chance of the edge ever being refined (since mutation only happens where the agent is). Likewise, once an edge has been passively disinhibited by lowering $\potentialbar$ (moving it from $\segElt{\potentialbar}$ to $\segEgte{\potentialbar}$), there is no mechanism to \emph{re-inhibit} that edge. While there are many possible ways to make active inhibition and passive disinhibition reversible, we took inspiration from homeostatic mechanisms in biology.

When our agent fails to find a path because $\allpathsf[][]{\raisemath{0.1ex}{(\segverts, \eactive{\segedges})}}{}{\vertex, \vertex'} = \varnothing$, the agent both lowers $\potentialbar$ \emph{and} sets $\eforcedinactive{\segedges} = \varnothing$, thus ``clearing'' the actively inhibited edges, and resetting the failure-driven enumeration. Likewise, after every traversal failure, $\potentialbar$ is raised slightly, discouraging low-affordance (and by proxy, low-reliability) edges. This further complicates the ``enumeration picture'' for the agent, as now it becomes possible that edges that haven't been tried can be re-inhibited by raising $\potentialbar$, and edges that have already failed can fail again before other non-failed edges can even be tried. In fact, it's technically possible that some completely reliable paths will never be tried at all, because they always end up below $\potentialbar$ before the agent can actually sample them!

Actually, it is refinement that recovers the ability of our agent to perform an enumeration. Consider that the goal for our agent, over a given segraph $\segraph$, is to enumerate $\achset{\segraph}{\SSK}$. Suppose that to enumerate $\achset{\segraph}{\SSK}$, the agent needs to use the edges $\segedges_{\text{req}} \subset \segedges$, with $\potential_{\text{min}}$ and $\potential_{\text{max}}$ the minimum and maximum affordances of edges in $\segedges_{\text{req}}$. By definition, $\forall \edge \in \segedges_{\text{req}}, \rK(\edge)=1$. Now suppose that there is another set $\segedges_{\text{mask}} \subset \segedges$ disjoint to $\segedges_{\text{req}}$, such that $\forall \edge \in \segedges_{\text{mask}}, \potential(\edge) > \potential_{\text{min}}$. That is, these edges all have higher affordance than the required edge with the least affordance. It is possible that repeated failures on these edges can raise $\potentialbar$ over $\potential_{\text{min}}$, masking part or all of $\segedges_{\text{req}}$. For each edge $\edge \in \segedges_{\text{mask}}$, $\rK(\edge) \in (0, 1]$, since if $\rK(\edge)=0$, it would mean that $\potential(\edge)=0$, and thus would not be in $\segedges_{\text{mask}}$. If $\rK(\edge)=1$, then there will never be a traversal failure crossing $\edge$ to raise $\potentialbar$.

If $\rK(\edge)<1$, then eventually, there will be an accumulation of failures and successes over $\edge$ that will trigger it's refinement. We can assume that for each child-edge, it's affordance will be $r_{\text{ref}} \cdot \potential(\edge)$, with $0 < r_{\text{ref}} < 1$. Let $\segedges_{\text{ref}}(\edge)$ be the set of ``child'' edges produced by refining $\edge$. This removes $\edge$ from $\segedges_{\text{mask}}$ and adds to $\segedges_{\text{mask}}$ each $\edge' \in \segedges_{\text{ref}}(\edge)$ for which $\potential(\edge') > \potential_{\text{min}}$. If some of these edges have $\rK(\edge')=1$, then they will never fail and thus never raise $\potentialbar$. Otherwise, the same logic holds, until every edge that was in $\segedges_{\text{mask}}$ is removed (because it's affordance falls bellow $\potential_{\text{min}}$) or is perfectly reliable. At that point, even though there are probably new not-reachable achievement classes on this new graph, $\segraph'$, it is possible to try some subset of paths over $\segraph'$ to enumerate the achievement set of the original segraph $\segraph$, $\achset{\segraph}{\SSK}$. The same process guarantees that, at some point, there will be a $\segraph''$ which can be used to enumerate $\achset{\segraph'}{\SSK}$.

\subsection{Extension and Linking}
As long as every vertex (that needs to be refined) is eventually refined, to guarantee a balanced sequence of segraphs it is sufficient to simply extend each vertex \emph{before} it gets refined. In practice, we modify this condition slightly, instead extending every vertex after it has been visited. This means that some vertices will actually never be extended, because they will be refined before they can be extended. However, our intuition is that, if a vertex has never been visited, then there is no reason to expect that any neighboring states can be visited. If a vertex is refined, and some of it's child-vertices can be visited, then \emph{they} can be extended. Even though not every vertex actually gets visited, the statespace (or the parts that are physically accessible) are eventually covered.

As for the linking operation, we take an approach that is, frankly, heuristic. All vertices with overlapping fields are automatically linked to each other. If our agent starts a transition in $\vertex$ trying to reach $\vertex'$, for any other vertex it passes through, $\vertex''$, it adds an edge from $\vertex$ to $\vertex''$. This provides a local mechanism for increasing the connectivity of $\segraph$ without going all the way to fully-connecting $\segraph$. This is an area for further study.

\subsection{Ordering by Epistemic Value}
What is ``epistemic value''? In order to ultimately learn something, our agent will need to collect experiences trying to cross the edges of its cognitive graph $\segraph$, which amounts to a kind of exploration. If there are edges that haven't been traversed, the agent cannot know if they could be useful for reaching a particular goal. To ensure that ``no stone is left unturned'', we adopt a size-normalized count-based \emph{epistemic-score} for edges:
\begin{equation}
\epistemicscore(\edge) = \frac{\measure(\edge)}{n(\edge) + n_{\epistemicscore}(\edge) + 1}
\end{equation}
where $n(\edge) = n_s(\edge) + n_f(\edge)$, $1$ is added for numerical stability, and $n_{\epistemicscore}(\edge)$ is an extra variable that can be used to adjust an edge's score independently of the number of actual visitations. The point of this score is simple: larger edges (greater $\measure(\edge)$) have more potential plans to try, and less-visited (lower $n(\edge)$) edges have less certain robustness estimates. This means that while exploring, our agent doesn't pick a \emph{vertex} to find a path toward, but rather an \emph{edge} $\goal{\edge}$ to find a path toward. We can slightly adapt our notation to reflect this: $\allpathsf[][]{\segraph}{*}{\vertex, \goal{\edge}}$ is the set of all paths over $\segraph$ that start at $\vertex$ and end by crossing $\goal{\edge}$: if $\goal{\edge} = (\vertex', \goal{\vertex})$, then $\allpathsf[][]{\segraph}{*}{\vertex, \goal{\edge}} = \{\p + \goal{\vertex} : \p \in \allpathsf[][]{\segraph}{*}{\vertex, \vertex'}\}$.

We can model our agent's choice of a goal with the discrete probability distribution $\goalP$ over all of the edges $\segedges$ over its segraph. Without being prematurely specific, we can say that $\goalP(\edge)$ is a monotonically increasing function of $\epistemicscore(\edge)$, so the higher $\epistemicscore(\edge)$ is, the more likely $\edge$ is to be selected as a goal. If the agent reaches the edge $\edge$ and tries to cross it (successfully or unsuccessfully), then $n(\edge)$ will increment by $1$ and $\epistemicscore(\edge)$ will decrease, thereby decreasing the chance that $\edge$ is selected as a goal. This doesn't automatically mean that the agent will attempt to reach every edge in $\segedges$ ``in-order'', but under some mild ergotic assumptions, it does mean that every edge will eventually be visited (or, \emph{attempt} to be visited). However, reaching every edge is not the same as trying every path!

We model our agent's choice of a path from a starting vertex to a goal edge with the discrete probability distribution $\pathP{\segraph}{\vertex}{\goal{\edge}}$ over the set of paths $\allpathsf[][]{\segraph}{*}{\vertex, \goal{\edge}}$. The final form of this distribution will be determined by several factors not yet introduced, but for now, it is sufficient to begin by assuming that $\pathP{\segraph}{\vertex}{\goal{\edge}}$ is uniform over $\allpathsf[][]{\segraph}{*}{\vertex, \goal{\edge}}$, making any path from $\vertex$ to $\goal{\edge}$ equally likely. If that is the case, then randomly sampling targets from $\goalP$ and paths from $\pathPsym$ guarantees that eventually, every path in $\allpaths[][]{\segraph}{*}$ is eventually tried, giving us a trivial enumeration of $\allpaths[][]{\segraph}{*}$.

\subsection{Ordering by Simplicity}
All things being equal, a shorter path is a better path. If $\allpathsf[][]{\raisemath{0.1ex}{(\segverts, \eactive{\segedges})}}{k}{\vertex, \edge}$ is the set of all paths from $\vertex$ to $\edge$ of length $k$, then $\allpathsf[][]{\raisemath{0.1ex}{(\segverts, \eactive{\segedges})}}{*}{\vertex, \edge} = \bigcup_{n=1}^{\infty}{\allpathsf[][]{\raisemath{0.1ex}{(\segverts, \eactive{\segedges})}}{k}{\vertex, \edge}}$. Our agent can first try paths in $\allpathsf[][]{\raisemath{0.1ex}{(\segverts, \eactive{\segedges})}}{1}{\vertex, \edge}$, if a path is found that works, then a single instance of $\achclass{\segraph}{\vertex}{\edge}$ has been found and there is no need to try longer paths. If no length-1 path is found, the agent can begin trying paths in $\allpathsf[][]{\raisemath{0.1ex}{(\segverts, \eactive{\segedges})}}{2}{\vertex, \edge}$, and so-on until a successful path is found.

In fact, in combination with the failure-driven inhibition logic (see section \ref{sec:failure_order}), our agent can get away with \emph{only} sampling from among the shortest available paths. Let $k_{\text{min}}$ be the smallest $k$ for which $\allpathsf[][]{\raisemath{0.1ex}{(\segverts, \eactive{\segedges})}}{k}{\vertex, \edge} \neq \varnothing$, and assume that:
\begin{equation}
\pathP{\raisemath{0.1ex}{(\segverts, \eactive{\segedges})}}{\vertex}{\edge}(\p) = \begin{cases}
0 & \text{if }\abs{\p} > k_{\text{min}} \\
p(\p) & \text{otherwise, with $p(\p)>0$}
\end{cases}
\end{equation}
If a plan sampled from $\pathP{\raisemath{0.1ex}{(\segverts, \eactive{\segedges})}}{\vertex}{\edge}$ succeeds, then the search for an element in $\achclass{\segraph}{\vertex}{\edge}$ is complete. Otherwise, every plan sampled from $\pathP{\raisemath{0.1ex}{(\segverts, \eactive{\segedges})}}{\vertex}{\edge}$ fails, and by definition, is removed from further consideration: after each plan in $\allpathsf[][]{\raisemath{0.1ex}{(\segverts, \eactive{\segedges})}}{k}{\vertex, \edge}$ fails, $\allpathsf[][]{\raisemath{0.1ex}{(\segverts, \eactive{\segedges})}}{k}{\vertex, \edge}$ is empty! This means that $k_{\text{min}} \neq k$, and instead, $k_{\text{min}} > k$, meaning that when our agent samples a new plan, it will automatically come from the next ``length-set'' of plans, $\allpathsf[][]{\raisemath{0.1ex}{(\segverts, \eactive{\segedges})}}{k+n}{\vertex, \edge}$. Thus, selecting shortest paths along with failure-induced inhibition automatically induces an ordering of paths by length.

\section{Adaptive Realtime Metasearch over Segraphs}
To recap: we imagine an agent $\agent$ that wants to be able to reach any point in an abstract statespace $\SS = \mathbb{R}^d$ from any other point in $\SS$. It is endowed with a continuous sensorimotor controller $\K$, which cannot fully navigate $\SS$ on it's own. We equip $\agent$ with a segraph $\segraph$ that decomposes $\SS$ into chunks that $\K$ can more easily navigate between, reaching distant goals via paths over $\segraph$. By following a path over the segraph, the agent collects experiences that inform it about the traversability of individual edges. This information changes the probability distribution over paths, which changes which plans are tried by the agent, which subsequently, changes how the segraph itself is modified, creating a tight behavior/segraph feedback loop. Effectively, this feedback loops involves a search over paths driving (and being driven by) a search over segraphs, which occurs in real-time, and is adaptive. Because of this, we call the algorithm the Adaptive Real-time Metasearch over Segraphs, or ARMS, algorithm. We describe the ARMS algorithm, providing motivation in terms of balancing segraph mutation. For the description of the algorithm itself, we actually do not assume that the agent has a fixed goal: rather, we imagine the agent's goal is simply to explore as much of the space of actions as possible, so that any particular goal can be reached on-demand.

\subsection{Algorithm}
\begin{figure}[h]
\includegraphics[width=\textwidth]{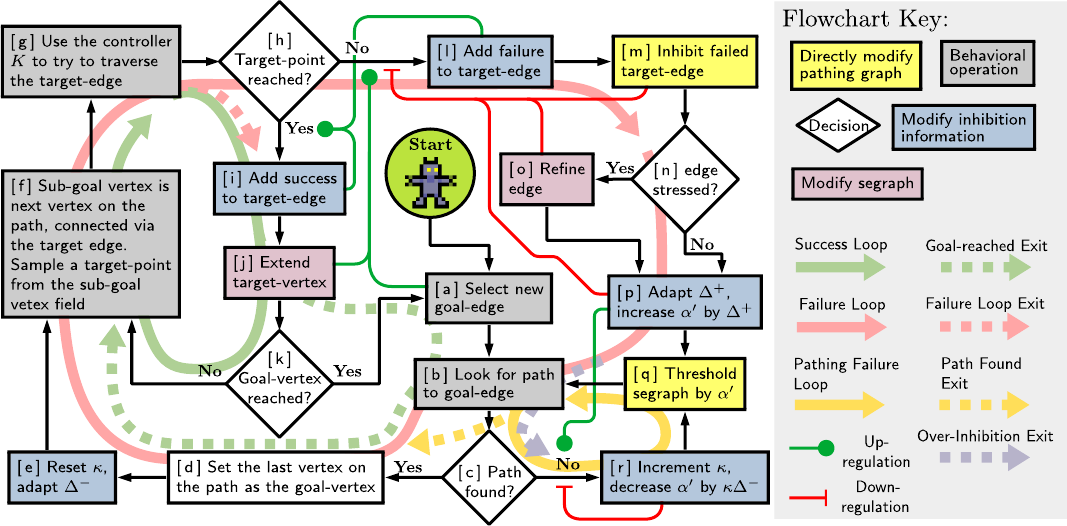}
\caption{\label{fig:flowchart}The search algorithm can be decomposed into three major interconnected loops: the first is the ``success'' loop (solid green arrow [\textsf{f}, \textsf{g}, \textsf{h}, \textsf{i}, \textsf{j}, \textsf{k}]). Staying on this loop results in reaching the selected goal, triggering the selection of a new goal and ideally leading back to the success loop (dashed green arrow). However, selection of ambitious goals and mismatch between $G^{\star}$ and the environment will usually lead to failed edge-traversals, triggering the ``failure-loop'' (solid red arrow, [\textsf{f}, \textsf{g}, \textsf{h}, \textsf{l}, \textsf{m}, \textsf{n}, \textsf{o?}, \textsf{p}, \textsf{q}, \textsf{b}, \textsf{c}, \textsf{d}, \textsf{e}]). Both the success and failure loops update the reliability estimates of edges, allowing the agent to more easily distinguish good from bad edges, which long-term, promotes the success-loop. Repeated failure raises $\potentialbar$ and can result in a pathing failure (over-inhibition exit). Each pathing failure lowers $\potentialbar$, so the ``pathing failure loop'' (yellow arrow) is self-inhibiting, and will eventually lead back to the success or failure loops.}
\end{figure}

Each edge has an \emph{estimated affordance} $\epotential(\edge)$ which is compared to a threshold $\potentialbar$. If $\epotential(\edge) < \potentialbar$, then $\edge$ is \emph{inhibited}. Inhibiting an edge prevents it from being used during pathfinding, allowing entire classes of paths to be instantaneously turned off, or if an edge is disinhibited, turned back on. Edges can also be manually (dis)inhibited, independent of their $\potential$-value. Disinhibiting low $\potential$ edges encourages revisiting less reliable or more-fine-grained options, and inhibiting low $\potential$ edges encourages the use of more reliable paths. A major element that controls the behavior of the algorithm is the exact way that $\potentialbar$ is adapted. Because pathfinding can be computationally expensive, trying to find a path and failing often is very expensive, so lowering $\potentialbar$ too slowly can be catastrophic for computational efficiency. On the other hand, lowering it too fast can make pathfinding under-selective. The same is true of raising $\potentialbar$. We represent the rate of lowering $\potentialbar$ by $\inhbneg$, and the rate of raising it by $\inhbpos$.

Fig. \ref{fig:flowchart} shows a basic conceptual overview of the algorithm as a flowchart. The agent starts out at a point inside of the field of a \emph{start vertex} $\start{\vertex}$ (the segraph $\segraph$ may have an arbitrary initial configuration, perhaps with only a single vertex). \itemA{} The agent calculates an \emph{epistemic score} $\epistemicscore(\edge)$ for each edge $\edge$, equal to an edge's size divided by its total number of traversals. Then, the agent selects the edge with the highest $\epistemicscore$ and sets it as it's \emph{goal-edge}, $\goal{\edge}$, analogous to various \emph{count-based} exploration methods \cite{bellemare2016unifying}. $\goal{\edge}$ is manually \emph{dis}inhibited, and all manual inhibition is cleared, then  \itemB{} the agent uses its pathfinding algorithm to sample a path from $\pathP{(\segverts, \eactive{\segedges})}{\start{\vertex}}{\goal{\edge}}$ to find a path $\p$ beginning at $\start{\vertex}$ and ending (by passing through) the edge $\goal{\edge}$. \itemC{} If such a path is found, \itemD{} the last vertex on $\p$ is designated the \emph{goal-vertex}, $\goal{\vertex}$. \itemE{} Then $\inhbneg$ is adapted to regulate the number of pathfinding failures (we will return to this).

Next \itemF{} the agent takes the next vertex from the path and sets it as its \emph{sub-goal vertex} $\subgoal{\vertex}$, the edge between $\agent$'s current vertex $\start{\vertex}$ and the subgoal vertex is designated the \emph{target edge} $\subgoal{\edge}$, which $\agent$ will try to traverse next. $\agent$ uniformly samples a target point from $\field(\subgoal{\vertex})$ and passes the displacement to its controller $\K$. Then \itemG{} $\agent$ tries to physically traverse the edge, ultimately ending up at a new point $\s'$. \itemH{} If $\agent$ reached the target point, \itemI{} the traversal is marked as a success, and the agent updates its current vertex to $\start{\vertex} \leftarrow \subgoal{\vertex}$, after which \itemJ{} the vertex the agent just reached is extended. Then \itemK{} the agent checks if it has reached the goal vertex $\goal{\vertex}$. If it has, the agent selects a new goal (back to \itemA{}), otherwise, the agent continues along the path \itemF{}.

If at \itemH{} the agent \emph{doesn't} reach the target point, then \itemL{} the traversal is marked as a failure, and the agent recalculates what vertex it is in. \itemM{} $\agent$ \emph{manually inhibits} the failed edge (inhibited independently of its actual $\potential$-value) by adding it to $\eforcedinactive{\segedges}$ so that it can't immediately be used again in a path. If the edge is stressed ($\stress(\edge)>1$), the edge is refined, in the hopes that some of the child-edges are more reliable than the original edge. Here, the $c$ term in the stress score becomes a \emph{control parameter} that determines the overall refinement rate of the algorithm. Then, $\inhbpos$ is adapted to reflect the difference between activity potential of the edge and $\potentialbar$, and $\potentialbar$ is increased to discourage further failures on other edges.

Then \itemQ{} the graph is threshholded by $\potentialbar$, producing a new pathing graph so that \itemB{} $\agent$ can look for a new path. Repeated failures may result in $\potentialbar$ being so high that a path cannot be found between $\start{\vertex}$ and the goal edge $\goal{\edge}$, resulting in a pathing failure (\itemC{} branches to \itemR{}), which increments $\APFC$ (the number of consecutive pathing failures) and decreases $\potentialbar$ by $\APFC \inhbneg$ (decreasing the odds of another pathing failure), which then goes back to \itemQ{}. Jumping back to \itemE{}, upon pathfinding \emph{success}, $\APFC$ is reset back to $0$, and $\inhbneg$ is adapted to reflect the difference between $\potentialbar_0$ (initial threshold value before pathing failure) and the current threshold value, to regulate the number of consecutive pathing failures necessary before the threshold is low enough to find a path. This controls the balance between pathfinding selectivity and pathfinding computational cost. Pathing failure also clears all manual inhibition to ensure that a path can be found. Additionally, this also triggers $n_{\epistemicscore}(\goal{\edge})$ to incremented, to discourage ``fixation'' on an unreachable goal.

The ARMS algorithm has several means of self-regulation. Overall the algorithm has three interlocking loops, shown in Fig. \ref{fig:flowchart} as thick colored arrows: the ``success'', ``failure'', and ``pathing failure'' loops. Edge-traversal collects information that helps the agent discriminate between edges, so long-term the success loop is up-regulated by both \itemI{} (the success loop) and \itemL{} (the failure loop). However, the success loop causes the addition of new edges via extension \itemJ{} and reliability-agnostic selection of goal-edges \itemA{}, which causes the success loop to up-regulate the failure loop. The failure loop down-regulates itself by (1) inhibiting an edge after failure to prevent immediate repeat-failure, (2) potentially refining an edge to create better options for pathfing in the long-term, and (3) raising $\potentialbar$ to lower the chance of other low-reliability edges being used. This however also up-regulates the ``pathing-failure loop'', as over-inhibition can prevent a path from being found. To counteract this, the pathing-failure loop down-regulates itself by \emph{lowering} $\potentialbar$, allowing the system to return to either the success or failure loops. See \ref{meth:lambda} for details of $\potentialbar$ adaptation.

\subsection{Resource constraints and Prior Knowledge}

One important addition of the algorithm, strictly beyond the scope of our theory, is how to deal with resource constraints. With a fixed size neural network for representing the segraph, there is an upper limit on how many vertices the segraph can have. We could just stop the algorithm at this point, but instead, we use an ad-hoc score (see \ref{meth:ghost}) to determine which vertices might be acceptable to delete. As the agent gets closer to the its built-in vertex limit, it slows down the rate of refinement (by lowering $c$), and once it reaches the vertex-cap, it starts to delete apparently useless vertices. The slow-down in refinement allows the segraph to naturally stabilize to a more-or-less ``final'' configuration, which is only further modified by better and better estimates of edge-reliability.

We were also interested in testing the incorporation of prior knowledge into the ARMS algorithm. The algorithm described above corresponds to a ``naive'' agent. The simplest mechanism we could think of was to allow the ARMS algorithm some ability to decide how large new vertex-fields should be, based on sensory data. If a wide area near the agent seems to be open, then the agent can decide, when extending a vertex, that it should place a larger field there: if the area is instead a barrier, it can decide to add much smaller vertices there (making them more likely to be inhibited by the affordance threshold). We call such an agent ``astute''. The ``sensory information'' that we give the astute agent is just the value $\rK(\vect{s}, \vect{s}')$, it then samples points $\vect{s}'$ around a field that needs to be extended, calculates $\rK(\vect{s}, \vect{s}')$, then stores that value in a HaRK memory (Fig \ref{fig:results_1}A). At the same points, it performs a query at different resolutions: starting at a low resolution (blurring stored $\rK$ values together on a large spatial scale), it checks for points with high average local $\rK$, adding new vertices at that scale if it's above a threshold. Any points that remain uncovered, are quaried at a higher and higher resolution, until the vertex is fully extended. We can model a ``misled'' agent by simply inverting $\rK(\vect{s})$.

\section{Neural and Algorithmic Instantiation}

Aside from the theoretical interest we have in characterizing and mimicking the adaptive abilities of living organisms, we are also interested in demonstrating that, at least in principle, our method of \emph{achieving} hyperadaptability can be implemented in the nervous system. Rather than try to develop a very detailed neurobiological model, we focused on a more abstract implementation of artificial neural networks, relying heavily on linear algebra and Fourier theory.

In particular, we take inspiration from the hippocampus and entorhinal cortex. We consider that a \emph{segraph} can be thought of as a very simplified model of a \emph{cognitive graph} and a \emph{cognitive map}. Place cells \cite{o1976place} in the hippocampus and grid cells \cite{hafting2005microstructure} in the entorhinal cortex have been implicated as instantiating cognitive graphs and maps, providing a neurological basis for these abstract constructs. We associated place-cells with the vertices of a segraph, and we associated grid cells and band cells with the mechanism for granting vertices their fields.

We use a ``binding'' operator corresponding to a mathematically simple form of one-shot Hebbian learning, both to construct the graph, and to ground it in the statespace $\SS$. We choose to represent graph-vertices as one-hot vectors, as this seems to provide a reasonable facsimile of place cells, and gives us some technical benefits we will describe later.  The statespace is represented by a complex-valued model of \emph{band cells} \cite{krupic2012neural}. By ``binding'' vertex-vectors together, we encode the \emph{topology} of the segraph, and by binding vertex-vectors to band-cell population vectors, we endow the vertices with their \emph{fields}.

\subsection{Hebbian Learning}
The segraph itself must be able to be modified rapidly in order to achieve hyperadaptability through the means we have already outlines. We achieve this rapid mutability by defining a simple Hebbian-type learning rule that can accomplish the association of arbitrary vectors in a single operation, which we refer to as \emph{binding}.

\begin{box_definition}{Binding}{binding}
The \emph{bind} between two vectors $\vect{a} \in \mathbb{C}^N$ and $\vect{b} \in \mathbb{C}^M$ is the matrix:
\begin{equation*}
\vect{a} \bind \vect{b} = \frac{\vect{b}\vect{a}^*}{\norm{\vect{a}}^2} \in \mathbb{C}^{M \times N}
\end{equation*}
where if $\vect{c} \in \mathbb{C}^N$, then $(\vect{a} \bind \vect{b}) \vect{c} = \vect{b} \tfrac{\norm{\vect{c}}}{\norm{\vect{a}}} \cosim{\vect{a}}{\vect{c}}$, with $\cosim{\vect{a}}{\vect{c}} = \tfrac{\vect{a}^*\vect{b}}{\norm{\vect{a}}\cdot\norm{\vect{b}}}$ the cosine similarity between $\vect{a}$ and $\vect{c}$.
\end{box_definition}
\noindent{}We can interpret a sum of such ``bind'' matrices as a dictionary, as
\begin{equation*}
\left[\sum_{i=1}^{k}{\vect{a}_i \bind \vect{b}_i}\right] \vect{c} = \sum_{i=1}^{k}({\vect{a}_i \bind \vect{b}_i}) \vect{c} = \sum_{i=1}^{k}{\vect{b}_i \frac{\norm{\vect{c}}}{\norm{\vect{a}_i}}\cosim{\vect{a}_i}{\vect{c}}}
\end{equation*}
If all of the ``key'' vectors $\vect{a}_i$ are orthogonal to each other (have zero cosine-similarity), and $\vect{c} = \vect{a}_j$, then:
\begin{equation*}
\left[\sum_{i=1}^{k}{\vect{a}_i \bind \vect{b}_i}\right] \vect{a}_j = \sum_{i=1}^{k}\vect{b}_i\frac{\norm{\vect{a}_j}}{\norm{\vect{a}_i}}\cosim{\vect{a}_i}{\vect{a}_j} = \vect{b}_j\frac{\norm{\vect{a}_j}}{\norm{\vect{a}_j}}\cosim{\vect{a}_j}{\vect{a}_j} = \vect{b}_j
\end{equation*}
meaning it's possible to selectively retrieve stored associations. Our approach is comparable to \cite{schlag2020learning} and \cite{ba2016using}. See \ref{app:hebb} for details.

\subsection{Graph Working Memory}
Let $\graph = (\vertices, \edges)$ be a directed graph with vertices $\vertices$ and edges $\edges$. Each vertex $\vertex_i\in \vertices$ is assigned a unique $M$-dimensional one-hot vector $\vect{v}_i$, with all elements $0$ except for the $i^{\text{th}}$, which is $1$. If an edge $\edge \in \edges$ is an ordered tuple of vertices with one-hot vectors $\vect{v}_{i}$ and $\vect{v}_{j}$, then the edge $\edge_{i,j}$ is represented by a matrix $\mat{E}_{i,j} = \vect{v}_{i} \bind \vect{v}_{j}$. The graph $G$ is represented by the matrix $\mat{G} = \textstyle\sum_{\edge_{i,j} \in E} \mat{E}_{i,j}$, so that for any vertex $\vertex_i$, $\mat{G}\vect{v}_i$ will be the sum of one-hot vectors for the neighbors of $\vertex_i$ (Fig. \ref{fig:gwmHaRK}A). This makes $\mat{G}$ essentially an adjacency matrix. Being represented by one-hot vectors, they are easy to individually identify.

\subsection{Wave-based Pathfinding}\label{subsec:pathfinding}
As plans must be sampled from paths, and paths must be computed, efficient online pathfinding is essential for our agent. Our algorithm could in-principle be substituted for any standard pathfinding algorithm, but we designed this algorithm with two things in mind: first, taking advantage of parallelism (at least, in a physical neural substrate), and second, the intuition that when imagining a plan to reach a goal, it is common to \emph{first} imagine some ``intermediate'' steps, and gradually decompose the problem into more and more fine-grained intermediate steps.

\begin{figure}
\includegraphics[width=1\textwidth]{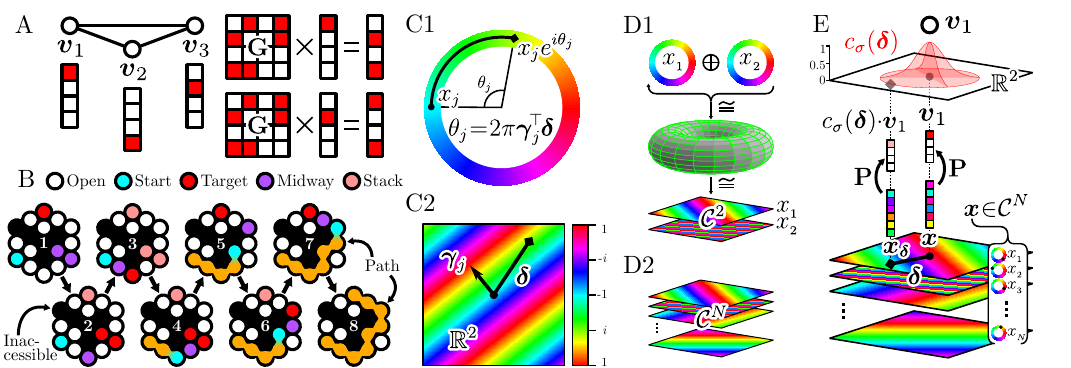}
\caption{\label{fig:gwmHaRK}(A) Each vertex of the graph is assigned a one-hot vector, which are bound together via Hebbian learning in a matrix $\mat{G}$. (B) An example of the recursive decomposition used by the pathing algorithm, excluding wave-propagation for brevity. Adjacent cells are connected, pathing proceeds by recursively finding the vertices midway between some start and target vertices. (C1) With $\mathcal{C}$ the set of unit-complex numbers, each band-cell's state is $x_j \in \mathcal{C}$. An update $\vect{\delta}\in\mathbb{R}^d$ projects onto the oriented frequency \smash{$\vect{\gamma}_j$} of the band-cell. (C2) The $j^{\textrm{th}}$ band-cell's state is a sinusoidal grating in $\mathbb{R}^d$ ($d=2$ example shown). (D1) The Cartesian product of two such cells is a torus $\mathcal{C}^2$. (D2) In higher dimensions we just show the stacked gratings. (E) In order to ground the graph in space, each vertex one-hot vector is bound via Hebbian learning inside the matrix $\mat{P}$ to a vector of band-cell activations $\vect{x}_{\vect{p}}$ encoding a specific point in space $\vect{p}\in\mathbb{R}^d$. The Hebbian learning is modulated so that the response of the vertex-vector decays as a Gaussian with respect to displacement from the ``center'' of the stored location. By binarizing this activity, we get a ``place-field''.}
\end{figure}

We take advantage of the one-hot-vector graph-representation to implement the $\pwave$ pathfinding algorithm. By propagating across the graph a forward ``wave-front'' from a set of starting vertices and a backward ``wave-front'' from some target vertices (compare to \cite{soulignac2006fast}), we can find the set of ``mid-point'' vertices where the waves overlap in $O(k)$ matrix multiplications, where $k$ is the length of the shortest path between the start and target vertices. By treating the mid-point vertices as a new set of target vertices and ignoring the cost of matrix multiplication, $\pwave$ recursively solves the pathfinding problem in time $O(k\log(k))$, and moreover, finds the \emph{first} vertex on the path in time $O(k)$, meaning the agent can begin following the path before the whole path has been found. A toy-example is shown in Fig. \ref{fig:gwmHaRK}B.

The pathfinding algorithm $\pwave$ proceeds by recursive application of a simple midway-point-finding process. A combination of passive $\potentialbar$-thresholded inhibition and manual inhibition over a segraph $\segraph$ will yield a subgraph $\segraph' = (\segverts, \eactive{\segedges})$ that is used for pathfinding. These edges can be represented via Hebbian learning by a matrix $\mat{G}$. The $\pwave$ algorithm takes as input a set of starting vertices $V'_s \subset V$ and a set of goal vertices $V'_g \subset V$, which have corresponding multi-hot vector representations $\vect{m}'_s$ and $\vect{m}'_g$. Then, a ``forward wave-front'' and ``backward wave-front'' propagate out from $\vect{m}'_s$ and $vect{m}'_g$ respectively, until the two wavefronts overlap at a set of vertices represented by the multi-hot vector $\vect{m}$. The ``goal'' vertices $\vect{m}'_g$ are pushed to a stack, and the process repeats with $\vect{m}'_s$ and $\vect{m}$ representing a ``sub-problem''.

This process repeats recursively, until the sub-problem is trivial. At this point, vertices may be chosen and added to the agent's path, with the last sub-goal popped from the stack. See \ref{meth:pathing} for details.

\subsection{Harmonic Relational Keys}
Having introduced our method of constructing cognitive \emph{graphs}, we must now explain how to construct cognitive \emph{maps} representing $\mathbb{R}^d$ statespaces. We take inspiration from band cells, and represent the state of each band cell as a unit-magnitude complex number $x_j \in \mathbb{C}$, with an oriented frequency $\vect{\gamma}_j \in \mathbb{R}^d$ (Fig. \ref{fig:gwmHaRK}C). The population of band cells is a vector $\vect{x} \in \mathbb{C}^N$ ($N$ large, Fig. \ref{fig:gwmHaRK}D), and each $\vect{\gamma}_j$ is sampled from a distribution $p_{\vect{\gamma}}(\vect{\gamma})$ over $\mathbb{R}^d$. If we associate the key $\vect{x}_{\vect{p}}$ with a point $\vect{p}\in\mathbb{R}^d$, then the key corresponding to the point $\vect{p} + \vect{\delta}$ is given by $\vect{x}_{\vect{p} + \vect{\delta}} = f_{\mathrm{HaRK}}(\vect{x}_{\vect{p}}, \vect{\delta}) =  \vect{x}_{\vect{p}} \odot e^{2\pi i \mat{\Gamma} \vect{\delta}}$, where $\odot$ is an element-wise product, and $\mat{\Gamma}$ is the matrix of oriented frequencies $\vect{\gamma}_j$.

$f_{\mathrm{HaRK}}$ can assign a unique $\mathbb{C}^N$ key to each point in $\mathbb{R}^d$. We can use element-wise modulation to also control the \emph{resolution} of information stored in $\mathbb{R}^d$. The bind matrix between an input-modulated key-vector $\vect{x} \odot \vect{\mu}$ and the value-vector $\vect{y}$ is $[(\vect{x} \odot \vect{\mu}) \bind \vect{y}]$. We then query this matrix with a $\vect{\delta}$-offset key $\vect{x}_{\vect{\delta}} = f_{\mathrm{HaRK}}(\vect{x}, \vect{\delta})$ modulated by $\vect{\eta}$, yielding $\tilde{\vect{y}}(\vect{\delta}) = [(\vect{x} \odot \vect{\mu}) \bind \vect{y}] (\vect{x}_{\vect{\delta}} \odot \vect{\eta})$, which can be decomposed as $\tilde{\vect{y}}(\vect{\delta}) = c(\vect{\delta}) \vect{y}$, where $c(\vect{\delta}) = \mathcal{F}_{\vect{\gamma}}^{-1}[g(\vect{\gamma})p_{\vect{\gamma}}(\vect{\gamma})]$, with $\mathcal{F}^{-1}$ being the inverse Fourier transform and $g(\vect{\gamma})$ a function mapping oriented frequencies $\vect{\gamma}$ to modulation weights: in other words, $g(\vect{\gamma})$ determines $\vect{\mu}$ and $\vect{\eta}$. For this initial work we choose $c(\vect{\delta})$ to be an isometric Gaussian $c_{\sigma}(\vect{\delta})$ with width $\sigma$ (Fig. \ref{fig:gwmHaRK}E), though this is not strictly necessary. Manipulating $\sigma$ can be used to control the resolution of information on storage, on retrieval, or both to achieve statespace band-pass filtering.

Our approach to harmonic representations can be thought of as generalization and simplification of the model presented in \cite{rodriguez2019hexagonal}. See \ref{app:hark} for details.

\section{Results}
Initial development of the ARMS algorithm occurred in 2D mazes, since this allowed for easy visual inspection and intuitive understanding. We began by testing the ability of the ARMS algorithm in it's naive, astute, and misled variants to learn to navigate around a fixed set of regular rectangular and  irregular polygonal mazes. To measure the agent's ability in a way that was irrespective of maze-size, we calculate the agent's segraphs' \emph{reliability} $R(\segraph)$, which is a distance-weighted average of the reliability of the graph for navigating between pairs of points sampled from the statespace, allowing the agent to re-path if necessary to reflect the operation of the ARMS algorithm (see Methods \ref{meth:r_of_g}). $R(\segraph)=0$ means complete unreliability, $R(\segraph)=1$ means total reliability. $R(\segraph)$ can be thought of as distance-normalized and environment-size-normalized variant of the affordance of the graph, $\potential(\segraph)$. We also checked the resilience of ARMS to shifts in the environment, its sensitivity to initial field-size guess, and its ability to handle higher dimensional statespaces. Finally, we tested ARMS in a dynamic environment with rewards.

\subsection{2D mazes}

\begin{figure}[h]
\includegraphics[width=\textwidth]{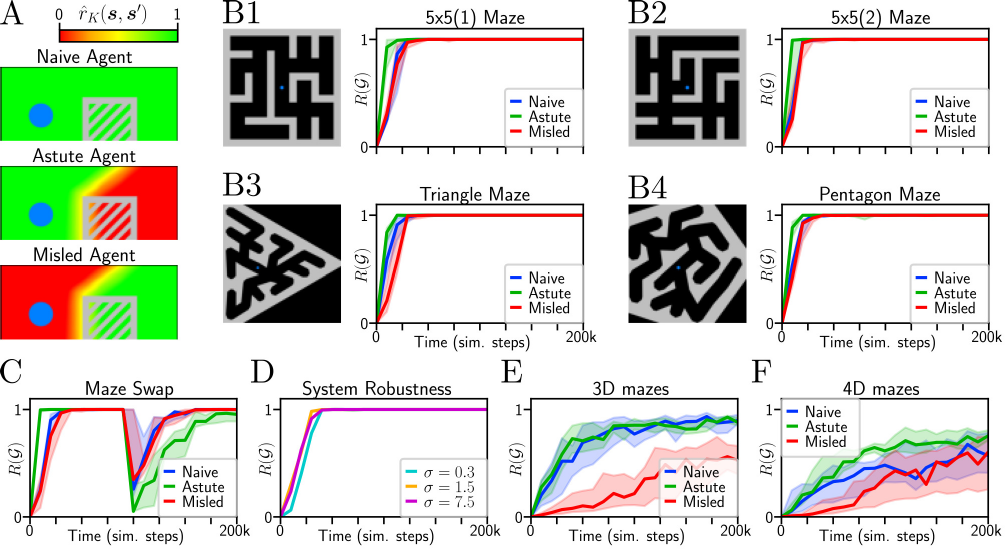}
\caption{\label{fig:results_1} (A) Agents can have different knowledge of spatial navigability, either being Naive, Astute, or Misled. (B) Naive, Astute, and Misled agent's ability to construct a reliable segraph in different maze-environments. (C) Naive, Astute, and Misled agent's ability to recover when the maze is changed (at 100k sim. steps). (D) The robustness of the ARMS algorithm to different initial field sizes. (E) Naive, Astute, and Misled agents in random 3D mazes. (F) Naive, Astute, and Misled agents in random 4D mazes.}
\end{figure}

To build mazes, we generated randomly connected 25-node graphs, converting the graph into a maze by 1) finding a random spanning tree of the graph and 2) converting the topology of the spanning tree into a continuous geometric environment with barriers. Two 5x5 mazes (5x5(1) and 5x5(2)), a triangular maze, and a pentagonal maze were randomly generated (see Methods \ref{meth:maze}) and used to determine if 1) the ARMS algorithm could successfully learn to navigate around these mazes and 2) whether incorporating prior knowledge about environment navigability could influence the ability of the ARMS algorithm to learn these mazes (Figure \ref{fig:results_1}A). We ran 8 simulations for each condition (maze $\times$ prior knowledge) for 200,000 timesteps, measuring $R(\segraph)$ every 10,000 timesteps (because the measurement is quite computationally expensive to make). The median (solid line) and interquartile range (shaded area) are shown for each condition in Figure \ref{fig:results_1}. Across all conditions, ARMS achieves nearly perfect navigation ability within approximately 30,000 timestep, and as the first four rows of Table \ref{table:1} show, by the end of each simulation, all ARMS agents reached perfect navigation ability across all mazes and all prior knowledge conditions. At the beginning, it appears that having some accurate prior knowledge does speed up learning, but surprisingly, misleading prior knowledge only seems to have a negative effect compared to the naive agent in the Triangular maze, suggesting that our means of adding prior knowledge to ARMS is conceptually shallow and does not correspond to the kinds of prior knowledge that we might expect animals or humans to have.

\begin{table}
\begin{center}
\begin{tabular}{|c||c|c|c|}
\hline
Maze & Naive Agent & Astute Agent & Misled Agent \\
\hline \hline
5x5(1) & (1.000, \textbf{1.000}, 1.000) & (1.000, \textbf{1.000}, 1.000) & (1.000, \textbf{1.000}, 1.000) \\
\hline
5x5(2) & (0.999, \textbf{1.000}, 1.000) & (1.000, \textbf{1.000}, 1.000) & (1.000, \textbf{1.000}, 1.000) \\
\hline
Triangle & (1.000, \textbf{1.000}, 1.000) & (1.000, \textbf{1.000}, 1.000) & (1.000, \textbf{1.000}, 1.000) \\
\hline
Pentagon & (1.000, \textbf{1.000}, 1.000) & (1.000, \textbf{1.000}, 1.000) & (1.000, \textbf{1.000}, 1.000) \\
\hline
Maze Swap & (1.000, \textbf{1.000}, 1.000) & (0.888, \textbf{0.955}, 1.000) & (0.998, \textbf{0.999}, 1.000) \\
\hline
3D Mazes & (0.864, \textbf{0.897}, 0.951) & (0.851, \textbf{0.880}, 0.921) & (0.397, \textbf{0.526}, 0.661) \\
\hline
4D Mazes & (0.410, \textbf{0.570}, 0.781) & (0.712, \textbf{0.755}, 0.806) & (0.256, \textbf{0.604}, 0.745) \\
\hline %
\end{tabular}
\caption{\label{table:1} $R(\segraph)$ values after 200,000 time-steps of simulation, reported for the Naive, Astute, and Misled agents in each maze as (Q1, \textbf{median}, Q3).}
\end{center}
\end{table}

Next, we were interested in the ability of the ARMS algorithm to recover from major environmental perturbations. To simulate this, we started naive, astute, and misled agents inside of the 5x5(1) maze, then at 100,000 timesteps, swapped to the 5x5(2) maze, and continued the simulation for another 100,000 timesteps. As before, we measured $R(\segraph)$ every 10,000 timesteps, ran 8 trials per condition, and plot median and interquartile range (Figure \ref{fig:results_1}C). It seems that prior knowledge (the astute agent) speeds up early convergence, but it appears that this initial strength comes at a cost: the astute agent recovers from the perturbation more slowly than the naive and misled agents, which both manage to recover to perfect navigability of the new mazes by the end of the simulation (see Table \ref{table:1}). 

If the ARMS algorithm creates too many tiny fields, it can waste it's time on irrelevant details of the environment, while if the ARMS algorithm only creates large fields, it may never actually master the maze. The ARMS algorithm starts with an initial state with a predefined field-size, which is a hyperparameter. The size of this initial field will control to some extent the size of subsequently added fields. How sensitive is the operation of the ARMS algorithm to this initial choice? We tested the naive agent in the 5x5(1) maze across three different initial vertex sizes ($\sigma=0.3$, $\sigma=1.5$, $\sigma=7.5$), measured by the size of their Gaussian activation functions. Running 8 trials per condition for 200,000 timesteps, we find that there is only a weak initial sensitivity to initial field size on performance: in all cases, the ARMS algorithm quickly achieves perfect navigation ability in the maze.

\subsection{Higher dimensional Mazes}

We also wanted to see if the ARMS algorithm could work in higher-dimensional state-spaces. To test this, we generated random 3D and 4D mazes composed of overlapping capsules to make collision-checking simple, with topological complexity equivalent to the 2D mazes we initially tested the ARMS algorithm on. Again, we simulated naive, astute, and misled ARMS agents for 200,000 timesteps, though this time we ran 16 trials per conditions due to the high variability induced by randomizing mazes per-trial. An additional difference is that, due to the computational cost of collision-checking in these mazes, we allowed the agent to ``jump'' towards it's target, stopping at any obstacles along the way. This makes the effective times-scale of the physics a bit faster compared to the 2D mazes.

In smaller mazes (not shown), ARMS handily solves 3 and 4 dimensional mazes, but in 25-node 3D and 4D mazes (in theory of similar complexity to the 25-node 2D mazes that it can master) it begins to struggle. Strangely, the relationship between prior knowledge and ability begins to become less logical. In 3D mazes, it appears that the Naive agent performs better than the Astute agent, with even the Misled agent slightly surpassing the Astute agent by the end of the simulation. In 4D, the situation is even stranger: while the Astute agent performs almost as well in 4D as it did in 3D, the Naive agent's performance totally collapses, with the Misled agent splitting the difference. These patterns suggest that our method of adding prior knowledge to ARMS is conceptually inadequate. We tested an alternative method of choosing when to refine an edge, replacing the stress score $\stress$ with just the number of failures multiplied by a constant, and reran the 3D (Figure \ref{fig:results_1}E) and 4D (Figure \ref{fig:results_1}F) simulations, which yielded overall similar results but a more logical relationship between prior knowledge and maze-navigation ability.

\subsection{MountainCar}
We further tested the ARMS algorithm to see if it could perform well on a standard reinforcement-learning task, continuous MountainCar. Instead of the statespace being 2-spatial dimensions, it is 1-space dimension and 1-velocity dimension. This makes low-level control far less trivial, as it is usually impossible to just ``move towards the goal'', due to the coupling between these two dimensions. We used a PID controller with hand-tuned gains for ARMS. We tested against a strong baseline, Proximal Policy Optimization with Random Network Distillation, equipped with the exact same low-level controller as used by the ARMS algorithm, to keep the baseline fair. This baseline, technically PPO-RND-PID, we refer to simply as ``PPO+''. We found on ordinary MountainCar that both algorithms converged extremely fast, because using a PID controller, it is sufficient to pick a goal to the left, then pick a goal moving to the right in the valley, then pick a goal moving to the right on top of the hill. The presence of the wall means that the exact goal location is irrelevant, so the task becomes fairly trivial.

We made a very simple modification to the ARMS algorithm in order to convert it into an RL algorithm, allowing ARMS to spontaneously switch between an ``exploration'' mode (the default ARMS algorithm) and and ``exploitation'' mode, where the agent selects as its goal the vertex with the highest time-average reward. The decision between exploration and exploitation is made by calculating a weighted measure of the mean-relative reward deviation across all vertices: if the deviation is high (there are some states with \emph{much} higher reward than average), then the agent is more likely to choose to ``exploit'' the segraph, rather than exploring it. Learning still occurs during exploitation, as ARMS is otherwise unchanged, so experience about paths \emph{to} the reward is still being collected. Basically, we convert ARMS into an RL algorithm by simply adding a reward-weighted \emph{bias} to it's exploration goals.

\begin{figure}
\includegraphics[width=\textwidth]{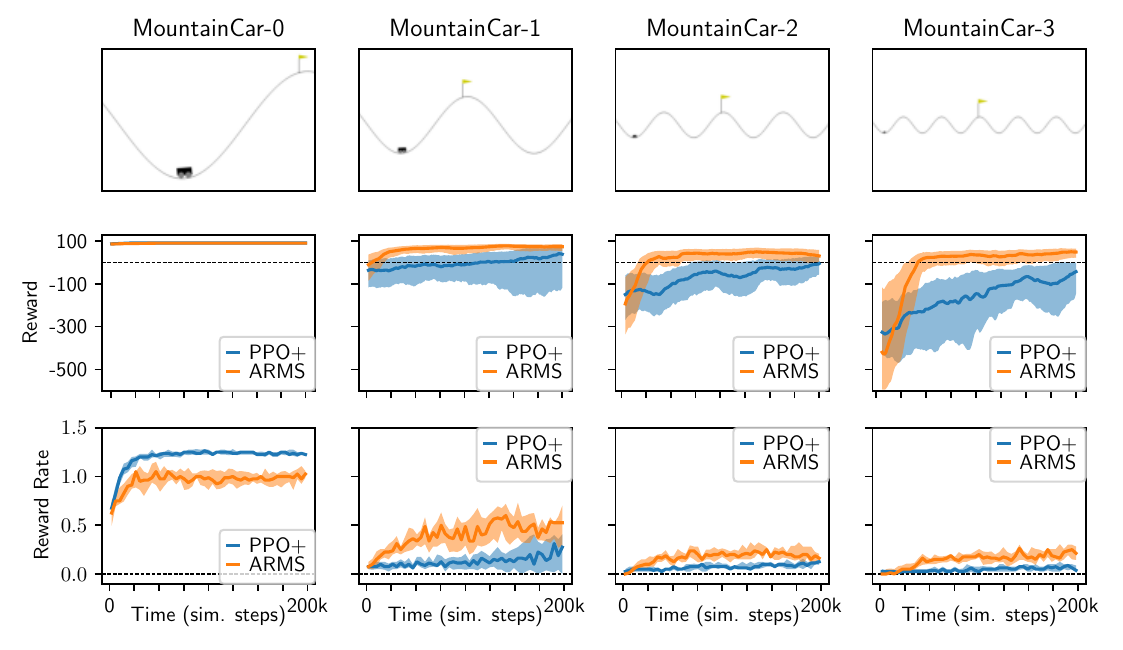}
\caption{ARMS (orange) vs PPO+ (blue) on traditional ContinuousMountainCar (MountainCar-0), and variants with larger environments and stricter reward conditions, MountainCar-1, MountainCar-2, and MountainCar-3. In MountainCar-0, both ARMS and PPO+ instantly solve the problem: PPO+ is actually slighty faster, and achieves a slightly higher performances. For the harder variants, ARMS quickly edges out PPO+. In MountainCar-1, ARMS immediately performs better, and while PPO+ slowly catches up, it never quites matches ARMS within the 200k timesteps of the simulation. The difference is even more dramatic in MountainCar-2 and MountainCar-3, where ARMS rapidly achieves a large positive total reward, indicating that it reliably reaches the target, while PPO+ struggles or fails to even get a net-positive reward. PPO+ is in fact solving the problem in these case, as can be seen by its slowly increasing reward rate in the bottom row of plots.}
\label{fig:rl}
\end{figure}

To more strongly test ARMS, we developed much more difficult variants of the MountainCar environment. Instead of the usual +100 reward being granted for merely crossing the goal position, we converted the goal-\emph{position threshold} to a goal-\emph{radius} in the (position, velocity) statespace, so that the reward is only granted to the agent when it is at the top of the hill \emph{and} is also moving very slowly. We also extended the environment past the first hill, so that it is possible to overshoot the first hill, meaning the agent cannot simply slam into the wall to achieve the goal. We developed three levels of difficult (see top row of Figure \ref{fig:rl}): the first level, MountainCar-1, is as we just described, a valley, then the hill with the reward, then another valley. The second level, MountainCar-2, has three hills with a reward in the middle (only when moving slowly), while the third level, MountainCar-3, has five hills. We did a small hyperparameter search on MountainCar-1 for ARMS, which was used for all environments, whereas for PPO+ we were obliged to do an extensive hyperparameter search for each individual environment. Median total per-episode reward (first and third quartiles shown as shaded region) over 200k time-steps of training are shown for both ARMS (orange) and PPO+ (blue) in the middle row of Figure \ref{fig:rl}.

To confirm that PPO+ was actually learning the more difficult tasks (since for MountainCar-3 reward never goes above 0) we also measured the reward-rate, plotted on the bottom row of Figure \ref{fig:rl}, which shows that PPO+ is actually gradually improving at the task and reaching the reward. 

\section{Discussion}
We have introduced the outlines of a theory of adaptive behavior which ties the ability of a system to robustly solve a problem to the potential of that system to generate any possible behavior, and explained how enumeration of behaviors with sequential modification of a cognitive graph can accomplish this. While the theory has been guided by formal considerations, it was \emph{motivated} by the intuition that, in some sense, universal adaptation should be easy: just don't repeat your mistakes. Formalizing this idea in a way that makes sense in continuous state-spaces with temporally extended actions naturally lends itself to a search paradigm, and forms the meat of this paper. To validate the basic concepts of the theory, we developed the ARMS algorithm as one possible manifestation of the theory, with the addition of more specific biological motivations such as the use of feedback loops to adapt a threshold for edges.

The algorithm can be minimally extended to accommodate rewards to specify a particular task, and performs well against a strong RL baseline. The algorithm makes very minimal assumptions about the underlying statespace, and can use any goal-directed low-level controller, as the algorithm only picks targets for the controller to follow. While our results in 2D environments (2D mazes and MountainCar variants) are very promising, results in higher-dimensional environments suggest that there is room for improvement, especially at the level of algorithm implementation. The strong sensitivity in higher dimensions to prior knowledge suggests that prior knowledge can be very helpful, but it remains an open question what \emph{kinds} of prior knowledge should be incorporated, and perhaps more importantly, \emph{how} they should be incorporated into ARMS. While only preliminary, this work provides the outlines of a system that can exhaustively, but not naively, search a space of continuous behaviors.

Our neural implementation of segraphs suggests that, at least in principle, the mechanisms of the ARMS algorithm could actually be used in the brain, perhaps especially at early stages of learning where trial-and-error dominates. The neural architecture is highly inspired by the functionality of the hippocampus and entorhinal cortex, with segraph vertices playing the role of place cells, vertex-fields playing the role of place-fields, and the underlying HaRK memory system playing the role of grid cells and, more specifically, band cells. The heavy requirement for both active and passive inhibition and disinhibition of edges in our system is interesting, because a population of cells known as \emph{astrocytes}, common in the areas where cognitive graphs are believed to exist, actually project onto and modulate the behavior of individual synapses on both short and long timescales \cite{kofuji2021astrocytes}, perhaps providing the needed degree of control.

\subsection{Related Work}

Our work, proceeding as it does from a theoretical analysis of behavior and problem-solving, touches on a broad range of topics, and ultimately, comes to conclusions that are echoed through the literature on reinforcement learning, problem-solving, and planning. Our specific neural implementation of cognitive graphs creates an overlap of concern, both with neurologically-inspired models of hippocampal learning, and with cognitively-inspired hybrid models. And of course, many different RL algorithms have been developed to handle the issues of sparse rewards, sample efficiency, and lack of resets.

Two prominent continuous statespace planning algorithms for robotics, Rapidly-exploring Random Tress (RRT) \cite{lavalle1998rapidly} and Probabilistic Roadmaps \cite{kavraki1996probabilistic} are interesting to compare to our algorithm. Both algorithms have good performance in high-dimensional statespaces, but both also assume that the problem can be purely represented in terms of a free-space, which can be queried at-will. This setting is quite different from the one we consider, where the only way for the agent to ``query'' the space is to actually physically attempt an action. This actually highlights a potential problem with our formulation of ARMS: because queries cannot be performed freely, but must be \emph{local}, there is a real cost to exploration that is performed inefficiently. It could be the case that more careful control over which edges the agent tries to use could boost performance significantly, even under the constraints of locality.

Planning (search) and learning have been combined quite fruitfully in various RL systems, notably MuZero \cite{schrittwieser2020mastering}, which learns latent dynamics of a problem and performs planning using Monte-Carlo tree search. While MuZero has been successfully applied to continuous domains \cite{hubert2021learning}, the overall algorithm family is significantly different: it uses a learned forward model with Monte Carlo tree search to simulate different futures using sampled actions, with the tree itself being thrown away and rebuilt at every time-step. In this work, we instead grow a dynamic graph that is continuously used and updated, which stores information about the possibility of state-state transitions, with latent dynamics only implicitly represented, and actual actions handled by a low-level controller.

Related more to our usage of a neural graph as a model, the Tolman-Eichenbaum Machine (TEM) \cite{whittington2020tolman} represents one of the most mature models of hippocampal learning. However, they focus on learning good state-representations, compositional inference, and prediction, working in problem domains defined over a static graph. Instead, we dynamically construct a discrete structure (the segraph) \emph{over} a continuous space, using extension and refinement to ensure that any areas or details of the state-space can eventually be represented by the edges of the segraph.

In between pure reinforcement learning and more cognitive-science considerations are works such as \cite{gershman2018successor}, which cast the hippocampus as a reinforcement learning engine that builds successor representations to efficiently decompose tasks in a flexible way. In our work, we adopt a significantly more primitive RL formulation for ARMS, though in-principle it would be easy to propagate values over the segraph-vertices to define a value-function. On the other hand, we emphasize the potential for cognitive graphs to be used for rapid adaptation and problem solving, which is not the focus of their work. As it so happens, hippocampus contains both place cells with firing fields that stabilize with experience (more suggestive of gradually consolidated successor representations), and place cells with unstable firing fields that are ``freshly'' recruited even in the same contexts \cite{ziv2013long}, perhaps indicative of something approximated by our proposal.

\subsection{Current Limitations}
Our work makes several highly limiting assumptions: we assume the statespace is Euclidean, that it is fully visible (there is no notion here of a hidden ``latent space''), there is no noise, and we provide no mechanism for learning a more powerful low-level controller, if one is not already provided. Furthermore, due to the difficulty of constructing appropriate low-level controllers, we only tested on fairly simple environments, such as spatiotemporally continuous mazes and the MountainCar environment variants.

The use of hyperball vertex-fields is a natural consequence of our construction of robust behaviors, but place-fields in the hippocampus are often non-circular, with fields along hallways often elongated. Under many circumstances, it is likely that radially symmetric fields do not provide the optimal decomposition of the statespace. Additionally, our mechanism for adding prior knowledge to the ARMS algorithm seems to be inadequate at best or even incoherent at worst, especially when applied to higher-dimensional statespaces. Our suspicion is that our mechanism of thresholding by affordance contains a strong dimensionality dependency that is currently unmodeled, making our choice of affordance-threshold adaptation hyperparameters critical. We also don't provide any mechanism for consolidating the graphs constructed by ARMS into either implicit procedural knowledge or explicit semantic knowledge that could be reused in the future.

We also used a fairly naive pathfinding algorithm, which being unweighted cannot take into account costs along the path. In the tasks that we tested on, this was not a major barrier to the success of the algorithm, but in more sophisticated tasks, or tasks with different reward structure, incorporating path costs could be critical. Additionally, planning occurs over a ``flat'' graph: there is no mechanism in the planning algorithm for taking advantage of clustering or hierarchical structure in the graph itself.

\subsection{Future Extensions}
The aforementioned limitations provide a natural focus for future growth. In order to tackle more complex tasks with hidden stochastic dynamics, it might be sufficient to use one of several latent-space estimation techniques such as variational or denoising autoencoders to construct the statespace $\SS$ used by the ARMS algorithm. This may also allow for the representation of more complex manifolds than simply a flat Euclidean geometry. Additionally, in experiments we performed but did not include in this work, it is apparent that higher-quality controllers can be trained through self-supervision by executions over the graph. Using mechanisms of exploration for the low-level controller, well-studied in the RL literature, it could be possible to bootstrap better low-level controllers.

Regarding radially symmetric vertex-fields, this point is relatively easy to expand upon: the underlying HaRK memory makes it trivial to construct arbitrary and even disconnected vertex-fields. The problem becomes 1) what should actually control the shape of vertex fields, and 2) how can our agent sample waypoints from arbitrarily-shaped fields? It may be the case that some kind of coupled perceptual-generative network could fill the role of defining and sampling from vertex-fields with more complex geometries. Using such a network to guess good initial configurations of vertex-fields, as well as estimate the reliability of transitions between vertex-fields, could enable application of prior knowledge without compromising the algorithm's robustness.

While we used a custom pathfinding algorithm that takes advantage of the one-hot encoding of the segraph for parallel pathfinding, in-principle, any pathfinding algorithm could be used, including algorithms that incorporate edge-weights such as time, energy, or reward. One possibility that we are currently considering, is whether the midpoints recursively found by the pathfinding algorithm could somehow be used to ``cache'' previous planning results, perhaps constructing a hierarchical representation of the segraph's structure that can somehow speed up planning.

One idea that we would like to emphasize, in closing, is that intelligently-ordering an exhaustive search over behaviors is a general-purpose problem-solving strategy, which can be complete \emph{without} being naive. Here, we have performed an extremely simple kind of search: the space of behaviors has been restricted to those generated by a low-level-controller, ``states'' are mere hyperballs, the pathfinding algorithm is fixed, and the mechanisms for ordering behaviors have been hard-coded into the algorithm. In-principle, however, it would be possible to incorporate considerably more intelligence into the \emph{ordering} of behaviors, perhaps if the ARMS algorithm was replaced with an RL algorithm that performed the same function, a much more flexible algorithm could be created.

In this work, we used the ``affordance'' of an edge to induce an order over paths, in the future, we hope that the theory can be extended to inducing orders over paths using arbitrary functions, including the use of neural networks. If a general theory could be developed for how to do this, then it would be possible to design hyperadaptable agents with arbitrary priors, allowing for the incorporation of both expert and learned knowledge. An analysis of the set of all possible orders (corresponding to different biases or priors) could be extremely interesting and fruitful. One possibility we are particularly interested in studying is ordering paths by potential \emph{empowerment} \cite{klyubin2005all}.

Much of current deep reinforcement learning research is, indirectly, inspired by older work on habit-formation. Technically, we could think of ``problem-solving'' as a kind of very sophisticated habit, but we think that perhaps exploring the opposite perspective is worthwhile: if we equate ``problem-solving'' with ``search'', maybe a \emph{habit} is just a cached \emph{search result}. To the extent that model-free RL can handle old situations, and model-based RL is needed for new ones, this framing dissolves the dichotomy, as ``model-free'' behavior is just what happens when the first result of behavioral search is correct, whereas ``model-based'' behavior occurs when subsequent search steps are required. We leave to future work the development of a more general theory of behavioral search, in which prior knowledge merely re-orders behaviors in the infinite space of possibilities.

\appendix
\section{Notation Glossary}\label{sec:notation_glossary}
\begin{notation}
\entry{$\agent$}{An agent} \notsep
\entry{$\SS$}{A Euclidean \textbf{statespace}} \notsep
\entry{$\SS_0$}{The subset of $\SS$ physically accessible from the agent's starting state} \notsep
\entry{$\s, \goal{\s}$}{A state in $\SS$, and a goal-state in $\SS$ (together, a \textbf{problem})} \notsep
\entry{$\K$}{A statefull, goal-oriented \textbf{controller} that tries to reach a goal-state} \notsep
\entry{$\Ks(\vect{h}_t)$}{Controller \textbf{success} ($0$ for not yet, $1$ for success)} \notsep
\entry{$\Kf(\vect{h}_t)$}{Controller \textbf{failure} ($0$ for not yet, $1$ for failure)} \notsep
\entry{$\termtime{\s}{\goal{\s}}$}{Termination time (either $\Ks(\vect{h}_t)=1$ or $\Kf(\vect{h}_t)=1$) for controller $\K$ trying to reach $\goal{\s}$ from $\s$} \notsep
\entry{$\trajK(\s, \goal{\s})$}{Finite physical \textbf{trajectory} generated by controller $\K$ going from $\s$ to $\goal{\s}$} \notsep
\entry{$\rK(\s, \goal{\s})$}{\textbf{Reachability} for controller $\K$ from $\s$ to $\goal{\s}$ (failure=$0$, success=$1$)} \notsep
\entry{$\conK(\s)$}{The points reachable by controller $\K$ from state $\s$} \notsep
\entry{$\preK(\s)$}{The points that can be reached from state $\s$ using controller $\K$} \notsep
\entry{$\plan$}{A \textbf{plan}, or sequence of states (waypoints)} \notsep
\entry{$\trajK(\plan)$}{Finite physical \textbf{trajectory} generated by controller $\K$ following plan $\plan$} \notsep
\entry{$\rK(\plan)$}{\textbf{Reachability} for controller $\K$ following plan $\plan$ (failure=$0$, success=$1$)} \notsep
\entry{$\ball{\s}{\epsilon}$}{A hyperball around $\s$ with radius $\epsilon$, $\ball{\s}{\epsilon} = \{\s' \in \SS : \norm{s' - s} \leq \epsilon\}$} \notsep
\entry{$\splans[][]{}{}$}{The \textbf{set of all plans}. This symbol admits many variations. The set of plans of length $n$ is $\splans[][]{}{n}$, of any length, $\splans[][]{}{*}$. The set of plans with waypoints in the set $A$ is $\splans[][]{A}{}$. The set of robust plans is $\splans[r][]{}{}$. The set of plans that start at $\s$ and end at $\s'$ is $\splans[][]{}{}(\s, \s')$} \notsep
\entry{$\tube{\eps}(\plan)$}{The radius-$\eps$ tube around plan $\plan$ of alternative plans} \notsep
\entry{$\stube{\eps}(\plan)$}{The subset of $\tube{\eps}(\plan)$ that is successful} \notsep
\entry{$\rKe{\eps}(\plan)$}{The fraction of $\tube{\eps}(\plan)$ that succeeds, if $\rKe{\eps}(\plan)=1$ for $\eps>0$ then $\plan$ is \textbf{robust}} \notsep
\entry{$\SSK$}{The subset of statespace reachable using the controller $\K$} \notsep
\entry{$\plan_1 \sim \plan_2$}{Plans are equivalent if they are both solutions to the same problem, starting at the same point $\start{\s}$, ending at the same point $\goal{\s}$, and succeeding, meaning $\rK(\plan_1) = 1$ and $\rK(\plan_2) = 1$. Equivalently, $\plan_1, \plan_2 \in \splansf[r][]{}{*}{\start{\s}, \goal{\s}}$} \notsep
\entry{$\achclass{\SSK}{\s}{\s'}$}{Equivalence class of plans from $\s$ to $\s'$ with waypoints in $\SSK$, equal to $\splansf[r][]{\SSK}{*}{\s, \s'}$} \notsep
\entry{$\achset{}{\SSK}$}{The set $\splans[r][]{\SSK}{*}$ partitioned by $\sim$, i.e. the \textbf{set of solveable problems} in $\SSK$, or the set of all equivalence classes of plans with waypoints in $\SSK$} \notsep
\entry{$\achset{\agent}{\SSK}$}{The set of equivalence classes of solutions with waypoints in $\SSK$ that the agent $\agent$ can generate} \notsep
\entry{$\WP$}{A discrete set of waypoints in the statespace $\SS$} \notsep
\entry{$E$}{A function assigning an $\epsilon$ to each waypoint in $\WP$} \notsep
\entry{$B_E$}{A function assigning a hyperball to each waypoint in $\WP$ with radius given by $E$} \notsep
\entry{$\hball_{\s}$}{The hyperball assigned to the waypoint $\s$} \notsep
\entry{$\hballs$}{A set of hyperballs. $\hballs(\WP)$ are the hyperballs corresponding to points in $\WP$} \notsep
\entry{$\bath$}{A sequence of hyperballs or \textbf{path}. If $\plan$ is a plan drawn from $\splans[][]{\WP}{*}$, then $\bath_{\plan}$ is the corresponding sequence of hyperballs} \notsep
\entry{$\product{\bath}$}{The bundle of plans ($\eps$-tube) represented by the hyperball sequence $\bath$} \notsep
\entry{$\rK(\bath)$}{The $\eps_{\bath}$-reliability of $\bath$, where $\eps_{\bath}$ is the vector of corresonding hyperball radii. $\bath$ is \textbf{robust} if $\rK(\bath)=1$} \notsep
\entry{$\refine$}{The \textbf{refinement mutation}, replaces a hyperball with several smaller hyperballs in the same area, increasing the local resolution of $\hballs(\WP)$} \notsep
\entry{$\allbaths[][]{}{}$}{The \textbf{set of all paths}. This symbol admits many variations. The set of paths of length $n$ is $\allbaths[][]{}{n}$, of any length, $\allbaths[][]{}{*}$. The set of paths with hyperballs in the set $A$ is $\allbaths[][]{A}{}$, all paths that are robust is $\allbaths[r][]{}{}$. All paths from $\hball_1$ to $\hball_2$ is $\allbaths[][]{}{}(\hball_1, \hball_2)$} \notsep
\entry{$\bath_1 \sim \bath_2$}{Paths are equivalent if they are both robust, start at the same hyperball, and end at the same hyperball. Equivalently, $\bath_1, \bath_2 \in \allbaths[r][]{}{*}(\hball, \hball')$} \notsep
\entry{$\achclass{\hballs}{\hball}{\hball'}$}{Equivalence class of paths from $\hball$ to $\hball'$ with hyperballs in $\hballs$, equal to $\allbaths[r][]{\hballs}{*}(\hball, \hball')$} \notsep
\entry{$\achset{\hballs}{\hballs}$}{The set $\allbaths[r][]{\hballs}{*}$ partitioned by $\sim$, i.e. the set of all equivalence classes of \emph{paths} with hyperballs in $\hballs$} \notsep
\entry{$\achset{\hballs}{\SSK}$}{The set of equivalence classes of \emph{plans} with members inside a path in $\allbaths[r][]{\hballs}{*}$} \notsep
\entry{$\extend$}{The \textbf{extension mutation}, adds new larger hyperballs around an existing hyperball, increasing the area of statespace covered by $\hballs(\WP)$} \notsep
\entry{$\segraph=(\segverts, \segedges)$}{A \textbf{segraph}, a graph with directed edges $\segedges$ that connect vertices $\segverts$ that segment statespace, which is used to organize the enumeration of plans} \notsep
\entry{$\vertex_i$}{A segraph \textbf{vertex} $\vertex_i$ representing a ``chunk'' of state-space, a corresponding hyperball $\hball_i \subset \SS$} \notsep
\entry{$\edge_{i,j}$}{The directed segraph \textbf{edge} between vertices $\vertex_i$ and $\vertex_j$, representing a chunk of connection space $\connection_{i, j} = \hball_i \times \hball_j \subset \SS^2$} \notsep
\entry{$\measure$}{The \textbf{measure} of a vertex or edge, $\measure(\vertex_i) = \abs{\hball_i}$, $\measure(\edge_{i, j}) = \abs{\connection_{i,j}}$} \notsep
\entry{$\rK(\edge_{i,j})$}{The theoretical \textbf{reliability} of the edge $\edge_{i, j}$, equal to $\rK((\hball_i, \hball_j))$} \notsep
\entry{$n_s(\edge)$}{The number of \textbf{successful traversals} by $\agent$ over $\edge$} \notsep
\entry{$n_f(\edge)$}{The number of \textbf{failed traversals} by $\agent$ over $\edge$} \notsep
\entry{$n(\edge)$}{The total number of \textbf{traversals} by $\agent$ over $\edge$, equal to $n_s(\edge) + n_f(\edge)$} \notsep
\entry{$\erK(\edge)$}{The empirically \textbf{estimated reliability} of $\edge$} \notsep
\entry{$\p$}{A \textbf{path}, a sequence of unique vertices} \notsep
\entry{$\allpaths[][]{}{}$}{The \textbf{set of all paths}. This symbol admits many variations. The set of paths of length $n$ is $\allpaths[][]{}{n}$, of any length, $\allpaths[][]{}{*}$. The set of paths with vertices in the set $A$ is $\allpaths[][]{A}{}$. The set of robust paths is $\allpaths[r][]{}{}$. The set of paths that start at $\vertex$ and end at $\vertex'$ is $\allpaths[][]{}{}(\vertex, \vertex')$} \notsep
\entry{$\goalPsym$}{The probability distribution over goals used to represent the way an agent selects problems to try to solve} \notsep
\entry{$\pathPsym$}{The probability distribution over paths used to represent the way an agent selects plans to try to solve a problem} \notsep
\entry{$\rK(\p)$}{The reliability of a path, equal to the reliability of the corresponding hyperball sequence. When $\rK(\p)=1$, the path is \textbf{robust}} \notsep
\entry{$\achclass{\segraph}{\vertex}{\vertex'}$}{The equivalence class of robust paths over $\segraph$ between $\vertex$ and $\vertex'$} \notsep
\entry{$\achset{\segraph}{\segverts}$}{The set of all equivalence classes of robust paths over $\segraph$} \notsep
\entry{$\try(\p)$}{A function modeling the action of an agent \textbf{trying to follow a path}} \notsep
\entry{$\efailed{\segedges}$}{The set of edges that fail when an agent tries to follow a path. If $\efailed{\segedges}=\varnothing$, then the path succeeds} \notsep
\entry{$\eforcedinactive{\segedges}$}{The set of edges that are inhibited (excluded from pathfinding by the agent) due to having failed while trying paths} \notsep
\entry{$\eactive{\segedges}$}{The set of edges that are not inhibited and can be used in pathfinding} \notsep
\entry{$\potential(\p)$}{The total affordance provided by a path $\p$} \notsep
\entry{$\q(\edge)$}{The utility of an edge $\edge$ in terms of the incremental affordance it provides} \notsep
\entry{$\q_{\text{ref}}(\edge)$}{The total utility of the edges produced by refining $\edge$} \notsep
\entry{$\Delta q(\edge)$}{The theoretical change in utility resulting from refining $\edge$} \notsep
\entry{$\stress(\edge)$}{An empirical estimate of $\Delta q(\edge)$} \notsep
\entry{$\potentialbar$}{An affordance-threshold used to regulate the order in which paths are tried}
\entry{$\potential(\edge)$}{The affordance of the edge $\edge$, $\potential(\edge) = \rK(\edge) \cdot \measure(\edge)$} \notsep
\entry{$\epotential(\edge)$}{The estimated affordance of the edge $\edge$, $\epotential(\edge) = \erK(\edge) \cdot \measure(\edge)$} \notsep
\entry{$\segElt{\potentialbar}$}{The set of edges that are below the affordance threshold $\potentialbar$} \notsep
\entry{$\segEgte{\potentialbar}$}{The set of edges that are above the affordance threshold $\potentialbar$} \notsep 
\entry{$\allpathsf[][]{\raisemath{0.1ex}{\segraph}}{}{\vertex, \edge}$}{The set of all paths over $\segraph$ that start at $\vertex$ and end by crossing $\edge$} \notsep
\entry{$\pathP{\segraph}{\vertex}{\edge}$}{The probability distribution of paths from $\vertex$ to $\edge$ over $\segraph$} \notsep
\entry{$\epistemicscore(\edge)$}{The epistemic value of an edge $\edge$, used to weigh the probability that $\edge$ is chosen as a goal, modeled by the distribution $\goalP$ over edges}
\end{notation}

\section{Detailed Methods}
We cover some technical details that important for understanding the operation of the ARMS algorithm, as well as its neural implementation.

\subsection{Extension and Prior Knowledge} \label{meth:extPKNA}
Extension adds new vertices around an existing vertex so as to increase the coverage of $\segraph$. During extension, the agent randomly samples points $\vect{p}_j$ around the exterior of the field of the vertex $\vertex$. If the agent can estimate the reachability $r_j$ or be provided with it, then it can store that information using HaRKs in a matrix $\mat{R} = \sum_j\vect{x}_{\vect{p}_j} \bind r_j$, along with a normalization matrix $\mat{C} = \sum_j\vect{x}_{\vect{p}_j} \bind 1$. If the agent is Naive, then $r_j=1$ regardless of the actual reachability (we could also call the Naive agent ``optimistic'').

Then, the agent tries to add new fields. Suppose the size of the field $\field(\vertex)$ is $\sigma$, then the agent starts by trying to add new fields with a size of $\sigma' = \sqrt{2}\sigma$. It does this by checking the value $\hat{r}_j = \frac{\mat{R}(\vect{x}_{\vect{p}_j} \odot \vect{\eta}_{\sigma'})}{\mat{C}(\vect{x}_{\vect{p}_j} \odot \vect{\eta}_{\sigma'})}$, where $\hat{r}_j$ is the local ``average'' of reachability, with ``local'' defined by the scale of $\sigma'$. All $\vect{p}_j$ are evaluated and ordered from highest to lowest $\hat{r}_j$, and the points with the highest $\hat{r}_j$ (above a certain threshold) are added as new vertex with fields of size $\sigma'$ (new vertices fields that are too redundant with older vertices are not added).

After doing this check for $\sigma' = \sqrt{2}\sigma$, the agent does this check for $\sigma' = \sigma$ and $\sigma' = \frac{\sqrt{2}}{2}\sigma$. For the last check, it doesn't matter if $\tilde{r}_j$ is above the pre-defined threshold, the space surrounding $\field(\vertex)$ is filled with new vertices of field-size $\frac{\sqrt{2}}{2}\sigma$. The ``Astute'' agent uses a direct collision detection calculation to determine $r_j$, while the ``Misled'' agent uses $1 - \hat{r}_j$ instead of $\hat{r}_j$, effectively inverting the judgments of the ``Astute'' agent.

\subsection{ARMS $\potentialbar$ Regulation} \label{meth:lambda}
Internally, information about inhibition dynamics is stored in a tuple $\Lambda = (\potentialbar, \potentialbar_0, \inhbpos, \inhbneg, \APFC)$, with three main functions that modify $\Lambda$, $\ftaua$, $\ftaub$, and $\ftauc$. Referring back to the ARMS flowchart in Fig. \ref{fig:flowchart}, there are three points where $\Lambda$ is updated, those being steps \itemP{} (where $\ftaua$ is called on the edge that just failed), \itemE{} (where $\ftaub$ is called after a path has been found), and \itemR{} (where $\ftauc$ is called on the edge with the highest $\potential$ that is inhibited after a pathing failure). \newline{}

\noindent
\begin{minipage}[t]{.5\textwidth}
\begin{algorithmic}
\ttfamily
\Function{$d\potentialbar$}{$\potentialbar$, $\Delta$}
	\State $n \sim U([0, 2])$
	\State $\potentialbar \leftarrow \potentialbar + n \cdot \Delta$
	\If{$\potentialbar < 0$}
		\State $\potentialbar \leftarrow 0$
	\EndIf
	\State \Return $\potentialbar$
\EndFunction
\end{algorithmic}
\end{minipage}%
\begin{minipage}[t]{.5\textwidth}
\begin{algorithmic}
\ttfamily
\Function{$\ftaua$}{$\Lambda$, $\edge$}
	\State $(\potentialbar_0, \potentialbar, \inhbpos, \inhbneg, \APFC) \leftarrow \Lambda$
	\State $\Delta\potentialbar \leftarrow \potential(\edge) - \potentialbar$
	\State $\inhbpos \leftarrow (1-\beta_r) \cdot \inhbpos + \beta_r\cdot\lambda_{\tau}^+\cdot\max(0, \Delta\potentialbar)$
	\State $\potentialbar \leftarrow d\potentialbar(\potentialbar, \Delta^+)$
	\State $\Lambda \leftarrow (\potentialbar_0, \potentialbar, \inhbpos, \inhbneg, \APFC)$
	\State \Return $\Lambda$
\EndFunction
\end{algorithmic}
\end{minipage}\newline{}\bigskip{}

\noindent
\begin{minipage}[t]{.5\textwidth}
\begin{algorithmic}
\ttfamily
\Function{$\ftaub$}{$\Lambda$}
	\State $(\potentialbar_0, \potentialbar, \inhbpos, \inhbneg, \APFC) \leftarrow \Lambda$
	\If{$\APFC > 0$}
		\State $\Delta\potentialbar \leftarrow \potentialbar_0 - \potentialbar$
		\State $\inhbneg \leftarrow (1 - \beta_r) \cdot \inhbneg + \beta_r \cdot \lambda_{\tau}^- \cdot \Delta\potentialbar$
		\State $\APFC \leftarrow 0$
	\EndIf
	\State $\Lambda \leftarrow (\potentialbar_0, \potentialbar, \inhbpos, \inhbneg, \APFC)$
	\State \Return $\Lambda$
\EndFunction
\end{algorithmic}
\end{minipage}%
\begin{minipage}[t]{.5\textwidth}
\begin{algorithmic}
\ttfamily
\Function{$\ftauc$}{$\Lambda$, $\edge$}
	\State $(\potentialbar_0, \potentialbar, \inhbpos, \inhbneg, \APFC) \leftarrow \Lambda$
	\If{$\APFC = 0$}
		\State $\potentialbar_0 \leftarrow \potentialbar$
	\EndIf
	\State $\APFC \leftarrow \APFC + 1$
	\State $\potentialbar \leftarrow \min(\potentialbar, \potential(\edge))$
	\State $\potentialbar \leftarrow d\potentialbar(\potentialbar, -\inhbneg)$
	\State $\Lambda \leftarrow (\potentialbar_0, \potentialbar, \inhbpos, \inhbneg, \APFC)$
	\State \Return $\Lambda$
\EndFunction
\end{algorithmic}
\end{minipage}

\subsection{Derivation of $\Delta q$} \label{meth:deltaQ}
While ``refinement'' is an operation that occurs to an individual \emph{vertex}, the relevant information about when to refine a vertex actually exists on an \emph{edge}, as we are ultimately interested in increasing the resolution of the corresponding $\eps$-tubes. Henceforth, when we refer to ``refinement'', we will be talking about an operation that occurs \emph{on an edge}, with the decision to ``refine an edge'' resulting in the refinement of \emph{one of its vertices}. To understand when our agent should refine an edge, we need to understand what exactly the agent gains through refinement. Technically, the overall utility of an edge is related to how helpful the edge is for getting the agent to the rest of the environment... a global measure that depends on all the paths that pass through that edge, and their reliability. We can quantify this as:
\begin{equation*}
\q(\edge) = \sum_{\mathclap{\p \in \paths_{\segraph}(\edge)}}{p(\p)\cdot\potential(\p)}
\end{equation*}
where $\paths_{\segraph}(\edge)$ is the set of all paths containing $\edge$, $p(\p)$ is the probability that $\p$ is sampled, and $\potential(\p)$ is a generalization of potential affordance to paths, with $\potential(\p) = \rK(\p) \cdot \measure(\p\idx{0}) \cdot \measure(\p\idx{\n{1}})$. To make $\q(\edge)$ tractable to calculate, we have to make some assumptions. First, we assume that $p(\p) \propto \rK(\edge)$, with an unknown proportionality constant $c_p$. Second, we assume that $\rK(\p) \propto \rK(\edge)$, with a proportionality constant of $c_r$. Third, we assume that $\measure(\p\idx{0})$ and $\measure(\p\idx{\n{1}})$ are independent of each other and of $\edge$, so we factor them out into their average across the whole graph, denoted $c_F$. Fourth, we assume that the number of paths including $\edge$ is proportional to the number of traversals over that edge, $n(\edge)$, giving us:
\begin{align*}
\q(\edge) &= \sum_{\mathclap{\p \in \paths_{\segraph}(\edge)}}{p(\p)\cdot\potential(\p)} = \sum_{\mathclap{\p \in \paths_{\segraph}(\edge)}}{c_p \rK(\edge)\cdot \rK(\p) \cdot \measure(\p\idx{0}) \cdot \measure(\p\idx{\n{1}})} \\
&= \sum_{\mathclap{\p \in \paths_{\segraph}(\edge)}}{c_p\rK(\edge) \cdot c_r\rK(\edge) \cdot c_F} = c_p c_r c_F \cdot \sum_{\mathclap{\p \in \paths_{\segraph}(\edge)}}{\rK(\edge)^2} = c_q \cdot n(\edge) \cdot \rK(\edge)^2
\end{align*}
where $c_q = c_p c_r c_F$. Assuming that refining the edge will split it into $k$ smaller edges $\{\edge'_1, ... \edge'_k\}$ with disjoint fields, the total ``utility'' of these refined edges would be:
\begin{equation*}
\q_{\text{ref}}(\edge) = \sum_{i=1}^k{q(\edge'_i)} = \sum_{i=1}^k{c_qn(\edge'_i) \rK(\edge'_i)^2}
\end{equation*}
We assume that each ``child''-edge $\edge'_i$ is in approximately $\frac{1}{k}$th as many paths as $\edge$, and that $\rK(\edge'_i) \sim \text{B}(c \cdot \rK(\edge), c \cdot (1 - \rK(\edge)))$, the beta distribution re-parameterized with $\rK(\edge)$ the average of the distribution, $c$ a parameter encoding the intrinsic properties of the statespace. Taking $k\rightarrow\infty$ for analytical tractability, we get:
\begin{align*}
q_{\text{ref}}(\edge) &= \sum_{i=1}^k{c_q n(\edge'_i) \rK(\edge'_i)^2} = \sum_{i=1}^k{c_q \frac{n(\edge)}{k} \rK(\edge'_i)^2} = c_q \cdot n(\edge) \cdot \frac{1}{k} \sum_{i=1}^k{\rK(\edge'_i)^2} \\
&= c_q \cdot n(\edge) \cdot \expect{\rK(\edge'_i)^2} = c_q \cdot n(\edge) \cdot \left(\frac{\rK(\edge)(1 - \rK(\edge))}{c + 1} + \rK(\edge)^2\right)
\end{align*}
The difference between these two terms, $q(\edge)$ and $q_{\text{ref}}(\edge)$, gives the expected utility of refining an edge:
\begin{align*}
\stress(\edge) &= q_{\text{ref}}(\edge) - q(\edge) = \left[c_q \cdot n(\edge) \cdot \left(\frac{\rK(\edge)(1 - \rK(\edge))}{c + 1} + \rK(\edge)^2\right)\right] - \left[c_q \cdot n(\edge) \cdot \rK(\edge)^2\right] \\
&= c_q \cdot n(\edge) \cdot \left(\frac{\rK(\edge)(1 - \rK(\edge))}{c + 1} + \rK(\edge)^2 - \rK(\edge)^2\right) = c_{\stress} n(\edge) \rK(\edge) (1 - \rK(\edge))
\end{align*}
where $c_{\stress}$ folds the unknown value $c+1$ in with the also unknown $c_q$. One small caveat: our agent has to estimate $\rK(\edge)$ as $\erK(\edge)$, but pseudo-counts in this case are unhelpful... to avoid them, we re-arrange terms:
\begin{equation}
\stress(\edge) = c_{\stress}n(\edge)\frac{n_s(\edge)}{n(\edge)}\frac{n_f(\edge)}{n(\edge)} = c_{\stress}\frac{n_s(\edge) n_f(\edge)}{n(\edge)} \approx c_{\stress}\frac{n_s(\edge) n_f(\edge)}{n(\edge) + \epsilon}
\end{equation}
with $\epsilon$ some constant for numerical stability (we arbitrarily pick $\epsilon = 1$).

This score, an indicator of how useful it would be to refine an edge, is actually quite intuitive. An edge that is never used (or has never been used) is unlikely to benefit from refinement. If an edge is completely reliable \emph{or} completely unreliable, then $\stress(\edge)=0$ and it is useless to refine the edge. If the edge is of intermediate reliability, then $\stress(\edge)>0$, because refining the edge creates the possibility that some child-edges will be \emph{more} reliable than the parent, while others less: these lesser children can be filtered out (inhibited), allowing our agent to separate the wheat from the chafe, so to speak.

\subsection{Unobserved vertices and Vertex deletion} \label{meth:ghost}
A state that has never been entered by the agent has a special status in the ARMS system. Any edge $\edge_{i,j} = (\vertex_i, \vertex_j)$ for which $\vertex_i$ has not been visited is considered a \emph{ghost edge}. Such an edge is considered to have zero epistemic value, and is inhibited until $\vertex_i$ is visited. Not maintaining this ``ghost/non-ghost'' distinction can cause the agent to fixate on impossible-to-reach goals.

Once the agent has reached the maximum number of vertices that it's memory-matrix can allow, to keep progressing it needs a way to delete useless vertices. To determine this, we developed an ad-hoc score, which incorporates the fraction of edges that a vertex $\vertex$ has which are ``ghosts'', $F_{\text{ghost}}(\vertex)$. This score is computed as:
\begin{equation*}
g(\vertex) = -\frac{(1 - F_{\text{ghost}}(\vertex)) (n_{\text{transits}}(\vertex) + 1) (n_{\text{visits}}(\vertex) + 1) (\measure(\vertex) + 1)}{\sqrt{T_{\text{last}}(\vertex)} + 1}
\end{equation*}
where $n_{\text{transits}}(\vertex)$ is the number of times the agent successfully entered and then subsequently left $\vertex$, $n_{\text{visits}}(\vertex)$ is the number of times $\vertex$ has been visited, and $T_{\text{last}}(\vertex)$ is the time since $\vertex$ was last visited. When the agent wants to add a new vertex but no space is left, it deletes the vertex which has the maximum $g$.

\subsection{Pathfinding Algorithm} \label{meth:pathing}
The $\pwave$ pathfinding algorithm is based on recursive problem decomposition via wave-based midpoint finding. The algorithm can be described by the following functions:\newline{}\\
\bigskip{}
\noindent{}\begin{minipage}[t]{.5\textwidth}
\begin{algorithmic}
\ttfamily
\Function{FIND\_MID}{$\vect{m}_s$, $\vect{m}_g$, $\mat{G}$}
	\State $\vect{f}_s \leftarrow \vect{m}_s$, $\vect{f}_g \leftarrow \vect{m}_g$
	\While{True}
		\State $\vect{f}_s \leftarrow f_{\mathrm{prop}}(\mat{G}, \vect{f}_s)$
		\If{$\vect{f}_s \odot \vect{f}_g \neq \vect{0}$}
			\State \Return $\vect{f}_s \odot \vect{f}_g$
		\EndIf
		\State $\vect{f}_g \leftarrow f_{\mathrm{prop}}(\mat{G}, \vect{f}_g)$
		\If{$\vect{f}_s \odot \vect{f}_g \neq \vect{0}$}
			\State \Return $\vect{f}_s \odot \vect{f}_g$
		\EndIf
		\If{$\vect{f}_s = \vect{0}$ OR $\vect{f}_g = \vect{0}$}
			\State \Return $\vect{0}$
		\EndIf
	\EndWhile
\EndFunction
\end{algorithmic}
\end{minipage}%
\begin{minipage}[t]{.5\textwidth}
\begin{algorithmic}
\ttfamily
\Function{$\pwave$}{$\vect{m}_s'$, $\vect{m}_g'$, $\mat{G}$}
	\State stack $\leftarrow$ [], $\p \leftarrow$ []
	\State $\vect{m}_s \leftarrow \vect{m}_s'$, $\vect{m}_g \leftarrow \vect{m}_g'$
	\While{True}
		\State $\vect{m} \leftarrow$ FIND\_MID($\vect{m}_s$, $\vect{m}_g$, $\mat{G}$)
		\If{$\vect{m} = \vect{0}$}
			\State \Return None
		\EndIf
		\If{$\vect{m} \odot \vect{m}_g = \vect{0}$}
			\State stack.push($\vect{m}_g$)
			\State $\vect{m}_g \leftarrow \vect{m}$
		\Else
			\State $\vect{p} \leftarrow$ RANDOM($\vect{m} \odot \vect{m}_s$)
			\State $\p$.append($\vect{p}$)
			\State $\vect{m}_s \leftarrow \vect{p}$
			\State $\vect{m}_g \leftarrow$ stack.pop()
			\If{$\vect{p} \odot \vect{m}_g' \neq \vect{0}$}
				\State \Return $\p$
			\EndIf
		\EndIf{}
	\EndWhile
\EndFunction
\end{algorithmic}
\end{minipage}\bigskip

The \texttt{RANDOM} function returns a random one-hot vector from a multi-hot vector, and corresponds to random selection of a vertex from a set of vertices.

\subsection{Maze Generation} \label{meth:maze}
We use two different kinds of random mazes, the first are mazes laid out on a rectilinear lattice, and the second are randomly distributed within the boundaries of a specific polygon. In both cases, we generate a set of ``node'' points (on a grid for rectilinear mazes, and randomly but evenly distributed using Lloyd's algorithm \cite{lloyd1982least} for polygonal mazes) for which we compute the Delaunay triangulation, forming a planar graph. The topology of the maze is determined by randomly selecting a spanning tree of this graph. The interior of the maze is just determined by the geometric union of many ``hallway'' polygons corresponding to edges of this graph, with a small amount of ``smoothing'' to reduce the complexity of the polygon to speed up collision-detection.

\subsection{Estimating Graph Reliability $R(\segraph)$} \label{meth:r_of_g}
In the text, we say that the reliability of a segraph $R(\segraph)$ measures the ability of the agent to go from any point in a statespace to any other point in the statespace. Technically, this is not true for two reasons. First of all, we restrict ourselves to subset of statespace that is physically \emph{reachable} by the agent, meaning we only include points that are inside of the maze. Second of all, there are a continuum of points within the maze, so measuring the reliability for all points is computationally impossible. We compromise by selecting ``special'' points in the maze, which correspond to the ``node'' points used for constructing the maze in the first place.

So, consider two node points from the maze, $p_1$ and $p_2$. We map these points to vertices of $\segraph$, $\vertex_1$ and $\vertex_2$, respectively. If either of these points isn't inside a vertex-field of $\segraph$, then the reliability of $\segraph$ for going between $p_1$ and $p_2$ is marked as $0$. Otherwise, $\agent$ uses the $\pwave$ algorithm to find a path across the graph, using a value of $\potentialbar$ that is a decaying running average of the actual $\potentialbar$ used by the agent up until the point of measurement. If no path can be found, the agent uses its regular mechanisms for lowering the threshold until a path can be found. We then simulate the agent following the path. If the sampled trajectory hits a wall, we use the normal mechanism for handling traversal failure of inhibiting the failed edge and raising $\potentialbar$, then looking for a new path. If at any point no path can be found, the reliability of the graph for going from $p_1$ to $p_2$ is marked as $0$.

The ``true'' length of the path between $p_1$ and $p_2$ is determined using the graph used to initially generate the maze, denoted $d(p_1, p_2)$. If the reliability of the path is $r(p_1, p_2) \in \{0, 1\}$, and the set of node-points for the maze is $P$, then the formula for graph reliability is given as:
\begin{equation}
R(\segraph) = \frac{1}{\sum_{p_1, p_2 \in P}d(p_1, p_2)} \sum_{p_1, p_2 \in P} d(p_1, p_2) \cdot r(p_1, p_2)
\end{equation}

\section{Hebbian Learning}
\noindent\textbf{Definitions:}
\begin{enumerate}
\item $\vect{x} \bind \vect{y} := \frac{\vect{y}\vect{x}^*}{\norm{\vect{x}}^2}$, the ``bind'' matrix between vectors $\vect{x}$ and $\vect{y}$
\item $\cosim{\vect{x}}{\vect{y}} := \tfrac{\vect{x}^*\vect{y}}{\norm{\vect{x}}\cdot\norm{\vect{y}}}$, the cosine-similarity between vectors $\vect{x}$ and $\vect{y}$
\item $\vect{a} \odot \vect{b}$ is the Haddamard or element-wise product of $\vect{a}$ and $\vect{b}$
\end{enumerate}
\subsection{Algebra of Hebbian Learning} \label{app:hebb}
Suppose we have two vectors, $\vect{x}$ and $\vect{y}$, and we want to associate $\vect{y}$ with $\vect{x}$ via some matrix $\mat{W}$ so that $\mat{W}\vect{x} = \vect{y}$. The matrix $\vect{x} \bind \vect{y}$ satisfies this requirement. To see this, first observe that:
\begin{equation}
(\vect{x} \bind \vect{y}) \vect{z} = \tfrac{\vect{y}\vect{x}^*}{\norm{\vect{x}}^2}\vect{z} = \vect{y} \tfrac{1}{\norm{\vect{x}}} \tfrac{\vect{x}^*\vect{z}}{\norm{\vect{x}}} = \vect{y} \tfrac{\norm{\vect{z}}}{\norm{\vect{x}}} \tfrac{\vect{x}^*\vect{z}}{\norm{\vect{x}}\cdot\norm{\vect{z}}} = \vect{y} \tfrac{\norm{\vect{z}}}{\norm{\vect{x}}} \cosim{\vect{x}}{\vect{z}}
\label{eq:bind_identity}
\end{equation}
If $\vect{z} = \vect{x}$, we have $(\vect{x} \bind \vect{y}) \vect{x} = \vect{y}\tfrac{\norm{\vect{x}}}{\norm{\vect{x}}}\cosim{\vect{x}}{\vect{x}} = \vect{y} \cdot 1 \cdot 1 = \vect{y}$. From Equation \ref{eq:bind_identity} it follows that:
\begin{equation}
[\textstyle \sum_{i=1}^{k}{\vect{x}_i \bind \vect{y}_i}] \vect{z} = \textstyle \sum_{i=1}^{k}({\vect{x}_i \bind \vect{y}_i}) \vect{z} = \textstyle \sum_{i=1}^{k}{\vect{y}_i \tfrac{\norm{\vect{z}}}{\norm{\vect{x}_i}}\cosim{\vect{x}_i}{\vect{z}}}
\label{eq:bind_interference}
\end{equation}
If all of $\vect{x}_i$ are mutually orthogonal ($j\neq\ell \implies \cosim{\vect{x}_j}{\vect{x}_{\ell}}=0$), then $[\textstyle \sum_{i=1}^{k}{\vect{x}_i \bind \vect{y}_i}] \vect{x}_j = \vect{y}_j$ for all $j\in\{1...k\}$. In this sense, $\mat{W} = \textstyle \sum_{i=1}^{k}{\vect{x}_i \bind \vect{y}_i}$ acts like a dictionary, with $\vect{x}_i$ as ``keys'' and $\vect{y}_i$ as ``values''.
From Equation \ref{eq:bind_interference} we can see that the cosine-similarity of key-vectors controls the amount of interference between stored values.

\subsubsection{Algebraic Properties of Hebbian Learning}
\begin{align}
\intertext{Left distribution over addition:}
\vect{x} \bind (\vect{y} + \vect{z}) &= \tfrac{(\vect{y} + \vect{z})\vect{x}^*}{\norm{\vect{x}}^2} = \tfrac{\vect{y}\vect{x}^*}{\norm{\vect{x}}^2} + \tfrac{\vect{z}\vect{x}^*}{\norm{\vect{x}}^2} \nonumber{} \\
&= (\vect{x} \bind \vect{y}) + (\vect{x} \bind \vect{z})
\intertext{Right distribution over addition:}
(\vect{x} + \vect{y}) \bind \vect{z} &= \tfrac{\vect{z}(\vect{x}+\vect{y})^*}{\norm{\vect{x}+\vect{y}}^2} = \tfrac{\vect{z}\vect{x}^*}{\norm{\vect{x}+\vect{y}}^2} + \tfrac{\vect{z}\vect{y}^*}{\norm{\vect{x}+\vect{y}}^2} = \tfrac{\norm{\vect{x}}^2}{\norm{\vect{x}+\vect{y}}^2}\tfrac{\vect{z}\vect{x}^*}{\norm{\vect{x}}^2} + \tfrac{\norm{\vect{y}}^2}{\norm{\vect{x}+\vect{y}}^2}\tfrac{\vect{z}\vect{y}^*}{\norm{\vect{y}}^2} \nonumber{}  \\
 &= \tfrac{\norm{\vect{x}}^2}{\norm{\vect{x}+\vect{y}}^2}(\vect{x} \bind \vect{z})+ \tfrac{\norm{\vect{y}}^2}{\norm{\vect{x}+\vect{y}}^2}(\vect{y} \bind \vect{z}) 
\intertext{Left and right distribution over addition:}
(\vect{w} + \vect{x}) \bind (\vect{y} + \vect{z}) &= [(\vect{w} + \vect{x}) \bind \vect{y}] + [(\vect{w} + \vect{x}) \bind \vect{z}] \nonumber{} \\
&= \tfrac{\norm{\vect{w}}^2}{\norm{\vect{w} + \vect{x}}^2} (\vect{w} \bind \vect{y}) + \tfrac{\norm{\vect{x}}^2}{\norm{\vect{w} + \vect{x}}^2} (\vect{x} \bind \vect{y}) + \tfrac{\norm{\vect{w}}^2}{\norm{\vect{w} + \vect{x}}^2} (\vect{w} \bind \vect{z}) + \tfrac{\norm{\vect{x}}^2}{\norm{\vect{w} + \vect{x}}^2} (\vect{x} \bind \vect{z}) \nonumber{} \\
&= \tfrac{\norm{\vect{w}}^2}{\norm{\vect{w} + \vect{x}}}[\vect{w} \bind (\vect{y} + \vect{z})] + \tfrac{\norm{\vect{x}}^2}{\norm{\vect{w} + \vect{x}}^2}[\vect{x} \bind (\vect{y} + \vect{z})] 
\intertext{Left distribution over element-wise multiplication:}
\vect{x} \bind (\vect{y} \odot \vect{z}) &= \tfrac{(\vect{y} \odot \vect{z})\vect{x}^*}{\norm{\vect{x}}^2} = \tfrac{\vect{y}\vect{x}^*}{\norm{\vect{x}}^2} \odot \tfrac{\vect{z}\vect{1}^*}{\norm{\vect{1}}^2}\norm{\vect{1}}^2 \nonumber{} \\
&= \norm{\vect{1}}^2(\vect{x} \bind \vect{y}) \odot (\vect{1} \bind \vect{z}) = \norm{\vect{1}}^2(\vect{1} \bind \vect{y}) \odot (\vect{x} \bind \vect{z}) 
\intertext{Right distribution over element-wise multiplication:}
(\vect{x} \odot \vect{y}) \bind \vect{z} &= \tfrac{\vect{z}(\vect{x} \odot \vect{y})^*}{\norm{\vect{x} \odot \vect{y}}^2} = \tfrac{\vect{z} (\vect{x}^* \odot \vect{y}^*)}{\norm{\vect{x} \odot \vect{y}}^2} = \tfrac{\vect{z}\vect{x}^*}{\norm{\vect{x}}^2}\tfrac{\norm{\vect{x}}^2}{\norm{\vect{x}\odot\vect{y}}^2} \odot \tfrac{\vect{1}\vect{y}^*}{\norm{\vect{y}}^2}\norm{\vect{y}}^2 \nonumber{} \\
 &= \tfrac{\norm{\vect{x}}^2 \norm{\vect{y}}^2}{\norm{\vect{x} \odot \vect{y}}^2}(\vect{x}\bind\vect{z})\odot(\vect{y} \bind \vect{1}) = \tfrac{\norm{\vect{x}}^2 \norm{\vect{y}}^2}{\norm{\vect{x} \odot \vect{y}}^2}(\vect{x}\bind\vect{1})\odot(\vect{y} \bind \vect{z}) \label{eq:right_haddamard_distribution}
\end{align}

\section{Modulation of Harmonic Relational Keys} \label{app:hark}
\noindent\textbf{Definitions:}
\begin{enumerate}
\item $\vect{x} \bind \vect{y} := \frac{\vect{y}\vect{x}^*}{\norm{\vect{x}}^2}$, the ``bind'' matrix between vectors $\vect{x}$ and $\vect{y}$
\item $\cosim{\vect{x}}{\vect{y}} := \tfrac{\vect{x}^*\vect{y}}{\norm{\vect{x}}\cdot\norm{\vect{y}}}$, the cosine-similarity between vectors $\vect{x}$ and $\vect{y}$
\item $\vect{a} \odot \vect{b}$ is the Haddamard or element-wise product of $\vect{a}$ and $\vect{b}$
\item $\vect{a} \oslash \vect{b}$ is the Haddamard or element-wise quotient of $\vect{a}$ and $\vect{b}$
\item $\mathcal{C} := \{z\in\mathbb{C} : \abs{z}=1\}$, the set of complex numbers with magnitude 1
\end{enumerate}
Now, our goal is to construct a method of generating spatiotemporally-specific ``key'' vectors. Consider the function:
\begin{equation}\label{eq:qpkm}
\vect{x}_{\vect{\delta}} = f_{\textrm{HaRK}}(\vect{x}, \vect{\delta}) = \vect{x} \odot e^{2\pi i \mat{\Gamma}\vect{\delta}}
\end{equation}
where $\vect{\delta} \in \mathbb{R}^d$, $\mat{\Gamma} \in \mathbb{R}^{N \times d}$, and $\vect{x}, \vect{x}_{\vect{\delta}} \in \mathcal{C}^N$ (see above definition). In the future, we let $\vect{\gamma}_j^{\top} := \mat{\Gamma}_{j,}$. For our purposes, it can be helpful to think of $\vect{\delta}$ as existing in some low ($d \leq 100$) dimensional ``physical'' space. Then $\mat{\Gamma}$ represents a matrix of $N$ frequencies oriented in $d$-dimensional space, where $N$ is quite large ($\geq 10000$). It is easy to show that $f_{\textrm{HaRK}}(\vect{x}, \vect{0}) = \vect{x}$, and that $f_{\textrm{HaRK}}(f_{\textrm{HaRK}}(\vect{x}, \vect{\delta}_1), \vect{\delta}_2) = f_{\textrm{HaRK}}(\vect{x}, \vect{\delta}_1 + \vect{\delta}_2)$, meaning the structure of addition in $\mathbb{R}^d$ is preserved by $f$. This means that $\vect{x}_{\vect{\delta}}$ implicitly encodes $\vect{\delta}$, at least relative to some ``origin'' vector $\vect{x}_{\vect{0}}$. Projecting from $\mathbb{R}^d$ to $\mathcal{C}^N$ in this way grants us (1) a simple model for grid cells, (2) many more than $d$-orthogonal keys, (3) uniform magnitude of key-vectors, and (4) a relative (rather than absolute) coordinate system, which is cognitively attractive.

\subsection{Hebbian Learning with QPKM}
\def \xterm {\alpha \cdot \vect{\mu} \odot \vect{x}}
\def \zterm {\beta \cdot \vect{\eta} \odot \vect{x}_{\vect{\delta}}}
Now, we investigate the consequences of using key-vectors generated by $f$ when the keys are element-wise multiplied by some ``modulatory'' vectors $\vect{\mu}$ and $\vect{\eta}$, with scalar multipliers $\alpha$ and $\beta$. Consider when $\mat{W} = (\xterm)\bind\vect{y}$ and we query $\mat{W}$ with $(\zterm)$:
\begin{align}\label{eq:response_curve}
\mat{W}(\zterm) &= [(\xterm)\bind\vect{y}](\zterm) = \vect{y} \frac{\norm{\zterm}}{\norm{\xterm}} \cosim{\xterm}{\zterm} \nonumber{} \\
&= \vect{y}\frac{\beta}{\alpha}\frac{\norm{\vect{\eta}}}{\norm{\vect{\mu}}}\frac{\langle\xterm, \zterm\rangle}{\norm{\xterm} \cdot \norm{\zterm}} = \vect{y}\frac{\beta}{\alpha}\frac{\norm{\vect{\eta}}}{\norm{\vect{\mu}}}\frac{\langle\vect{\mu} \odot \vect{x}, \vect{\eta} \odot \vect{x}_{\vect{\delta}}\rangle}{\norm{\vect{\mu}}\cdot\norm{\vect{\eta}}} \nonumber{} \\
&= \vect{y}\frac{\beta}{\alpha}\frac{1}{\norm{\vect{\mu}}^2}\langle\vect{\mu} \odot \vect{x}, \vect{\eta} \odot \vect{x}_{\vect{\delta}}\rangle = \vect{y}\frac{\beta}{\alpha}\frac{1}{\norm{\vect{\mu}}^2}\langle\vect{\mu} \odot \vect{x}, \vect{\eta} \odot \vect{x} \odot e^{2\pi i \mat{\Gamma}\vect{\delta}}\rangle \nonumber{} \\
&= \vect{y}\frac{\beta}{\alpha}\frac{1}{\norm{\vect{\mu}}^2}\sum_{j=1}^N{\overline{(\mu_jx_j)}(\eta_jx_je^{2\pi i \mat{\Gamma}_{j,}\vect{\delta}})} = \vect{y}\frac{\beta}{\alpha}\frac{1}{\norm{\vect{\mu}}^2}\sum_{j=1}^N{\bar{\mu}_j\eta_j\overline{x}_jx_je^{2\pi i \vect{\gamma}_j^{\top}\vect{\delta}}} \nonumber{} \\
&= \vect{y}\frac{\beta}{\alpha}\frac{1}{\norm{\vect{\mu}}^2}\sum_{j=1}^N{\bar{\mu}_j\eta_je^{2\pi i \vect{\gamma}_j^{\top}\vect{\delta}}}
\end{align}
Let $c(\vect{\delta}) = \frac{\beta}{\alpha}\frac{1}{\norm{\vect{\mu}}^2}\sum_{j=1}^N{\bar{\mu}_j\eta_je^{2\pi i \vect{\gamma}_j^{\top}\vect{\delta}}}$ be our ``interference'' function, so that we have:
\begin{equation}
[(\xterm)\bind\vect{y}](\zterm) = c(\vect{\delta}) \cdot \vect{y}
\end{equation}
Why do we interpret $c(\vect{\delta})$ as an ``interference'' function? Looking at the second line of the derivation in Equation \ref{eq:response_curve}, we see that $c(\vect{\delta}) = \frac{\beta}{\alpha}\frac{\norm{\vect{\eta}}}{\norm{\vect{\mu}}}\cosim{\vect{\mu}\odot\vect{x}}{\vect{\eta}\odot\vect{x}_{\vect{\delta}}}$, which is just a scaled cosine-similarity between $\vect{x}$ and $\vect{x}_{\vect{\delta}}$ (modulated by $\vect{\eta}$ and $\vect{\mu}$, respectively). The ``shape'' of $c(\vect{\delta})$ controls the ``amount'' of $\vect{y}$ that is retrieved by using the query vector $\vect{x}_{\vect{\delta}}$. In general, we will want $c(\vect{0})=1$ so that when we query with $\vect{x}$ (the original storage vector), we get back $\vect{y}$, and additionally, as $\norm{\vect{\delta}} \rightarrow \infty$, we would like $c(\vect{\delta})=0$, so that values stores ``far-away'' from each other have minimal interference (though there are other possibilities). By carefully choosing $\alpha$, $\vect{\mu}$, $\beta$, $\vect{\eta}$, and $\mat{\Gamma}$, we can control the shape of $c(\vect{\delta})$. By doing this, will be able to approximately control the amount of interference between keys as a function of distance $\norm{\vect{\delta}}$ between them.

First, we want to enforce that $c(\vect{0}) = 1$, so that we have $[(\xterm)\bind\vect{y}](\beta \cdot \vect{\eta} \odot \vect{x}_{\vect{0}}) = \vect{y}$.
\begin{equation}
c(\vect{0}) = \frac{\beta}{\alpha}\frac{1}{\norm{\vect{\mu}}^2}\sum_{j=1}^N{\bar{\mu}_j\eta_je^{2\pi i \vect{\gamma}_j^{\top}\vect{0}}} = \frac{\beta}{\alpha}\frac{\sum_{j=1}^N{\bar{\mu}_j\eta_j}}{\norm{\vect{\mu}}^2} = 1 \implies \frac{\alpha}{\beta} = \frac{\sum_{j=1}^N{\bar{\mu}_j\eta_j}}{\norm{\vect{\mu}}^2}
\end{equation}

For the sake of simplicity, we split the problem into two situations: modulation on retrieval, or modulation on storage.

If we only perform modulation on retrieval, then we set $\alpha=1$ and $\vect{\mu}=\vect{1}$, yielding:
\begin{align*}
\frac{1}{\beta} = \frac{\sum_{j=1}^N{\eta_j}}{\norm{\vect{1}}^2} = \frac{\sum_{j=1}^N{\eta_j}}{N} \implies \beta = \frac{N}{\sum_{j=1}^N{\eta_j}} \implies c(\vect{\delta}) = \frac{1}{\sum_{j=1}^N{\eta_j}}\sum_{j=1}^N{\eta_je^{2\pi i \vect{\gamma}_j^{\top}\vect{\delta}}}
\end{align*}
If we only perform modulation on storage, then we set $\beta=1$ and $\vect{\eta}=\vect{1}$, yielding:
\begin{align*}
\alpha = \frac{\sum_{j=1}^N{\bar{\mu}_j}}{\norm{\vect{\mu}}^2} \implies c(\vect{\delta}) = \frac{\norm{\vect{\mu}}^2}{\sum_{j=1}^N{\bar{\mu}_j}}\frac{1}{\norm{\vect{\mu}}^2}\sum_{j=1}^N{\bar{\mu}_je^{2\pi i \vect{\gamma}_j^{\top}\vect{\delta}}} = \frac{1}{\sum_{j=1}^N{\bar{\mu}_j}}\sum_{j=1}^N{\bar{\mu}_je^{2\pi i \vect{\gamma}_j^{\top}\vect{\delta}}}
\end{align*}

The key thing to notice is that either way, $c(\vect{\delta}) = \frac{1}{\sum_{j=1}^N{g_j}}\sum_{j=1}^N{g_je^{2\pi i \vect{\gamma}_j^{\top}\vect{\delta}}}$, which happens to have the form of a Fourier decomposition. There is a caveat, however, which is that the Fourier decomposition implicitly assumes a uniform ``density'' of frequencies $\vect{\gamma}_j$: for both biological and representational reasons, we prefer to use a non-uniform density of frequencies.

\subsection{Derivation of $g_j$}

We will use the following statement, an expression of the ``law of the unconscious statistician'':
\begin{equation}
\left[\lim_{k \rightarrow \infty} \frac{1}{k} \sum_{j=1}^{k}{h(x_j)}, x_j \sim p_x(x)\right] = E_{p_x(x)}[h(x)] = \int_{\Omega}{h(x)p_x(x)dx}
\label{eq:lazy_stats}
\end{equation}
As well as these definitions for the Fourier and inverse Fourier transform.
\begin{align*}
\hat{f}(\vect{\xi}) &= \mathcal{F}_{\vect{x}}[f(\vect{x})](\vect{\xi}) = \int_{\mathbb{R}^d}{f(\vect{x})e^{-2\pi i \vect{x}^{\top} \vect{\xi}}d\vect{x}} & \textrm{(the Fourier transform)} \\
f(\vect{x}) &= \mathcal{F}^{-1}_{\vect{\xi}}[\hat{f}(\vect{\xi})](\vect{x}) =  \int_{\mathbb{R}^d}{\hat{f}(\vect{\xi})e^{2\pi i \vect{\xi}^{\top}\vect{x}}d\vect{\xi}} & \textrm{(the inverse Fourier transform)}
\end{align*}
Suppose that we distribute $\vect{\gamma}_j$ according to $p_{\vect{\gamma}}(\vect{\gamma})$. We will make $g_j$ be a function of $\vect{\gamma}_j$, so $g_j = g(\vect{\gamma}_j)$. This means that $c(\vect{\delta}) = \frac{1}{\sum_{j=1}^{N}{g(\vect{\gamma}_j)}}\sum_{j=1}^N{g(\vect{\gamma}_j)e^{2\pi i \vect{\gamma}_j^{\top}\vect{\delta}}}, \vect{\gamma}_j \sim p_{\vect{\gamma}}(\vect{\gamma})$

Now, consider $c_{\infty}(\vect{\delta}) = \lim_{N\rightarrow\infty}c(\vect{\delta})$, the function we want to approximate with $c(\vect{\delta})$:
\begin{align}
c_{\infty}(\vect{\delta}) &= \lim_{N\rightarrow\infty}c(\vect{\delta}) = \left[\lim_{N\rightarrow\infty}\frac{1}{\sum_{j=1}^{N}{g(\vect{\gamma}_j)}}\sum_{j=1}^N{g(\vect{\gamma}_j)e^{2\pi i \vect{\gamma}_j^{\top}\vect{\delta}}}, \vect{\gamma}_j \sim p_{\vect{\gamma}}(\vect{\gamma})\right] \nonumber{} \\
c_{\infty}(\vect{\delta})&= \left[\lim_{N\rightarrow\infty}\frac{N}{\sum_{j=1}^{N}{g(\vect{\gamma}_j)}}\frac{1}{N}\sum_{j=1}^N{g(\vect{\gamma}_j)e^{2\pi i \vect{\gamma}_j^{\top}\vect{\delta}}}, \vect{\gamma}_j \sim p_{\vect{\gamma}}(\vect{\gamma})\right] \nonumber{} \\
&= \left[\lim_{N\rightarrow\infty}\frac{N}{\sum_{j=1}^{N}{g(\vect{\gamma}_j)}}, \vect{\gamma}_j \sim p_{\vect{\gamma}}(\vect{\gamma})\right] \cdot \left[\lim_{N\rightarrow\infty}\frac{1}{N}\sum_{j=1}^N{g(\vect{\gamma}_j)e^{2\pi i \vect{\gamma}_j^{\top}\vect{\delta}}}, \vect{\gamma}_j \sim p_{\vect{\gamma}}(\vect{\gamma})\right] \nonumber{} \intertext{We let $C = \left[\lim_{N\rightarrow\infty}\frac{N}{\sum_{j=1}^{N}{g(\vect{\gamma}_j)})}, \vect{\gamma}_j \sim p_{\vect{\gamma}}(\vect{\gamma})\right]$ and substitute it in...}
&= C \cdot \left[\lim_{N\rightarrow\infty}\frac{1}{N}\sum_{j=1}^N{g(\vect{\gamma}_j)e^{2\pi i \vect{\gamma}_j^{\top}\vect{\delta}}}, \vect{\gamma}_j \sim p_{\vect{\gamma}}(\vect{\gamma})\right] \nonumber{} \intertext{We apply the ``law of the unconscious statistician'' (Equation \ref{eq:lazy_stats})...}
&= C \cdot \int_{\mathbb{R}^d}{g(\vect{\gamma})e^{2\pi i \vect{\gamma}^{\top} \vect{\delta}}}p_{\vect{\gamma}}(\vect{\gamma})d{\vect{\gamma}} = C \cdot \int_{\mathbb{R}^d}{[g(\vect{\gamma})p_{\vect{\gamma}}(\vect{\gamma})]e^{2\pi i \vect{\gamma}^{\top} \vect{\delta}}}d{\vect{\gamma}} \nonumber{} \intertext{Finding the definition of the \emph{inverse} Fourier transform, we substitute it in...}
c_{\infty}(\vect{\delta}) &= C \cdot \mathcal{F}^{-1}_{\vect{\gamma}}[g(\vect{\gamma})p_{\vect{\gamma}}(\vect{\gamma})](\vect{\delta}) \nonumber{} \intertext{We divide each side by $C$ and take the Fourier transform of both sides...}
\tfrac{1}{C}\mathcal{F}_{\vect{\delta}}[c_{\infty}(\vect{\delta})](\vect{\gamma}) &= g(\vect{\gamma})p_{\vect{\gamma}}(\vect{\gamma}) \nonumber{}  \intertext{Letting $\hat{c}_{\infty}(\vect{\gamma}) = \mathcal{F}_{\vect{\delta}}[c_{\infty}(\vect{\delta})](\vect{\gamma})$ be the Fourier transform of $c_{\infty}(\vect{\delta})$...}
\frac{1}{C}\cdot\frac{\hat{c}_{\infty}(\vect{\gamma})}{p_{\vect{\gamma}}(\vect{\gamma})} &= g(\vect{\gamma}) \nonumber{} \intertext{It turns out $C$ is a free parameter which will be normalized-out anyway, so we drop it, yielding...}
g(\vect{\gamma}) &= \frac{\hat{c}_{\infty}(\vect{\gamma})}{p_{\vect{\gamma}}(\vect{\gamma})}
\label{eq:gain_expression}
\end{align}
So, the modulation ``gains'' $g(\vect{\gamma}_j)$ just have to be the Fourier transform of the target cosine-similarity function evaluated at $\vect{\gamma}_j$ divided by the density of $\vect{\gamma}_j$, determined by our choice of $\mat{\Gamma}$.

\subsection{Derivation of $\hat{c}_{\infty}(\vect{\gamma})$}

We are primarily interested in two cases:
\begin{enumerate}
\item $c_{\sigma}(\vect{\delta}) = e^{-\norm{\tfrac{\delta}{\sigma}}^2}$, an isometric Gaussian with standard deviation $\sigma$
\item $c_{\vect{\sigma}}(\vect{\delta}) = e^{-\norm{\vect{\delta} \oslash \vect{\sigma}}^2}$, a non-isometric diagonal Gaussian with standard-deviations $\vect{\sigma}$
\end{enumerate}

According to Abramowitz and Stegun (1972, p. 302, equation 7.4.6), the Fourier transform of a Gaussian given by $e^{-x^2\sigma^{-2}}$ is:
\begin{equation}
\mathcal{F}_{x}\left[e^{-ax^2}\right](\xi) = \pi^{\frac{1}{2}}a^{-\frac{1}{2}}e^{-\pi^2\xi^2a^{-1}}
\end{equation}
In our case, $a = \sigma^{-2}$, so we substitute and get:
\begin{equation}
\mathcal{F}_{x}\left[e^{-x^2\sigma^{-2}}\right](\xi) = \pi^{\frac{1}{2}}(\sigma^{-2})^{-\frac{1}{2}}e^{-\pi^2\xi^2(\sigma^{-2})^{-1}} = \sqrt{\pi}\sigma e^{-\pi^2\sigma^2\xi^2}
\end{equation}
What is the Fourier transform for a Gaussian given by $e^{-\norm{\vect{x} \oslash \vect{\sigma}}^2}$, where $\vect{\sigma}$ is the standard deviation along each axis? Note that:
\begin{align*}
e^{-\norm{\vect{x} \oslash \vect{\sigma}}^2} = e^{-\sum_{j=1}^{d}{x_j^2\sigma_j^{-2}}} = \textstyle\prod_{j=1}^{d}{e^{-x_j^2\sigma_j^{-2}}}
\end{align*}
According to [https://see.stanford.edu/materials/lsoftaee261/chap8.pdf], this means that:
\begin{align}
\mathcal{F}_{\vect{x}}\left[e^{-\norm{\vect{x} \oslash \vect{\sigma}}^2}\right](\vect{\xi}) &= \prod_{j=1}^{d}{\mathcal{F}_{x}\left[e^{-x_j^2\sigma_j^{-2}}\right](\vect{\xi})} = \prod_{j=1}^{d}{\sqrt{\pi}\sigma_j e^{-\pi^2\sigma_j^2\xi_j^2}} \nonumber{} \\
&= \pi^{\frac{d}{2}} \left(\textstyle\prod_{j=1}^{d}{\sigma_j}\right) e^{-\sum_{j=1}^{d}{\pi^2\sigma_j^2\xi_j^2}} = \pi^{\frac{d}{2}} \left(\textstyle\prod_{j=1}^{d}{\sigma_j}\right) e^{-\pi^2\norm{ \vect{\sigma} \odot \vect{\xi}}^2}
\end{align}

From this, we directly get the expression for the Fourier transform of the non-isometric diagonal Gaussian with standard-deviation vector $\vect{\sigma}$:
\begin{equation}
\hat{c}_{\vect{\sigma}}(\vect{\gamma}) = \pi^{\frac{d}{2}} \left(\textstyle\prod_{j=1}^{d}{\sigma_j}\right) e^{-\pi^2\norm{ \vect{\sigma} \odot \vect{\gamma}}^2}
\label{eq:fourier_expression}
\end{equation}
Notice that $c_{\vect{\sigma}}(\vect{\delta}) = c_{\sigma}(\vect{\delta})$ when every $\sigma_j=\sigma$. We can use this result to get the Fourier transform of the isometric Gaussian with standard-deviation $\sigma$:
\begin{equation}
\hat{c}_{\sigma}(\vect{\gamma}) = \pi^{\frac{d}{2}} \sigma^d e^{-\pi^2\sigma^2\norm{\vect{\gamma}}^2}
\end{equation}

\subsection{Derivation of $p_{\vect{\gamma}}(\vect{\gamma})$}

Now we have to address the question of choosing $\mat{\Gamma}$, which we do by sampling individual $\vect{\gamma}$ from a chosen distribution, $p_{\vect{\gamma}}(\vect{\gamma})$. For the sake of simplicity, we choose $p_{\vect{\gamma}}(\vect{\gamma})$ to be symmetric, meaning that $\forall \vect{a}, \vect{b} \in \mathbb{R}^d$, $\norm{\vect{a}} = \norm{\vect{b}} \implies p_{\vect{\gamma}}(\vect{a}) = p_{\vect{\gamma}}(\vect{b})$. This implies that there exists some function $h(r)$ s.t. $p_{\vect{\gamma}}(\vect{\gamma}) = h(\norm{\vect{\gamma}})$. In order to sample individual $\vect{\gamma}$, we will actually specify a distribution over frequency \emph{magnitudes} $q_r(\norm{\vect{\gamma}})$, then assign a spherically uniform orientation to each frequency.

In order to relate $h(\norm{\vect{\gamma}})$ and $q_r(\norm{\vect{\gamma}})$, notice that the density of frequency \emph{magnitudes} must incorporate frequencies of all \emph{orientations}. What this means, geometrically, is that the density $h(r)$ of any individual frequency with magnitude $r$ must be equal to to the density of that magnitude, $q_r(r)$, \emph{divided} by the area of the hypersphere with radius $r$ (the area is proportional to $r^{d-1}$). We can use this to get the exact relationship between $p_{\vect{\gamma}}$ and $q_r(r)$. Ignoring constants, this is:
\begin{align}
h(\norm{\vect{\gamma}}) &\propto \frac{q_r(\norm{\vect{\gamma}})}{\norm{\vect{\gamma}}^{d-1}} \nonumber{} \\
p_{\vect{\gamma}}(\vect{\gamma}) = h(\norm{\vect{\gamma}}) &\propto q_r(\norm{\vect{\gamma}}) \cdot \norm{\vect{\gamma}}^{1-d} \nonumber{} \\
p_{\vect{\gamma}}(\vect{\gamma}) &= \tfrac{1}{\int_{\Omega}{q(\norm{\vect{\gamma}}) \cdot \norm{\vect{\gamma}}^{1-d}}d\vect{\gamma}} \cdot q_r(\norm{\vect{\gamma}}) \cdot \norm{\vect{\gamma}}^{1-d} \nonumber{} \intertext{Letting $r = \norm{\vect{\gamma}}$, and $\gamma_{\min}$ and $\gamma_{\max}$ the minimum and maximum of $\norm{\vect{\gamma}}$...}
p_{\vect{\gamma}}(\vect{\gamma}) &= \left[\frac{2\pi^{\frac{d}{2}}}{\Gamma(\frac{d}{2})} \int_{\gamma_{\min}}^{\gamma_{\max}}{q_r(r) \cdot r^{1-d} \cdot r^{d-1}dr}\right]^{-1}  \hspace{-1em} \cdot q_r(\norm{\vect{\gamma}}) \cdot \norm{\vect{\gamma}}^{1-d} \nonumber{} \\
p_{\vect{\gamma}}(\vect{\gamma}) &= \left[\frac{2\pi^{\frac{d}{2}}}{\Gamma(\frac{d}{2})} \int_{\gamma_{\min}}^{\gamma_{\max}}{q_r(r)dr}\right]^{-1}\hspace{-1em} \cdot q_r(\norm{\vect{\gamma}}) \cdot \norm{\vect{\gamma}}^{1-d} \nonumber{} \intertext{Letting $Q(r)$ be the antiderivative of $q_r(r)$...}
p_{\vect{\gamma}}(\vect{\gamma}) &= \left[\frac{2\pi^{\frac{d}{2}}}{\Gamma(\frac{d}{2})} [Q(\gamma_{\max}) - Q(\gamma_{\min})]\right]^{-1}\hspace{-1em} \cdot q_r(\norm{\vect{\gamma}}) \cdot \norm{\vect{\gamma}}^{1-d} \nonumber{} \\
p_{\vect{\gamma}}(\vect{\gamma}) &= \frac{\Gamma(\frac{d}{2})}{2\pi^{\frac{d}{2}}} \cdot \frac{q_r(\norm{\vect{\gamma}}) \cdot \norm{\vect{\gamma}}^{1-d}}{Q(\gamma_{\max}) - Q(\gamma_{\min})}
\end{align}
This gives us our analytical density over frequencies $p_{\vect{\gamma}}(\vect{\gamma})$ given a chosen density over frequency magnitudes $q_r(\norm{\vect{\gamma}})$.

\subsection{Numerical calculation of $g(\vect{\gamma})$}
It turns out that $q_r(\norm{\vect{\gamma}}) = \norm{\vect{\gamma}}^{-1}$ is a nice choice, giving us some scale-free and dimensionality-invariant properties, as well as some modicum of agreement with the observed distribution of scales in grid-cell responses.

$q_r(\norm{\vect{\gamma}}) = \norm{\vect{\gamma}}^{-1}$ implies the antiderivative of $q_r(\norm{\vect{\gamma}})$ is $Q(\norm{\vect{\gamma}}) = \ln(\norm{\vect{\gamma}})$. Recalling from Equation \ref{eq:gain_expression} that $g(\vect{\gamma}) = \frac{\hat{c}_{\infty}(\vect{\gamma})}{p_{\vect{\gamma}}(\vect{\gamma})}$ and from Equation \ref{eq:fourier_expression} that $\hat{c}_{\vect{\sigma}}(\vect{\gamma}) = \pi^{\frac{d}{2}} \left(\textstyle\prod_{j=1}^{d}{\sigma_j}\right) e^{-\pi^2\norm{ \vect{\sigma} \odot \vect{\gamma}}^2}$, we now have all the information we need to determine $g(\vect{\gamma})$ as a function of $\vect{\sigma}$:
\begin{align}
g(\vect{\gamma}) &= \frac{\hat{c}_{\vect{\sigma}}(\vect{\gamma})}{p_{\vect{\gamma}}(\vect{\gamma})} = \frac{\pi^{\frac{d}{2}} \left(\prod_{j=1}^{d}{\sigma_j}\right) e^{-\pi^2\norm{ \vect{\sigma} \odot \vect{\gamma}}^2}}{\norm{\vect{\gamma}}^{-1} \cdot \norm{\vect{\gamma}}^{1-d}} \nonumber{} \\&= \norm{\vect{\gamma}}^d\pi^{\frac{d}{2}} \left(\textstyle\prod_{j=1}^{d}{\sigma_j}\right) e^{-\pi^2\norm{ \vect{\sigma} \odot \vect{\gamma}}^2} \nonumber{} \\
&=  e^{\textstyle{\ln\left(\norm{\vect{\gamma}}^d\pi^{\frac{d}{2}} \left(\textstyle\prod_{j=1}^{d}{\sigma_j}\right) e^{-\pi^2\norm{ \vect{\sigma} \odot \vect{\gamma}}^2}\right)}} \nonumber{} \\
&=  e^{\textstyle{\ln(\norm{\vect{\gamma}}^d) + \ln(\pi^{\frac{d}{2}}) + \ln\left(\textstyle\prod_{j=1}^{d}{\sigma_j}\right) -\pi^2\norm{ \vect{\sigma} \odot \vect{\gamma}}^2}} \nonumber{} \\
&=  e^{\textstyle{d\ln(\norm{\vect{\gamma}}) + \tfrac{d}{2}\ln(\pi) + \textstyle\sum_{j=1}^{d}\ln(\sigma_j) -\pi^2\norm{ \vect{\sigma} \odot \vect{\gamma}}^2}} \nonumber{} \\
&= \left(e^{\textstyle\tfrac{d}{2}\ln(\pi) + \textstyle\sum_{j=1}^{d}\ln(\sigma_j)}\right)\left(e^{\textstyle{d\ln(\norm{\vect{\gamma}}) -\pi^2\norm{ \vect{\sigma} \odot \vect{\gamma}}^2}}\right) \nonumber{} \intertext{Note that with respect to $\vect{\gamma}$, the first term is a constant which will be normalized out later by $\alpha$ and $\beta$, so we drop it, leaving...}
&=  e^{{d\ln(\norm{\vect{\gamma}}) -\pi^2\norm{ \vect{\sigma} \odot \vect{\gamma}}^2}}
\label{eq:analytical_gain}
\end{align}
Calculation of this term can become numerically unstable on a computer. To correct this, we find the maximum value of the expression, and divide it out, re-arranging so the correction happens inside the exponential. To do this, we first have to realize that the relevant $\sigma$ is the minimum component of $\vect{\sigma}$, $\sigma_{\min}$. For clarity, we replace $\norm{\vect{\gamma}}$ with $\gamma$, and find where the derivative of Equation \ref{eq:analytical_gain} is equal to $0$ to find its maximum:
\begin{align}
0 &= \frac{d}{d\gamma} (e^{{d\ln(\norm{\vect{\gamma}}) -\pi^2\sigma_{\min}^2\gamma^2}}) = \left(\frac{d}{\gamma} - 2\gamma\pi^2\sigma_{\min}^2\right)e^{{d\ln(\norm{\vect{\gamma}}) -\pi^2\sigma_{\min}^2\gamma^2}}\nonumber{} \\
0 &= d - 2\gamma^2\pi^2\sigma_{\min}^2 \nonumber{} \\
\gamma_{\max} &= \frac{\sqrt{d}}{\sqrt{2}\pi\sigma_{\min}} \nonumber{}
\end{align}
Now we plug this back into Equation \ref{eq:analytical_gain}...
\begin{align}
e^{d\ln(\gamma_{\max}) -\pi^2\sigma_{\min}^2\gamma_{\max}^2} &= e^{d\ln\left(\frac{\sqrt{d}}{\sqrt{2}\pi\sigma_{\min}} \nonumber{}\right) -\pi^2\sigma_{\min}^2\left(\frac{\sqrt{d}}{\sqrt{2}\pi\sigma_{\min}}\right)^2} \nonumber{} \\
&= e^{d\ln\left(\frac{\sqrt{d}}{\sqrt{2}\pi\sigma_{\min}} \right) -\pi^2\sigma_{\min}^2\frac{d}{2\pi^2\sigma_{\min}^2}} \nonumber{} \\
&= e^{d\ln\left(\frac{\sqrt{d}}{\sqrt{2}\pi\sigma_{\min}} \right) - \frac{d}{2}} = e^{d\left(\ln\left(\frac{\sqrt{d}}{\sqrt{2}\pi\sigma_{\min}} \right) - \frac{1}{2}\right)} = e^{d\left(\ln\left(\sqrt{\frac{d}{2\pi^2\sigma_{\min}^2}} \right) - \frac{1}{2}\right)} \nonumber{} \\
&= e^{d\left(\frac{1}{2}\ln\left(\frac{d}{2\pi^2\sigma_{\min}^2} \right) - \frac{1}{2}\right)} = e^{\frac{d}{2}\left(\ln\left(\frac{d}{2\pi^2\sigma_{\min}^2} \right) - 1\right)}
\label{eq:numerical_correction}
\end{align}
Now, to ensure numerical stability, we divide Equation \ref{eq:analytical_gain} by the term derived in Equation \ref{eq:numerical_correction} so that the maximum value is fixed at $1$:
\begin{equation}
g(\vect{\gamma}) = \frac{e^{d\ln(\norm{\vect{\gamma}}) -\pi^2\norm{ \vect{\sigma} \odot \vect{\gamma}}^2}}{e^{\frac{d}{2}\left(\ln\left(\frac{d}{2\pi^2\sigma_{\min}^2}\right) - 1\right)}} = e^{\left[d\ln(\norm{\vect{\gamma}}) -\pi^2\norm{ \vect{\sigma} \odot \vect{\gamma}}^2 - \frac{d}{2}\left(\ln\left(\frac{d}{2\pi^2\sigma_{\min}^2}\right) - 1\right)\right]}
\label{eq:numerical_gain}
\end{equation}

This is a numerically stable expression which we can use to compute $g(\vect{\gamma})$ even for very large $d$, meaning that we can apply this system to high-dimensional sensory or action spaces, not just low-dimensional ``physical'' spaces.

\subsection{Picking $\gamma_{\min}$ and $\gamma_{\max}$}
As Equation \ref{eq:numerical_gain} is a Gaussian, there will be values of $\vect{\gamma}$ which are negligible. We can use Equation \ref{eq:numerical_gain} to find the minimum and maximum ``useful'' $\norm{\vect{\gamma}}$ for a given choice of $\sigma$, $\gamma^{\sigma}_{\min}$ and $\gamma^{\sigma}_{\max}$, respectively (we now restrict ourselves to the isometric case for the sake of simplicity). We can use this fact to go from a range of scales, $\sigma$, that we would like to represent, to a range of frequency magnitudes $\norm{\vect{\gamma}}$ that our system will need to represent those scales. Note that we are here assuming the isometric Gaussian case. Let $\epsilon$ be our ``minimum'' non-zero gain, then setting $\epsilon$ equal to Equation \ref{eq:gain_expression} yields:

\begin{align}
\epsilon &= e^{\left[d\ln(\gamma) - (\pi\sigma\gamma)^2 - \frac{d}{2}\left(\ln\left(\frac{d}{2\pi^2\sigma^2}\right)-1\right)\right]} = e^{\left[d\ln(\gamma) - (\pi\sigma\gamma)^2 - \frac{d}{2}\ln\left(\frac{d}{2\pi^2\sigma^2}\right)+\frac{d}{2}\right]} \nonumber{} \\
\epsilon &= e^{d\ln(\gamma)}e^{- (\pi\sigma\gamma)^2}e^{-\frac{d}{2}\ln\left(\frac{d}{2\pi^2\sigma^2}\right)}e^{\frac{d}{2}} = e^{\ln(\gamma^d)}e^{- (\pi\sigma\gamma)^2}e^{\ln\left(\left(\frac{2\pi^2\sigma^2}{d}\right)^{\frac{d}{2}}\right)}e^{\frac{d}{2}}\nonumber{} \\
\epsilon &= \gamma^d e^{-(\pi\sigma\gamma)^2} \left(\tfrac{2\pi^2\sigma^2}{d}\right)^{\frac{d}{2}} e^{\frac{d}{2}} = \gamma^d e^{-(\pi\sigma\gamma)^2} \left(\tfrac{2e\pi^2\sigma^2}{d}\right)^{\frac{d}{2}}\nonumber{} \\
\epsilon\left(\tfrac{d}{2e\pi^2\sigma^2}\right)^{\frac{d}{2}} &= \gamma^d e^{-(\pi\sigma\gamma)^2} \nonumber{} \\
\left(\epsilon\left(\tfrac{d}{2e\pi^2\sigma^2}\right)^{\frac{d}{2}}\right)^{\frac{2}{d}} &= \left(\gamma^d e^{-(\pi\sigma\gamma)^2}\right)^{\frac{2}{d}} \nonumber{} \\
\epsilon^{\frac{2}{d}}\tfrac{d}{2e\pi^2\sigma^2} &= \gamma^2 e^{-\frac{2}{d}\pi^2\sigma^2\gamma^2} \nonumber{} \\
-\tfrac{2}{d}\pi^2\sigma^2\epsilon^{\frac{2}{d}}\tfrac{d}{2e\pi^2\sigma^2} &= -\tfrac{2}{d}\pi^2\sigma^2\gamma^2 e^{-\frac{2}{d}\pi^2\sigma^2\gamma^2} \nonumber{} \\
-\tfrac{\epsilon^{\frac{2}{d}}}{e} &= -\tfrac{2}{d}\pi^2\sigma^2\gamma^2 e^{-\frac{2}{d}\pi^2\sigma^2\gamma^2} \nonumber{} \intertext{We can solve for $\gamma$ using the Lambert W function, the inverse of $f(x)=x \cdot e^x$...}
-\tfrac{2}{d}\pi^2\sigma^2\gamma^2 &= W_k\left(-\tfrac{\epsilon^{\frac{2}{d}}}{e}\right) \nonumber{} \\
\gamma^2 &= -\frac{d}{2\pi^2\sigma^2} W_k\left(-\tfrac{\epsilon^{\frac{2}{d}}}{e}\right) \nonumber{} \\
\gamma &= \sqrt{-\frac{d}{2\pi^2\sigma^2} W_k\left(-\tfrac{\epsilon^{\frac{2}{d}}}{e}\right)}
\end{align}
It turns out that the minimum and maximum values correspond to the $0$ and $1$-branches of the Lambert W function:
\begin{align}
\gamma^{\sigma}_{\min} &= \sqrt{-\frac{d}{2\pi^2\sigma^2} W_0\left(-\tfrac{\epsilon^{\frac{2}{d}}}{e}\right)} & & & \gamma^{\sigma}_{\max} &= \sqrt{-\frac{d}{2\pi^2\sigma^2} W_{-1}\left(-\tfrac{\epsilon^{\frac{2}{d}}}{e}\right)}
\end{align}
Thus, the frequency ``width'' is:
\begin{align}
\gamma^{\sigma}_{\mathrm{width}} &= \gamma^{\sigma}_{\max} - \gamma^{\sigma}_{\min} = \sqrt{-\frac{d}{2\pi^2\sigma^2} W_{-1}\left(-\tfrac{\epsilon^{\frac{2}{d}}}{e}\right)} - \sqrt{-\frac{d}{2\pi^2\sigma^2} W_0\left(-\tfrac{\epsilon^{\frac{2}{d}}}{e}\right)} \nonumber{} \\
&= \sqrt{\frac{d}{2\pi^2\sigma^2}} \sqrt{-W_{-1}\left(-\tfrac{\epsilon^{\frac{2}{d}}}{e}\right)} - \sqrt{\frac{d}{2\pi^2\sigma^2}} \sqrt{-W_0\left(-\tfrac{\epsilon^{\frac{2}{d}}}{e}\right)} \nonumber{} \\
&= \sqrt{\frac{d}{2\pi^2\sigma^2}}\left(\sqrt{-W_{-1}\left(-\tfrac{\epsilon^{\frac{2}{d}}}{e}\right)} - \sqrt{-W_0\left(-\tfrac{\epsilon^{\frac{2}{d}}}{e}\right)}\right) \nonumber{} \\
&= \frac{\sqrt{d}}{\sqrt{2}\pi\sigma}\left(\sqrt{-W_{-1}\left(-\tfrac{\epsilon^{\frac{2}{d}}}{e}\right)} - \sqrt{-W_0\left(-\tfrac{\epsilon^{\frac{2}{d}}}{e}\right)}\right)
\end{align}
In general, we want to represent more than one ``scale'', $\sigma$. Assuming we have already picked $\gamma_{\min}$ and $\gamma_{\max}$ to represent a range of scales that includes $\sigma$, what fraction of neurons will be more than $\epsilon$-active for the $\sigma$-scale? We use the fact that we chose $p_{\vect{\gamma}}(\vect{\gamma}) \propto \frac{1}{\norm{\vect{\gamma}}}$ to simplify things, by noting that when $\vect{\gamma} \sim p_{\vect{\gamma}}(\vect{\gamma})$, then $\ln(\norm{\vect{\gamma}})$ is uniform between $\ln(\gamma_{\min})$ and $\ln(\gamma_{\max})$. This means the probability mass between $\gamma^{\sigma}_{\min}$ and $\gamma^{\sigma}_{\max}$ is given by:
\begin{align}
\mathrm{P}(\gamma^{\sigma}_{\min} < \norm{\vect{\gamma}} < \gamma^{\sigma}_{\max}) &= \frac{\ln(\gamma^{\sigma}_{\max}) - \ln(\gamma^{\sigma}_{\min})}{\ln(\gamma_{\max}) - \ln(\gamma_{\min})} = \frac{\ln\left(\frac{\gamma^{\sigma}_{\max}}{\gamma^{\sigma}_{\min}}\right)}{\ln\left(\frac{\gamma_{\max}}{\gamma_{\min}}\right)} = \ln\left(\frac{\gamma_{\max}}{\gamma_{\min}}\right)^{-1}\ln\left(\frac{\gamma^{\sigma}_{\max}}{\gamma^{\sigma}_{\min}}\right) \nonumber{} \\
&= \ln\left(\frac{\gamma_{\max}}{\gamma_{\min}}\right)^{-1}\ln\left(\frac{\sqrt{-\frac{d}{2\pi^2\sigma^2} W_{-1}\left(-\tfrac{\epsilon^{\frac{2}{d}}}{e}\right)}}{\sqrt{-\frac{d}{2\pi^2\sigma^2} W_{0}\left(-\tfrac{\epsilon^{\frac{2}{d}}}{e}\right)}}\right) \nonumber{} \\
&= \frac{1}{2}\ln\left(\frac{\gamma_{\max}}{\gamma_{\min}}\right)^{-1}\ln\left(\frac{W_{-1}\left(-\tfrac{\epsilon^{\frac{2}{d}}}{e}\right)}{W_{0}\left(-\tfrac{\epsilon^{\frac{2}{d}}}{e}\right)}\right) \nonumber{} \\
&= \frac{1}{2}\ln\left(\frac{\gamma_{\max}}{\gamma_{\min}}\right)^{-1}\left[W_{0}\left(-\tfrac{\epsilon^{\frac{2}{d}}}{e}\right) - W_{-1}\left(-\tfrac{\epsilon^{\frac{2}{d}}}{e}\right)\right] \nonumber{} \intertext{When $\epsilon=0.001$, this can be extremely closely approximated by...}
&\approx 5.257\cdot\ln\left(\frac{\gamma_{\max}}{\gamma_{\min}}\right)^{-1}\left(\frac{\sqrt{d+1.83}}{d}\right)
\end{align}
From which it's evident that as $d$ increases, the overall number of neurons that will be ``active'' at a given $\sigma$ shrinks. In higher dimensions, a smaller range of scales can be represented using the same number of neurons. Another important point, though, is that our choice of density for $\vect{\gamma}$ has resulted in our representational power being scale-invariant, in the sense that the number of ``active'' neurons doesn't depend on $\sigma$.

\pagebreak
\noindent{}\textbf{Acknowledgments}

\noindent{}Funding: This work was supported by the Okinawa Institute of Science and Technology Graduate University (doctoral-student stipend). This research did not receive any specific grant from funding agencies in the public, commercial, or not-for-profit sectors.

\bibliographystyle{elsarticle-num} 

\input{aij_paper.bbl}


%
%
%
\end{document}

%% file: macros.tex
\usepackage{graphicx}%
\usepackage{multirow}%
\usepackage{amsmath,amssymb,amsfonts}%
\usepackage{amsthm}%
\usepackage{mathrsfs}%
\usepackage{mathtools}
\usepackage[title]{appendix}%
\usepackage{xcolor}%
\usepackage{textcomp}%
\usepackage{manyfoot}%
\usepackage{booktabs}%
\usepackage{algorithm}%
\usepackage{algorithmicx}%
\usepackage[noend]{algpseudocode}%
\usepackage{listings}
\usepackage{xfrac}
\usepackage[many]{tcolorbox}
\usepackage{tabularx}
\usepackage{xparse}
\usepackage{etoolbox}
\usepackage{stmaryrd}
\usepackage{framed}
\usepackage{relsize}
\usepackage{scalerel}
\usepackage{calc}
\usepackage{geometry}

\makeatletter
\newcommand{\raisemath}[2]{\mathpalette{\raise@math{#1}}{#2}}
\newcommand{\raise@math}[3]{%
  \raisebox{#1}[0pt][0pt]{$#2#3$}}   
\makeatother

\newcommand{\vect}[1]{\boldsymbol{#1}}
\newcommand{\mat}[1]{\mathbf{#1}}
\newcommand{\norm}[1]{\lVert #1 \rVert}
\newcommand{\bind}{\triangleright}
\newcommand{\abs}[1]{\vert #1 \vert}
\makeatletter
\newsavebox{\@brx}
\newcommand{\llangle}[1][]{\savebox{\@brx}{\(\m@th{#1\langle}\)}%
  \mathopen{\copy\@brx\kern-0.5\wd\@brx\usebox{\@brx}}}
\newcommand{\rrangle}[1][]{\savebox{\@brx}{\(\m@th{#1\rangle}\)}%
  \mathclose{\copy\@brx\kern-0.5\wd\@brx\usebox{\@brx}}}
\makeatother
\newcommand{\cosim}[2]{\llangle #1, #2 \rrangle}
\newcommand{\n}{\raisemath{0.3ex}{\text{-}}}
\newcommand{\idx}[1]{\pmb{[}#1\pmb{]}}
\DeclareMathOperator*{\cartprod}{\mathop{\scalerel*{\times}{\prod}}\displaylimits}

\newcommand{\preminus}[3]{\prescript{\raisemath{#1ex}{\texttt{-}}\hspace{#2ex}}{}{#3}}
\newcommand{\preplus}[3]{\prescript{\raisemath{#1ex}{\texttt{+}}\hspace{#2ex}}{}{#3}}
\newcommand{\start}[1]{#1^{\circ}}
\newcommand{\goal}[1]{#1^{\star}}
\newcommand{\subgoal}[1]{#1^{\triangleright}}

\renewcommand{\SS}{S}
\newcommand{\s}{\vect{s}}
\newcommand{\K}{K}
\DeclareRobustCommand{\SSK}{\ThisStyle{%
  \ensuremath{%
    \SavedStyle{\SS}_{\SavedStyle\scaleobj{0.5}{\hspace{-0.25ex}\boldsymbol{\K}}}%
  }%
}}
\newcommand{\WP}{\mathcal{S}}
\newcommand{\Kf}{\preminus{0}{-0.2}{\mathsf{x}}}
\newcommand{\Ks}{\preplus{0}{-0.2}{\mathsf{x}}}
\newcommand{\termtime}[2]{T^{\K}_{\raisemath{0.1ex}{#1, #2}}}
\newcommand{\plantermtime}[1]{T^{\K}_{#1}}
\DeclareRobustCommand{\rK}{\ThisStyle{%
	\ensuremath{%
		\SavedStyle{r}_{\SavedStyle\scaleobj{0.5}{\hspace{-0.25ex}\boldsymbol{\K}}}%
	}%
}}
\DeclareRobustCommand{\rKe}[1]{\ThisStyle{%
	\ensuremath{%
		\SavedStyle{r}^{#1}_{\SavedStyle\scaleobj{0.5}{\hspace{-0.25ex}\boldsymbol{\K}}}%
	}%
}}
\DeclareRobustCommand{\erK}{\ThisStyle{%
	\ensuremath{%
		\SavedStyle{\tilde{r}}_{\SavedStyle\scaleobj{0.5}{\hspace{-0.25ex}\boldsymbol{\K}}}%
	}%
}}
\newcommand{\epotential}{\tilde{\potential}}
\newcommand{\potentialbar}{\potential'}
\newcommand{\segElt}[1]{\segedges^{<}(#1)}
\newcommand{\segEgte}[1]{\segedges^{\geq}(#1)}
\newcommand{\traj}{\mathfrak{t}}
\newcommand{\trajK}{\traj_{\K}}
\newcommand{\plan}{\mathfrak{s}}
\newcommand{\con}{\operatorname{con}}
\newcommand{\pre}{\operatorname{pre}}
\newcommand{\conK}{\con_{\K}}
\newcommand{\preK}{\pre_{\K}}
\newcommand{\eps}{\vect{\epsilon}}
\newcommand{\plans}{\mathfrak{S}}
\newcommand{\tube}[1]{\mathcal{N}_{#1}}
\newcommand{\stube}[1]{\preplus{-0.1}{-0.5}{\tube{#1}}}
\newcommand{\ball}[2]{B_{#2}(#1)}

\newcommand{\q}{q}
\newcommand{\product}[1]{\llbracket#1\rrbracket}

\newcommand{\mapflag}[2]{%
  \def#2{}%
  \ifstrequal{#1}{r}{\def#2{\raisemath{-0.25ex}{\texttt{+}}\hspace{-.2ex}}}{}%
  \ifstrequal{#1}{e}{\def#2{\raisemath{-0.25ex}{\vect{\epsilon}}\hspace{-.15ex}}}{}%
  \ifstrequal{#1}{f}{\def#2{\raisemath{0.2ex}\triangleright\hspace{-.3ex}}}{}%
  \ifstrequal{#1}{c}{\def#2{\circ}}{}%
}

\NewDocumentCommand{\splans}{ O{} O{} m m }{%
  \mapflag{#1}{\splUp}%
  \mapflag{#2}{\splLow}%
  \ensuremath{
    \ifblank{\splUp}{%
      \ifblank{\splLow}{%
        \plans_{\hspace{-.2ex}#3}^{#4}%
      }{%
        \prescript{}{#2}{\plans}_{\hspace{-.2ex}#3}^{#4}%
      }%
    }{%
      \ifblank{\splLow}{%
        \prescript{\splUp}{}{\plans}_{\hspace{-.2ex}#3}^{#4}%
      }{%
        \prescript{\splUp}{\splLow}{\plans}_{\hspace{-.2ex}#3}^{#4}%
      }%
    }%
  }%
}

\NewDocumentCommand{\splansf}{ O{} O{} m m m}{
  \splans[#1][#2]{#3}{#4}\hspace{-0.4ex}(#5)
}

\newcommand{\paths}{\mathfrak{P}}

\NewDocumentCommand{\allpaths}{ O{} O{} m m }{%
  \mapflag{#1}{\splUp}%
  \mapflag{#2}{\splLow}%
  \ensuremath{
    \ifblank{\splUp}{%
      \ifblank{\splLow}{%
        \paths_{\hspace{-.2ex}#3}^{#4}%
      }{%
        \prescript{}{#2}{\paths}_{\hspace{-.2ex}#3}^{#4}%
      }%
    }{%
      \ifblank{\splLow}{%
        \prescript{\splUp}{}{\paths}_{\hspace{-.2ex}#3}^{#4}%
      }{%
        \prescript{\splUp}{\splLow}{\paths}_{\hspace{-.2ex}#3}^{#4}%
      }%
    }%
  }%
}

\NewDocumentCommand{\allbaths}{ O{} O{} m m }{%
  \mapflag{#1}{\splUp}%
  \mapflag{#2}{\splLow}%
  \ensuremath{
    \ifblank{\splUp}{%
      \ifblank{\splLow}{%
        \baths_{\hspace{-.2ex}#3}^{#4}%
      }{%
        \prescript{}{#2}{\baths}_{\hspace{-.2ex}#3}^{#4}%
      }%
    }{%
      \ifblank{\splLow}{%
        \prescript{\splUp}{}{\baths}_{\hspace{-.2ex}#3}^{#4}%
      }{%
        \prescript{\splUp}{\splLow}{\baths}_{\hspace{-.2ex}#3}^{#4}%
      }%
    }%
  }%
}

\NewDocumentCommand{\allpathsf}{ O{} O{} m m m}{
  \allpaths[#1][#2]{#3}{#4}(#5)
}

\newcommand{\achclasssym}{\xi}
\newcommand{\achclass}[3]{\achclasssym^{\raisemath{-0.1ex}{#1}}_{#2, #3}}
\newcommand{\achsetsym}{\Xi}
\DeclareRobustCommand{\achset}[2]{\achsetsym_{\raisemath{0.2ex}{\scaleobj{0.75}{#2}}}^{\scaleobj{0.75}{#1}}}

\newcommand{\hball}{\mathrm{\mathbf{b}}}
\newcommand{\hballs}{\mathcal{B}}
\newcommand{\refine}{\texttt{refine}}

\newcommand{\try}{\texttt{try}}
\newcommand{\bath}{\mathfrak{b}}
\newcommand{\baths}{\mathfrak{B}}
\newcommand{\extend}{\texttt{extend}}
\newcommand{\link}{\texttt{link}}
\newcommand{\eactive}[1]{#1^{\texttt{+}}}
\newcommand{\einactive}[1]{#1^{\texttt{-}}}
\newcommand{\efailed}[1]{#1^{\times}}
\newcommand{\eforcedinactive}[1]{#1^{\downarrow}}

\newcommand{\p}{\mathfrak{p}}

\newcommand{\actionspace}{A}

\newcommand{\pwave}{\mathcal{P}_{\hspace{-2pt}\circledcirc}}
\newcommand{\pathPsym}{\mathscr{P}}
\newcommand{\pathP}[3]{\pathPsym_{\raisemath{0.1ex}{#1}}^{#2\rightarrow#3}}
\newcommand{\goalPsym}{\mathscr{G}}
\newcommand{\goalP}{\goalPsym_{\epistemicscore}}
\newcommand{\agent}{\mathcal{A}}
\newcommand{\graph}{G}
\newcommand{\segraph}{\mathcal{G}}
\newcommand{\segverts}{\mathcal{V}}
\newcommand{\segedges}{\mathcal{E}}
\newcommand{\vertices}{V}
\newcommand{\edges}{E}
\newcommand{\vertex}{\mathrm{\mathbf{v}}}
\newcommand{\edge}{\mathrm{\mathbf{e}}}
\newcommand{\potential}{\alpha}
\newcommand{\measure}{M}
\newcommand{\connection}{\mathrm{\mathbf{c}}}

\newcommand{\field}{F}

\newcommand{\epistemicscore}{\Upsilon}
\newcommand{\inhbpos}{\Delta^+}
\newcommand{\inhbneg}{\Delta^-}
\newcommand{\expect}[1]{\mathbb{E}\left[#1\right]}

\newcommand{\APFC}{\kappa}
\newcommand{\ftaua}{f^{\Lambda}_{1}}
\newcommand{\ftaub}{f^{\Lambda}_{2}}
\newcommand{\ftauc}{f^{\Lambda}_{3}}
\newcommand{\stress}{\Delta \tilde{q}}

\newcommand{\mybox}[1]{\textsf{[\,#1\,]}}

\newcommand{\itemA}{\mybox{a}}
\newcommand{\itemB}{\mybox{b}}
\newcommand{\itemC}{\mybox{c}}
\newcommand{\itemD}{\mybox{d}}
\newcommand{\itemE}{\mybox{e}}
\newcommand{\itemF}{\mybox{f}}
\newcommand{\itemG}{\mybox{g}}
\newcommand{\itemH}{\mybox{h}}
\newcommand{\itemI}{\mybox{i}}
\newcommand{\itemJ}{\mybox{j}}
\newcommand{\itemK}{\mybox{k}}
\newcommand{\itemL}{\mybox{l}}
\newcommand{\itemM}{\mybox{m}}

\newcommand{\itemP}{\mybox{p}}
\newcommand{\itemQ}{\mybox{q}}
\newcommand{\itemR}{\mybox{r}}

\newcommand{\contains}{\gtrdot}